\documentclass[lettersize,journal]{IEEEtran}
\usepackage{amsmath,amsfonts}
\usepackage{algorithm}
\usepackage{multirow}
\usepackage{algpseudocode}
\usepackage{array}
\usepackage[caption=false,font=normalsize,labelfont=sf,textfont=sf]{subfig}
\usepackage{textcomp}
\usepackage{stfloats}
\usepackage{url}
\usepackage{verbatim}
\usepackage{graphicx}
\def\BibTeX{{\rm B\kern-.05em{\sc i\kern-.025em b}\kern-.08em
    T\kern-.1667em\lower.7ex\hbox{E}\kern-.125emX}}
\usepackage{balance}

\usepackage{enumitem}
\usepackage{tikz}
\usepackage{makecell}

\newcommand\circled[1]{\tikz[baseline=(char.base)]{
    \node[shape=circle,draw,inner sep=0.5pt] (char) {#1};}}

\newcommand{\etal}{\textit{et al.}}

\begin{document}

\title{DeRelayL: Sustainable Decentralized Relay Learning}

\author{Haihan Duan,~\IEEEmembership{Member,~IEEE},
Tengfei Ma,
Yuyang Qin,
Runhao Zeng,~\IEEEmembership{Member,~IEEE}, 
Wei Cai,~\IEEEmembership{Senior Member,~IEEE},
Victor C. M. Leung,~\IEEEmembership{Life Fellow,~IEEE},
Xiping Hu,~\IEEEmembership{Member,~IEEE}
\thanks{This work was supported by Guangdong-Hong Kong-Macao Joint Laboratory for Emotional Intelligence and Pervasive Computing, Artificial Intelligence Research Institute, Shenzhen MSU-BIT University. (Corresponding Author: Xiping Hu)}
\thanks{Haihan Duan and Runhao Zeng are with Artificial Intelligence Research Institute, Shenzhen MSU-BIT University, Shenzhen 518172, China, and also with Guangdong-Hong Kong-Macao Joint Laboratory for Emotion Intelligence and Pervasive Computing, Shenzhen 518172, China. E-mail: duanhaihan@smbu.edu.cn, runhaozeng.cs@gmail.com.}
\thanks{Tengfei Ma and Yuyang Qin are with Artificial Intelligence Research Institute, Shenzhen MSU-BIT University, Shenzhen 518172, China, also with Guangdong-Hong Kong-Macao Joint Laboratory for Emotion Intelligence and Pervasive Computing, Shenzhen 518172, China, and also with The Chinese University of Hong Kong, Shenzhen, Shenzhen 518172, China. E-mail: 121090406@link.cuhk.edu.cn, yuyangqin1@link.cuhk.edu.cn.}
\thanks{Wei Cai is with School of Engineering and Technology, University of Washington, Tacoma, WA 98402-3100, USA. E-mail: weicaics@uw.edu.}
\thanks{Victor C.M. Leung is with Artificial Intelligence Research Institute, Shenzhen MSU-BIT University, Shenzhen 518172 China, and also with the Department of Electrical and Computer Engineering, University of British Columbia, Vancouver, BC V6T 1Z4, Canada. E-mail: vleung@ieee.org.}
\thanks{Xiping Hu is with Artificial Intelligence Research Institute, Shenzhen MSU-BIT University, Shenzhen 518172, China, also with Guangdong-Hong Kong-Macao Joint Laboratory for Emotion Intelligence and Pervasive Computing, Shenzhen 518172, China, and also with the School of Medical Technology, Beijing Institute of Technology, Beijing 100090, China. E-mail: huxp@bit.edu.cn.}
}

\markboth{Journal of \LaTeX\ Class Files,~Vol.~18, No.~9, September~2020}%
{How to Use the IEEEtran \LaTeX \ Templates}

\maketitle

\begin{abstract}
In the era of big data, large-scale machine learning models have revolutionized various fields, driving significant advancements. However, large-scale model training demands high financial and computational resources, which are only affordable by a few technological giants and well-funded institutions. In this case, common users like mobile users, the real creators of valuable data, are often excluded from fully benefiting due to the barriers, while the current methods for accessing large-scale models either limit user ownership or lack sustainability. This growing gap highlights the urgent need for a collaborative model training approach, allowing common users to train and share models. However, existing collaborative model training paradigms, especially federated learning (FL), primarily focus on data privacy and group-based model aggregation. To this end, this paper intends to address this issue by proposing a novel training paradigm named decentralized relay learning (DeRelayL), a sustainable learning system where permissionless participants can contribute to model training in a relay-like manner and share the model. In detail, this paper presents the architecture and workflow of DeRelayL, designs incentive mechanisms to ensure sustainability, and conducts theoretical analysis and numerical simulations to demonstrate its effectiveness.
\end{abstract}


\section{Introduction}

\IEEEPARstart{I}{n} the era of big data, the rise of large-scale machine learning models has revolutionized various fields, from natural language processing to computer vision, healthcare, education, and beyond. These models are usually trained on enormous datasets and have demonstrated remarkable capabilities in terms of accuracy, generalization, and predictive ability, which drives significant advancements in technology and scientific discovery \cite{chang2024survey}. Moreover, the development and deployment of large models are no longer merely trends but necessities for fully capitalizing on the opportunities presented by the big data era. Therefore, as the volume of data continues to grow exponentially, advanced machine learning techniques need to harness the information from the available data. 

However, a major problem has emerged: although individuals, organizations, and even small enterprises often possess valuable data, completely training large-scale models is obviously beyond the financial and technical ability of common users \cite{kaplan2020scaling, zhang2024scaling}. Specifically, the computational resources required to train a large-scale model completely can only be afforded by a few technological giants and well-funded institutions \cite{chang2024survey}. These institutions may purchase training data from third parties, or even directly scrape data from websites without payment, thus the original creators of the knowledge (common users) find it hard to obtain profits. As a result, this forms a growing gap between the giants and common users with insufficient computational resources (e.g., common users using only mobile devices), limiting common users from enjoying the societal benefits of intelligent data-driven insights in the big data era created by themselves \cite{duan2024incentive}.

In practice, using the application of large language models (LLMs) \cite{chang2024survey} as an example, common users can actually utilize LLMs based on two approaches provided by the giants. (1) The first way is \textbf{close-source}, in which the common users need to pay for the use. The prospective users are usually charged by monthly subscription or accumulation of utilized input/output tokens. In this case, the common users can only obtain the rights to use the models online, but the model weights are not directly accessible to them, i.e., the common users do not truly possess the models. (2) The second approach is \textbf{open-source}, where some giants may voluntarily contribute well-trained models for the public to download freely. The common users can really obtain model weights in this situation, and the giants can earn non-monetary profit, such as reputation. However, the open-source approach is hard to achieve sustainability, since there lacks an explicit monetary incentive to maintain the model update \cite{alami2024free}. To this end, this study seeks to address this pressing issue by proposing a sustainable decentralized learning system in which participants can train like a relay and collaboratively share the model, named \underline{De}centralized \underline{Relay} \underline{L}earning (DeRelayL), i.e., participants who have sufficient contributions to the relay-like model training can possess the trained model weights, acting like \textbf{semi-open-source}.

In recent years, some researchers have discussed the collaborative model training methodologies. Among them, federated learning (FL) is the most notable framework for collaborative model training with privacy preservation \cite{mcmahan2017communication, ruan2024optimal}. Traditional FL relies on centralized model parameter aggregation and faces challenges like performance degradation with non-IID data \cite{mcmahan2017communication, hsieh2020non, karimireddy2020scaffold}. Moreover, other researchers also investigate decentralized FL \cite{beltran2023decentralized}, as well as blockchain-enabled FL \cite{qu2022blockchain}, which mainly studies decentralized model parameter aggregation \cite{he2018cola, li2022learning} and decentralization-related topics \cite{kim2019blockchained, li2020blockchain, qu2021proof, ramanan2020baffle, kim2019blockchain, fan2020hybrid, li2021blockchain, qin2024blockdfl, ekuban2023towards, zhang2021incentive}. However, the core motivation of FL differs significantly from the proposed DeRelayL, where FL typically revolves around the challenges of the group-based aggregation process, but DeRelayL focuses more on motivating independent participants to sustainably contribute to and benefit from model training. Besides the FL, other existing studies also have investigated collaborative model training, discussing collaborations in volunteer computing environments \cite{diskin2021distributed, ryabinin2020towards}, secure multi-party collaborative model training \cite{zheng2021cerebro, kang2024tiny}, decentralized LLM training \cite{gao2023gradientcoin}, etc. The most relevant study named relay learning presented by Bo \etal \cite{bo2023relay}, but they focused on security and privacy issues in relay-like model training between clinical multi-sites, so the motivation and application are quite different from our study.

In this paper, we aim to build a sustainable decentralized learning system based on blockchain, where permissionless participants can collaboratively train and share models. The models will be passed among participants in the learning system, following a relay-like learning process. In each round, the model evolves based on the previous round’s updates, creating a continuous chain of collaborative learning. Only the participants who have contributed to the model training or maintaining the system operation can share the models, which could ensure that each participant’s contribution is recognized and rewarded. Supported by blockchain, this procedure fosters a decentralized and collaborative learning environment, where different trainers can take over the process at different stages, allowing for a more flexible and efficient model development. Therefore, the paradigm operates like a relay, where the task is passed from one participant to another in sequence, with each participant contributing their part in a coordinated manner, so-called \textbf{\underline{De}centralized \underline{Relay} \underline{L}earning (DeRelayL)}.

The contributions of this paper can be concluded as follows:
\begin{itemize}[leftmargin=1.5em]
	\item This paper proposes a novel learning paradigm regarding the sustainable decentralized collaborative model training. Specifically, we introduce the architecture of DeRelayL based on blockchain and present a detailed system workflow. To the best of our knowledge, there are few existing studies that share the same considerations.
	\item To maintain the sustainability of DeRelayL, we formulate the utilities of all participants and design a corresponding incentive mechanism to guarantee the participants' Individual Rationality (IR) and Incentive Compatibility (IC).
        \item This paper conducts a detailed theoretical analysis to formulate a condition set for the incentive mechanism. Moreover, we also design a numerical simulation to demonstrate the effectiveness of the incentive mechanism and the sustainability of DeRelayL.
        \item Due to the complexity of DeRelayL, this paper can only present the key motivation and system workflow, while some techniques in realistic implementation are not mature enough. To this end, we also comprehensively discuss the potential challenges and research directions of DeRelayL.
\end{itemize}

\section{Related Work} \label{sec_related_work}

In this section, we mainly discuss existing collaborative model training paradigms, including federated learning (FL) and other related collaborative training methods, to clarify the different motivations and scenarios between the existing methodologies and the proposed DeRelayL.

\subsection{Federated Learning}

Federated learning (FL) is a distributed machine learning approach that enables multiple nodes to collaboratively train a shared model while keeping local data private \cite{mcmahan2017communication}. Generally, FL assumes that participants exchange model updates with a central server that aggregates them, instead of sharing raw data. Therefore, the typical FL algorithm studies the aggregation of gradients, such as FedAvg \cite{mcmahan2017communication}. On the other hand, some studies point out that FL faces the challenge of performance degradation in non-IID data \cite{hsieh2020non, karimireddy2020scaffold}. Many solutions have been proposed to solve the problem, including optimizing the model aggregation \cite{fraboni2021clustered, wang20federated}, knowledge distillation \cite{lin2020ensemble}, regularizing training in distributed nodes \cite{li2020federated}, and Bayesian reformulation \cite{chen21fedbe}. Besides the basics of FL, some researchers also study related topics, such as balancing personalization and generalization \cite{chen2024page}, acceleration \cite{zhou2024accelerating, wang2023fededge}, privacy and fairness preserving \cite{shi2022energy, zhang2024privacy}.

On the other hand, some researchers have also noticed that traditional FL relies on a centralized server for the aggregation of model parameters or gradients, so decentralized FL has been studied in recent years \cite{beltran2023decentralized}. The decentralized FL mainly focuses on distributed model parameter aggregation between the neighboring participants \cite{he2018cola, li2022learning}. Referring to decentralization, blockchain is the cutting-edge implementation of decentralized systems, and many researchers have paid attention to blockchain-enabled FL \cite{qu2022blockchain}, studying the architecture of decentralized FL \cite{kim2019blockchained}, consensus algorithm \cite{li2020blockchain, qu2021proof, cao2021toward}, decentralized aggregator assignment \cite{ramanan2020baffle, kim2019blockchain}, resource trading and allocation \cite{fan2020hybrid, li2021blockchain}, defending against poisoning attacks \cite{qin2024blockdfl, ekuban2023towards}, incentive mechanism design \cite{zhang2021incentive}, etc.

\begin{figure}[b!]
	\centering
	\includegraphics[width=1.0\columnwidth]{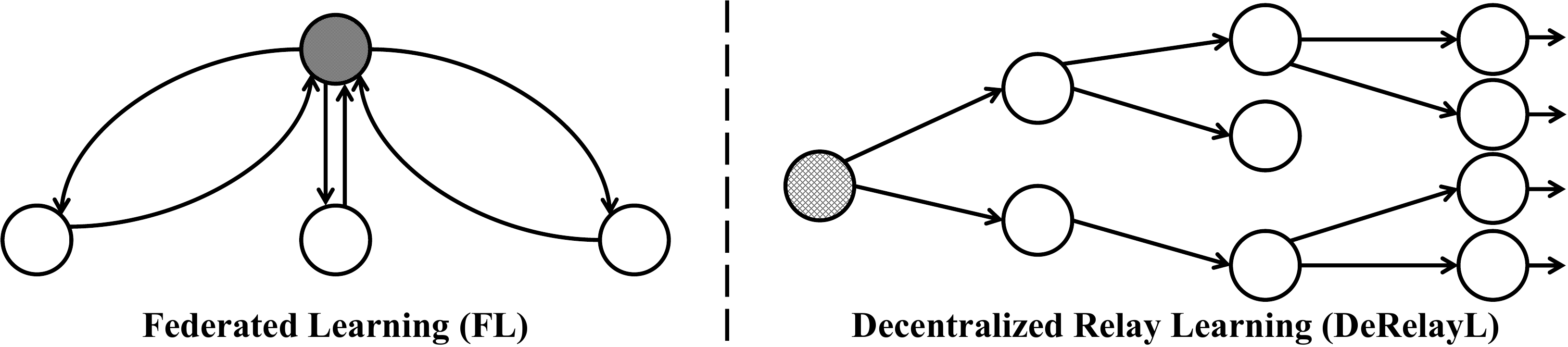}
	\caption{A comparative diagram between federated learning and relay learning.}
	\label{fig_FLvsRelayL}
\end{figure}

However, the motivation of FL shows a significant difference from that of the proposed DeRelayL. As shown in Figure \ref{fig_FLvsRelayL}, we demonstrate a comparative diagram with the simplest case to show the difference between FL and DeRelayL. FL mainly discusses the collaborative model training among a group, typically involving an aggregation process per round, but DeRelayL emphasizes incentivizing a sustainable collaborative model training, which is expected to present like a relay among multiple independent participants, who can obtain the model if they have contributions to the DeRelayL system. Although Buyukates \etal \cite{buyukates2023proof} presented similar considerations, where they proposed a proof-of-contribution-based design for collaborative machine learning on the blockchain, the trained model still belongs to the initiator rather than the participants. Similarly, subsequent studies of contribution proof can refer to Ebrahimi \etal \cite{ebrahimi2024blockchain} and Yazdaninejad \etal \cite{yazdaninejad2024blockchain}.

\subsection{Other Collaborative Model Training}

Although FL is the best-known paradigm of collaborative model training, some other methodologies present different considerations. Diskin \etal \cite{diskin2021distributed} proposed a framework for distributed deep learning in open collaborations (DeDLOC), addressing the challenges posed by volunteer computing environments. Ryabinin \etal \cite{ryabinin2020towards} introduced a decentralized mixture-of-experts (DMoE) model designed to leverage volunteer computing for training large neural networks, especially in distributing the computational workload across unreliable hardware. Zheng \etal \cite{zheng2021cerebro} presented Cerebro, a platform designed for secure multi-party cryptographic collaborative learning, avoiding exposing sensitive information when combining data from multiple organizations. The most similar motivation was mentioned by Gao \etal \cite{gao2023gradientcoin}, who proposed a theoretical design of a decentralized LLM and used GradientCoin to incentivize model training, but it lacks a verification of the training, so unreliable participants will tend to cheat for incentive without real training. The most relevant study was proposed by Bo \etal \cite{bo2023relay}, in which the authors also applied the term relay learning to present their paradigm. However, they mainly considered the security and privacy issues in relay-like model training between clinical multi-sites (one by one), so the motivation and application are quite different from our study.

Overall, few existing studies about collaborative model training have shown the same consideration as our proposal, which intends to build a decentralized collaborative model training system that can sustainably work.

\section{System Design}

This section will present the system design. First, we will discuss the motivation and challenges regarding the proposed blockchain-based DeRelayL system, which also reflects our core considerations during the architecture design. Then, we will illustrate the system architecture, addressing the aforementioned challenges. At last, the corresponding mechanism design and problem formulation will be investigated.

\subsection{Motivation and Challenges} \label{sec_challenges}

As discussed in Section \ref{sec_related_work} (also refer to Figure \ref{fig_FLvsRelayL}), the DeRelayL shows significant differences compared with FL \cite{yang2019federated}. In conclusion, the motivation of DeRelayL is \textbf{to build a sustainable decentralized learning system in which participants can train like a relay and collaboratively share the model}. To this end, this subsection will first clarify the challenges following with the logic of the proposed motivation.

\textbf{C1:} \textit{Sustainability of the training system.} Generally, the participants of the training system will keep the strategies that can maximize their utilities. Under this assumption, the open-source model is unsustainable, e.g., if everyone can obtain the model without cost, nobody will have the motivation to train the model, since the training process has cost, which will decrease the participant's utility. Therefore, the system design needs to guarantee that the model can be obtained only if the participants have contributed to the whole procedure of DeRelayL, e.g., participating in the model training or supporting the blockchain system operation.

\textbf{C2:} \textit{Model weight leakage before the model training.} Assuming that a participant wants to contribute to the system by training models, the most common and efficient way is to ask the model owner to send the model to the participant. This step faces a risk that the participant may not fulfill the training duty after obtaining the model, especially in a decentralized system, which means that the participant can obtain a model without any cost. To address the challenge, there should be punishment for the dishonest participants, so the prospective trainers should deposit a certain cost before obtaining the model (the deposit cost should be higher than the model's value), which will be returned after honest behavior. In an extreme situation, if the participants cannot finish the model training subjectively/objectively, the process is equivalent to a transaction between the participants and the system, where the participants spend the deposit cost to buy the model. 

\textbf{C3:} \textit{Dishonest model owners that provide fake models to the participants.} As we discussed in \textbf{C2}, the participants need to deposit a certain cost for requesting the model and withdraw the cost after training. However, dishonest model owners may provide fake models to the participants or even do not respond. In this case, the prospective trainers will lose the model and the deposit at the same time. Therefore, the model owner should also deposit a certain cost by constructing a smart contract with the prospective trainers, which will be returned at the same time as the deposit from the trainer is returned. More importantly, under this setting, the model owner and the trainer form a community of interests, so there should be a two-way selection mechanism between model owners and prospective trainers for them to find a reliable partner.

\textbf{C4:} \textit{Evaluation of model training.} In a decentralized system, it is difficult for model trainers to prove the completed training process. For instance, dishonest trainers can add some white noise to the model to pass the check of model hash, claiming that they finished the model training. Therefore, it requires an evaluation to validate the training, which should be provided by a random third party. Then, a new challenge appears, how to determine whether the trainer has honestly finished the training. Moreover, the black-box training of neural networks and different data distributions between trainers and the evaluation data providers will also influence the performance evaluation, e.g., it is hard to avoid that a dishonest but lucky trainer obtained the highest performance by only adding white noise. To this end, a relatively fair evaluation method is necessary.

\textbf{C5:} \textit{Model weight leakage during the performance evaluation.} To evaluate the performance, it requires the output from the trained model by inputting testing data. Obviously, it is unreliable that the trainers test their models by themselves, while the evaluation of a third party will face the risk of model weight leakage during the transmission. To address the challenge, this paper considers applying fully homomorphic encryption (FHE) \cite{martins2017survey} to transmit model weights, where FHE is an encryption scheme that enables functions to be run directly on encrypted data while yielding the same encrypted results as if the functions were run on plaintext \cite{kim2023sharp}. With FHE, it is not necessary to calculate testing output using the original trained model, avoiding the model weight leakage.

\textbf{C6:} \textit{Verification of the performance evaluation.} Following \textbf{C5}, the trained model after FHE can be broadcast in the system, so the encrypted model can be obtained by every user. This means that, if the testing data and public key of FHE are publicly available, every user can verify the claimed performance, according to the easy-to-check principle in a decentralized system. This public verification from other users is the fundamental guarantee of a valid training record. 

\textbf{C7:} \textit{Re-training the model after testing data publication.} The performance evaluation process contains a hidden pre-condition that the testing data should be published before the evaluation. Therefore, it is possible that trainers re-train their models based on training data to pass the performance evaluation. To solve the problem, the system should have a mechanism to guarantee the models in the performance evaluation are trained before publishing the testing data.

\textbf{C8:} \textit{Collusion between testing data publisher, performance verifier, and trainer.} Decentralized systems have a common challenge that some participants may collude with each other. The folking of blockchain can naturally address the challenge since other honest participants who find out the dishonest behavior will spontaneously follow the honest blocks. Globally, the collusion can only obtain short-term benefits, while, at a long-term level, honest participants will share more powerful models with the increase of the blockchain. Therefore, rational participants will behave normally to seek long-term benefits.

\subsection{System Architecture and Workflow} \label{sec_system}

\begin{figure*}[t!]
	\centering
	\includegraphics[width=2.0\columnwidth]{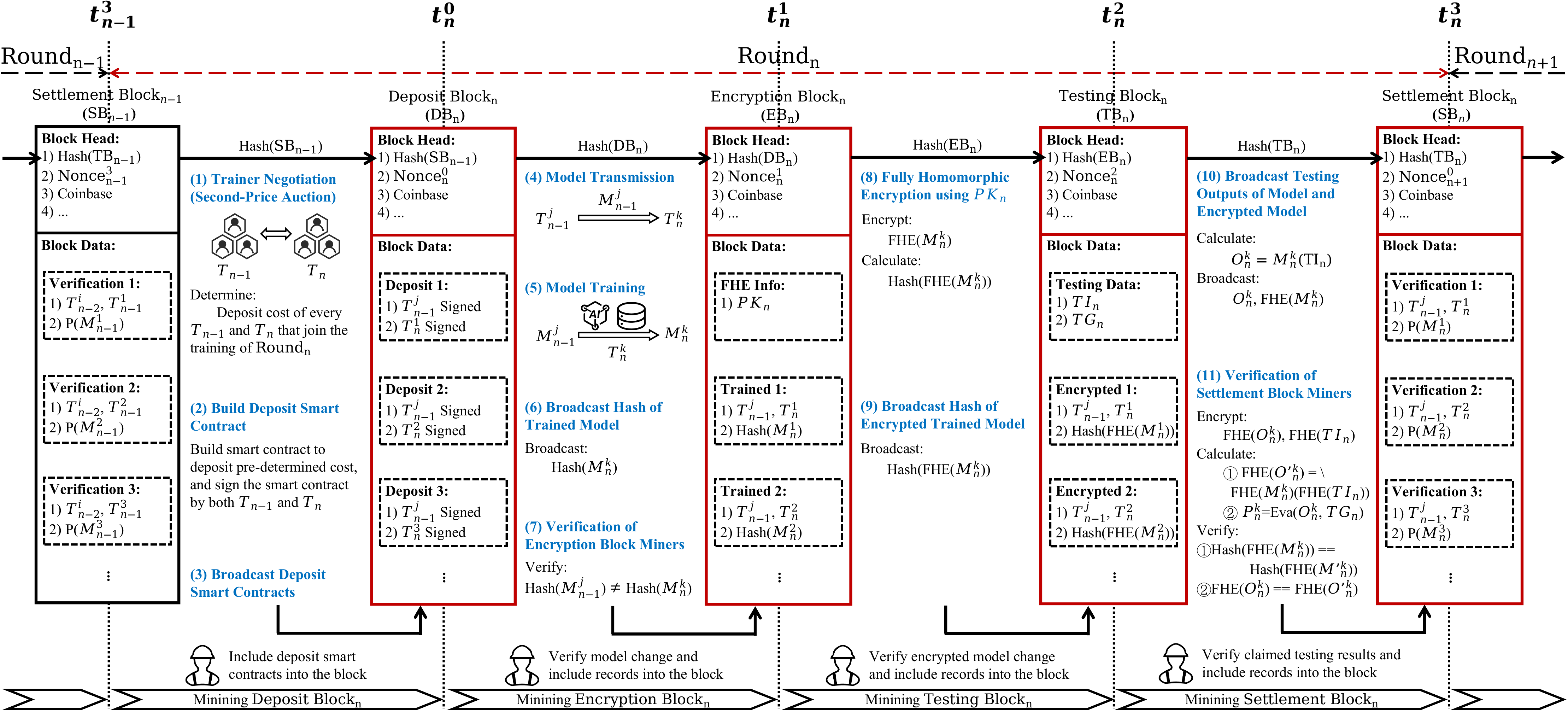}
        \caption{System architecture and workflow of sustainable decentralized relay learning (DeRelayL).}
	\label{fig_architecture}
\end{figure*}

After discussing the design motivation and challenges in Section \ref{sec_challenges}, this subsection will introduce the system architecture of sustainable \underline{De}centralized \underline{Relay} \underline{L}earning (DeRelayL), as shown in Figure \ref{fig_architecture}, by discussing the blockchain design, user roles, and workflow.

The DeRelayL system is based on blockchain, where we define four kinds of blocks according to the different functions during different stages, including deposit block, encryption block, testing block, and settlement block. \textbf{1) Deposit Block (DB):} to store deposit smart contracts, which are the fundamental guarantee of model transmission; \textbf{2) Encryption Block (EB):} to publish information about FHE and record the hash value of trained models; \textbf{3) Testing Block (TB):} to publish testing data and record the hash value of trained models encrypted by the public key in \textbf{Encryption Block} using FHE; \textbf{4) Settlement Block (SB):} to verify and store the performance of trained models. Note that, the mining process is identical for all blocks, where the core difference is that the miners should act in different roles and include different data corresponding to the stage. By default, we utilize the Proof of Work (PoW) \cite{nakamoto2008bitcoin} consensus model for block generation as an example, while other consensus models are also available. The detailed usage of each block will be introduced after the discussion of the workflow. In Figure \ref{fig_architecture}, we illustrate the $SB_{n-1}$ in $Round_{n-1}$ and all blocks in $Round_{n}$. The whole workflow of the DeRelayL system has 11 main steps:

\textbf{(1) Trainer Negotiation (Second-Price Auction).} As we mentioned, SB records the performance of trained models, as well as their corresponding trainers and resources (details of SB will be discussed in \textbf{Step (11)}). Therefore, after the confirmation of $SB_{n-1}$, all nodes in the system can find the models' performance of the (n-1)$-th$ training round. Abstractly, all participants can be denoted as trainers $T$, where $T_{n-1}$ are trainers in $Round_{n-1}$ and also the model owners in $Round_{n}$, and $T_{n}$ are trainers who will negotiate with model owners $T_{n-1}$ to obtain a model for training in $Round_{n}$. This negotiation will determine that the trainers $T_{n}$ will follow which model owner $T_{n-1}$ to train the model, which is a two-way selection procedure. Moreover, each of $T_{n-1}$ and $T_{n}$ should also determine the deposit cost to participate in $Round_{n}$. In this paper, we simply apply the second-price auction \cite{friedman2018double} to complete the procedure (details will be discussed in Section \ref{sec_negotiation}), while other two-way selection methods may also fit this scenario. Note that this paper assumes that the single trainer can only complete the training of one model in each round, but model owners can select multiple trainers if they have enough coins to deposit (also known as total deposit budget $\mathcal{B}$ in Section \ref{sec_negotiation}).

\textbf{(2) Build Deposit Smart Contract.} After the two-way selection, $T_{n-1}$ and $T_{n}$ who completed the procedure should build a smart contract to deposit the predetermined cost. The smart contract should be signed by both $T_{n-1}$ and $T_{n}$. The deposit will be returned if the trained model shows good performance in the testing data, but it also faces a risk of loss if the trainers cannot perform normally or even do not finish the training (details can refer to \textbf{Step (11)}). Therefore, the two-way selection in \textbf{Step (1)} is necessary, because both $T_{n-1}$ and $T_{n}$ hope to find a reliable partner to minimize risk.

\textbf{(3) Broadcast Deposit Smart Contracts.} The deposit smart contracts will be broadcast after the double signature. DB miners will include the smart contracts and construct $DB_{n}$ at $t^0_n$ when they find $Nounce^0_n$. The DB block contains a coinbase transaction to earn a mining reward, which depends on the included deposit smart contracts, prompting miners to pack records as much as possible. This setting is similar to gas fees on a public blockchain (e.g., BitCoin \cite{nakamoto2008bitcoin}), while the incentive is from the mechanism rather than the users.

\textbf{(4) Model Transmission.} After the publishing of DB at $t^0_n$, the system enters the training period, where trainers $T_{n}$ will receive the model from $T_{n-1}$. In Figure \ref{fig_architecture}, we illustrate an example that $T^k_{n}$ will receive a model $M^j_{n-1}$ from $T^j_{n-1}$.

\textbf{(5) Model Training.} After the model transmission, the trainer $T^k_{n}$ will train the model using the computational resource and data, where we use $M^k_{n}$ to denote the trained model. In an unreliable decentralized system, some trainers may fail to finish the model training due to unknown reasons. Correspondingly, as a punishment, the deposit of both $T_{n-1}$ and $T_{n}$ will not be returned.

\textbf{(6) Broadcast Hash of Trained Model.} After the model training, trainers will broadcast the hash value of trained models to claim that they completed the training process, denoted as $Hash(M^k_{n})$. 

\textbf{(7) Verification of Encryption Block Miners.} In this step, EB miners will check the hash value of the trained model, verifying $Hash(M^j_{n-1}) \ne Hash(M^k_{n})$, which means the updated model is at least different from the original one. Thanks to the transparency of blockchain, the required information can be easily obtained from the previous blocks, e.g., obtain $T^j_{n-1}$ from $DB_n$ according to $T^k_{n}$, and then obtain $Hash(M^j_{n-1})$ from $EB_{n-1}$ based on $T^j_{n-1}$. When finding $Nounce^1_n$ at $t^1_n$, the EB miners will include all records that passed the hash check, containing metadata ($T^j_{n-1}$ and $T^k_{n}$) and $Hash(M^k_{n})$. More importantly, there is a necessary step that EB miners need to calculate a public key $PK_n$ for FHE. EB miners are special compared with other block miners, because they can obtain trained models by their private key of FHE. On the one hand, due to the cost of generating the FHE key pairs, malicious miners may upload a random number as $PK_n$, influencing the subsequent steps of the system. On the other hand, the model is provided as a reward, incentivizing miners to actively participate and preventing EB miners to access valuable models and exit the system without further contribution. Therefore, obtaining a trained model can guarantee incentive compatibility (IC) for EB miners and motivate them to behave honestly. As long as EB miner does not disrupt the process and their actions contribute positively to the overall integrity of the system, the system remains sustainable.

\textbf{(8) Fully Homomorphic Encryption using $PK_n$.} After the publishing of EB at $t^1_n$, all participants of the system can obtain $PK_n$. Then, the trainers can encrypt their trained model using FHE by $PK_n$, denoted as $FHE(M^k_{n})$, and the corresponding hash value can be formulated as $Hash(FHE(M^k_{n}))$. This step can fix the trained models before publishing the testing data, ensuring the trainers cannot re-train the models. 

\textbf{(9) Broadcast Hash of Encrypted Trained Model.} The trainers will broadcast $Hash(FHE(M^k_{n}))$ after model encryption. The TB miners will include the hash values and construct $TB_{n}$ at $t^2_n$ when they find $Nounce^2_n$. Besides the hash values, the TB miners should also provide testing data to evaluate the performance of trained models, including testing input $TI_n$ and testing ground truth $TG_n$ (if the testing data is too large to store in the blockchain, the TB miners can also provide a decentralized storage address of the data). The testing data are publicly available, which corresponds to three advantages: 1) all participants can verify the results of performance evaluation, ensuring the procedure is valid with consensus of most participants; 2) the quality of the testing data can be supervised by all participants, and other participants can choose to folk the training blockchain if the quality is unsatisfactory; 3) the testing data can be utilized as training data in the next round, globally contributing additional information to the whole DeRelayL system.

\textbf{(10) Broadcast Testing Outputs of Model and Encrypted Model.} After the publishing of TB at $t^2_n$, all trainers are accessible to the testing data $TI_n$ and $TG_n$. The trainers can calculate the testing outputs of their models as $O^k_{n} = M^k_{n}(TI_n)$. Then, the trainers will broadcast the outputs $O^k_{n}$ and the encrypted models $FHE(M^k_{n})$ (discussed in \textbf{Step (8)}) for performance evaluation. 

\textbf{(11) Verification of Settlement Block Miners.} The last step is most critical, which involves four parts: model performance verification, packing valid training records, returning qualified deposits, and additional citation reward. \textbf{1) Model performance verification:} The SB miners should verify whether the received outputs $O^k_{n}$ are generated from $M^k_{n}$ based on FHE, without obtaining the original model $M^k_{n}$. The SB miners will encrypt received outputs $O^k_{n}$ and testing input $TI_n$ from $TB_n$ using the $PK_n$ from $EB_n$, denoted as $FHE(O^k_{n})$ and $FHE(TI_n)$. Then, SB miners will calculate new outputs using received encrypted models $FHE(M^k_{n})$ and encrypted testing input $FHE(TI_n)$, denoted as $FHE(O'^k_{n}) = FHE(M^k_{n})(FHE(TI_n))$. Moreover, the SB miners will also calculate a quantitative performance index $P^k_n$ based on pre-defined evaluation metrics (e.g., accuracy, mean-square error (MSE), precision), using the received outputs $O^k_{n}$. After that, the SB miners will firstly verify whether the received encrypted model is the one confirmed in $TB_n$, i.e., $Hash(FHE(M^k_{n}))$ should be equal to received $Hash(FHE(M'^k_{n}))$. Secondly, the SB miners will verify whether the received outputs are calculated from the claimed model, i.e., $FHE(O'^k_{n})$ should be equal to $FHE(O^k_{n})$ according to the features of FHE \cite{martins2017survey}. Note that, due to the transparency, the verification can be checked by any other participant, ensuring the validity. \textbf{2) Packing valid training records:} After the verification, SB miners will include valid training records into the block $SB_n$, including the original model owners $T^j_{n-1}$, current trainers $T^k_{n}$, and corresponding model performance indexes $P^k_n$. \textbf{3) Returning qualified deposit:} Generally, the verification records in $SB_n$ means that the trainer $T^k_{n}$ has finished the training process, and the smart contracts built in \textbf{Step (2)} will return the deposit to both $T^j_{n-1}$ and $T^k_{n}$. However, as discussed in \textbf{C4} of Section \ref{sec_challenges}, lazy workers can add very subtle white noise into the original model, which will not significantly influence the model performance. In this case, the updated model can also pass the verification, which means lazy workers can obtain a model at nearly free cost. To address the problem, we set a threshold to increase the risks of participating in the training system, where only the trained models which has performance ranking in top-$\mathcal{K}$ can return the deposit cost for both original model owners $T^j_{n-1}$ and trainers $T^k_{n}$. However, due to some uncertain factors, such as testing data distribution and randomness in model training, the top-$\mathcal{K}$ ranking mechanism cannot completely filter lazy workers, e.g., the model from a very lucky lazy worker may achieve better performance than others. The aforementioned case is very special with a low possibility, because a fundamental assumption of the system is that the model performance will generally increase with the honest model training behavior. In fact, the introduction of the top-$\mathcal{K}$ ranking mechanism will also change the trainer negotiation in \textbf{Step (1)}, in which both original model owners $T^j_{n-1}$ and trainers $T^k_{n}$ will be more serious when evaluating and selecting their partners. \textbf{4) Additional citation reward:} Besides returning the deposit, we also design a mechanism to reward the original model owners $T^j_{n-1}$ of the top-$\mathcal{K}$ models in $Round_{n}$, named citation reward. Note that, the original model owners $T^j_{n-1}$ may not be the top-$\mathcal{K}$ models in $Round_{n-1}$. Therefore, although there might be some very lucky lazy workers who occupied the top-$\mathcal{K}$ positions in $Round_{n-1}$, the citation reward can motivate honest trainers at a long-term level, since the trainers $T^k_{n}$ may not choose the model to follow only based on the ranking of $Round_{n-1}$, while other information of the original model owners $T^j_{n-1}$ (e.g., historical rankings, frequency of participation) will also be considered.

With the increase of training rounds, the above 11 steps will repeat until the model ability converges to an ultimate level without sufficient performance increment due to the limitation of the model size, referring to scaling law \cite{kaplan2020scaling, zhang2024scaling}.

\subsection{Formulation and Mechanism Design} \label{sec_mechanism}

In Section \ref{sec_system}, we utilize $T_{n-1}$ and $T_{n}$ to denote original model owners and trainers for a general understanding of the cyclic system. In the following parts, we will apply abbreviations of each role to better explain the formulation, i.e., model owner (MO), trainer (T), deposit block miner (DBM), encryption block miner (EBM), testing block miner (TBM), and settlement block miner (SBM).

\subsubsection{Trainer Negotiation Algorithm} \label{sec_negotiation}

In \textbf{Step (1)} of Section \ref{sec_system}, the trainers will negotiate with the original model owners to obtain an opportunity to join the model training, which is a two-way selection that also determines the deposit cost of the model owners and trainers. The two-way selection can be very complex by considering many factors such as historical ranking and participation frequency, but, to simplify the mechanism modeling in this paper, we design a second-price auction \cite{friedman2018double} that greedily selects model owners with higher ranking and trainers with higher deposit willingness. At first, the model owners will broadcast their pre-determined deposit cost $b^{MO}$ to the trainers. The prospective trainers (totally $Q_T$ trainers) will send sealed messages to model owners to honestly provide their reserve deposit bids $b^{T_i}$. After that, the model owners can rank the prospective trainers based on the deposit bids. Then, a model owner can greedily select prospective trainers following Algorithm \ref{alg_selection} constrained by the total deposit budget $\mathcal{B}$ (to select total $\lfloor \frac{\mathcal{B}}{b^{MO}} \rfloor$ trainers). The model owners will invite the selected trainers to build deposit smart contracts by depositing the second price, and, correspondingly, the prospective trainers will greedily accept invitations from higher-ranking model owners. 

\begin{algorithm}[h!]
\caption{Trainer Selection Based on Deposit Bids}\label{alg_selection}
\begin{algorithmic}[1]
\State \textbf{Input:} $\mathbb{T} = \{(T_1, b^{T_1}), (T_2, b^{T_2}), ..., (T_{Q_{T}}, b^{T_{Q_T}})\}$ as the set of trainers and their corresponding deposit bids $b^{T_i}$, MO's deposit cost for one trainer $b^{MO}$, and MO's total deposit budget $\mathcal{B}$.
\State \textbf{Output:} Selected Trainers and their deposit amounts.
\State $Q_{\text{Selected}} \gets \lfloor \frac{\mathcal{B}}{b^{MO}} \rfloor$ 
\State $\mathbb{T}_{\text{Sorted}} = Sort(\mathbb{T}, 2)$ \Comment{Sorted by bids in descending}
\For{$i = 1$ to $Q_{\text{Selected}} - 1$}
    \State $\mathbb{T}_{\text{Selected}}[i] \gets \mathbb{T}_{\text{Sorted}}[i][1]$ \Comment{Trainers} 
    \State $\mathbb{D}_{\text{Deposit}}[i] \gets \mathbb{T}_{\text{Sorted}}[i+1][2]$ \Comment{Deposits} 
\EndFor
\State $\mathbb{T}_{\text{Selected}}[Q_{\text{Selected}}] \gets \mathbb{T}_{\text{Sorted}}[Q_{\text{Selected}}][1]$ \Comment{Trainers}
\State $\mathbb{D}_{\text{Deposit}}[Q_{\text{Selected}}] \gets \mathbb{T}_{\text{Sorted}}[Q_{\text{Selected}}][2]$ \Comment{Deposits} 
\State \textbf{Return} $\mathbb{T}_{\text{Selected}}$, $\mathbb{D}_{\text{Deposit}}$
\end{algorithmic}
\end{algorithm}

\subsubsection{Problem Formulation and Incentive Mechanism Design} \label{sec_incentive}

To maintain sustainability, the incentive of DeRelayL (e.g., mining reward, model weights) should at least satisfy \textbf{Individual Rationality (IR)} and \textbf{Incentive Compatibility (IC)} \cite{fan2020hybrid, paris2014efficient, sun2022profit, duan2024incentive}:
\begin{itemize}[leftmargin=1.5em]
	\item \textbf{Individual Rationality:} All participants of the DeRelayL system should obtain a non-negative utility. Otherwise, the rational participants will not participate in the model training of the DeRelayL system.
	\item \textbf{Incentive Compatibility:} The incentive mechanism of the DeRelayL system should ensure that participants with normal behavior can obtain the maximum utility, which means that behaving normally is the optimal strategy for each participant.
\end{itemize}

Therefore, we will first formulate the utility (U) of each participant in the DeRelayL system based on the revenue (R) and cost (C). Key annotations are summarized in Table \ref{tab_symbol}.

\begin{table*}[!t]
	\caption{Key Annotations (In the Order of Appearance)}
	\label{tab_symbol}
	\resizebox{\textwidth}{!}{
    \renewcommand{\arraystretch}{1.4}
		\begin{tabular}{lp{9.5cm}|lp{9.5cm}}
  		\hline
            \multicolumn{1}{c}{\textbf{Symbol}} & \multicolumn{1}{c|}{\textbf{Description}} & \multicolumn{1}{c}{\textbf{Symbol}} & \multicolumn{1}{c}{\textbf{Description}}  \\
            \hline
            $U^{\text{Participant}}$ & Utility of Participant, $Participant \in \{MO, T, DBM, EBM, TBM, SBM\}$ & $R^{\text{Participant}}$ &  Revenue of Participant, $Participant \in \{MO, T, DBM, EBM, TBM, SBM\}$\\
            $C^{\text{Participant}}$ & Cost of Participant, $Participant \in \{MO, T, DBM, EBM, TBM, SBM\}$ & $R_{\text{Now}}^{MO}$ &  Immediate revenue of MO available now\\
            $R_{\text{Future}}^{MO}$ & Revenue of MO available in the future& $C_{\text{Deposit}}^{MO}$ & Cost incurred by MO due to potential deposit loss risk \\
            $C_{\text{Transmit}}^{MO}$ & Cost incurred by MO for transmitting model parameters & $\overline{Q_{\text{Selected}}^{MO}}$ & Average number of Trainers selected by MO \\
            $\mathcal{R}_{\text{Cited}}$ & Citation reward for MO when its model is successfully trained (cited) by one Trainer, noticing all $\mathcal{R}$ adjustable by the system & $\beta$ & Discount rate for T or MO on $\mathcal{R}_{\text{Cited}}$, used to estimate future potential rewards \\
            $Q_{\text{Selected}}$ & Number of Deposits signed by an MO in a given round & $s$ & Selection rate of trained models, $0 < s < 1$, e.g., $s = 0.5$ means half of the trained models are selected \\
            $b^{MO}$ & MO's bid: number of coins deposited by MO in the Deposit smart contract for each T & $k_{\text{Transmit}}$ & Transmission coefficient, multiplied by the number of model parameters, representing the cost of receiving or transmitting the model \\
            $|M|$ & Number of model parameters & $U_{N}^{MO}$ & Utility of MO under the strategy ``Normal"  \\
            $C_{\text{DepositLoss}}^{MO}$ & Cost to MO due to deposit loss from intentionally failing to transmit model parameters & $U_{NTm}^{MO}$ & Utility of MO under the strategy ``Not Transmitting" model parameters \\
            $R_{\text{Now}}^{T}$ & Immediate revenue of T available now & $R_{\text{Future}}^{T}$ &  Revenue of T available in the future \\
            $C_{\text{Train}}$ & Cost of training a model & $C_{\text{Deposit}}^{T}$ & Cost incurred by T due to potential deposit loss risk \\
            $C_{\text{RecM}}^{T}$ & Cost incurred by T when receiving model parameters from MO & $C_{\text{Encrypt}}$ & Cost of encrypting parameters using public key of FHE from EBM \\
            $C_{\text{Broadcast}}$ &  Cost of broadcasting the FHE-encrypted model parameters & $R_{\text{RecM}}$ & The model received by T itself, which could serve as a reward in the formulation  \\
            $R_{\text{TrainedM}}$ & The model trained by T itself, which could serve as a reward in the formulation  & $R_{\text{Cited}}^{T}$ &  Citation reward for T when its model is successfully trained (cited) by future Ts\\
            $V_{\text{RecM}}$ & Version of received model & $V_{\text{Now}}^{T}$ & Version of the latest model owned by T just now before receiving new model \\
            $\circled{C}$ & One unit of reward coin & $\overline{Q_{\text{Selected}}^{T}}$ & Average number of selected future Ts who will train model based on the T directly or indirectly \\
            $P_{\text{Comp}}$ & Price of computation per unit of time and per unit of data volume & $D$ & Data volume \\
            $\tau$ & Training time & $b^{T}$ & T's bid (the number of coins deposited by T)  \\
            $k_{\text{Encrypt}}$ & Cost of encrypting per unit of model parameters & $Q_{\text{Broadcast}}$ & Number (Quantity) of recipients that T needs to broadcast the encrypted model parameters to \\
            $k_{\text{Expand}}$ & Factor by which FHE encryption expands parameter size & $U_{N}^{T}$ & Utility of T under the strategy ``Normal" \\
            $C_{\text{DepositLost}}^{T}$ & Deposit lost by T due to poor behaviors & $U_{NTr}^{T}$ & Utility of T under the strategy ``Not Training" \\
            $U_{NBr}^{T}$ & Utility of T under the strategy ``Not Broadcasting" & $R^{DBM}_{\text{Include}}$ & Revenue for DBM from including deposit information into the blockchain \\
            $C_{\text{Mine}}$ & Cost of mining under the PoW consensus mechanism & $Q_{\text{Deposit}}$ & Quantity of deposits included on-chain by one DBM \\
            $\mathcal{R}_{\text{Deposit}}$ & Revenue per deposit included on-chain by DBM & $U^{DBM}_{N}$ & Utility of DBM under the strategy ``Normal" \\
            $Q_{\text{DepositLess}}$ & Quantity of deposits below the expected amount to be included on-chain & $U^{DBM}_{NPA}$ & Utility of DBM under the strategy ``Not Packing All the deposits" \\
            $U^{DBM}_{\text{PI}}$ & Utility of DBM under the strategy ``Packing Improperly"& $R^{EBM}_{\text{Include}}$ & Revenue for EBM from including FHE public key and hashes of trained model parameters on-chain \\
            $R_{\text{FHEM}}$ & The revenue represented by the FHE-encrypted model decrypted by EBM & $C^{EBM}_{\text{RecFHEM}}$ & Cost of process of receiving FHE-encrypted model \\
            $C_{\text{GenFHEKey}}$ & Cost of generating a pair of FHE keys including public key and secret key & $Q_{\text{HashM}}$ & Quantity of hashes of model submitted by trainers \\
            $\mathcal{R}_{\text{HashM}}$ & Revenue for EBM per hash of model & $V_{\text{FHEM}}$ & Version of FHE-encrypted model \\
            $V_{\text{Now}}^{EBM}$ & Version of the latest model owned by EBM just now before decrypting FHE-encrypted model& $U^{EBM}_{N}$ &  Utility of EBM under the strategy ``Normal"\\
            $U^{EBM}_{NG}$ & Utility of EBM under the strategy ``Not Generating" FHE keys & $R^{TBM}_{\text{Include}}$ & Revenue for TBM from including testing data cases and hashes of FHE-encrypted models on-chain \\
            $R_{\text{GenTDCases}}$ & Revenue from generating testing data cases & $C_{\text{GenTDCases}}$ & Cost of generating testing data cases \\
            $Q_{\text{EncryptedM}}$ & Quantity of hashes of FHE-encrypted models & $\mathcal{R}_{\text{EncryptedM}}$ & Revenue for TBM per hash of Encrypted model \\
            $Q_{\text{Cases}}$ & Quantity of testing data cases submitted by TBM & $\mathcal{R}_{\text{Case}}$ & Revenue for TBM per testing data case  \\
            $C_{\text{GenTDCase}}^{\text{Unit}}$ & Cost of generating one unit of testing data case & $U^{TBM}_{N}$ & Utility of TBM under the strategy ``Normal" \\
            $U^{TBM}_{IT}$ & Utility of TBM under the strategy ``Improper Testing data"& $R^{SBM}_{\text{Include}}$ & Revenue for SBM from including Testing outputs \(results\) and hashes of FHE-encrypted models on-chain\\
            $R_{\text{Verify}}$ & Revenue for SBM from verifying the performance of FHE-encrypted models using testing data cases & $C^{SBM}_{\text{RecFHEMs}}$ & Cost of process of receiving FHE-encrypted models \\
            $C_{\text{Verify}}$ & Cost of verifying the performance of FHE-encrypted models & $Q_{\text{VerifiedM}}$ & Quantity of verified FHE-encrypted models by SBM \\
            $\mathcal{R}_{\text{VerifiedM}}$ & Revenue for SBM per verified model from including performance of verified models on-chain & $\mathcal{R}_{\text{Verify}}$ & Revenue for SBM per verified model and per testing data case from computing the performance of verified models \\
            $C_{\text{Verify}}^{\text{Unit}}$ & Cost per verified model and per case& $U^{SBM}_{N}$ & Utility of SBM under the strategy ``Normal" \\
            $U^{SBM}_{IRa}$ & Utility of SBM under the strategy ``Improper Ranking"& & \\
            \hline
		\end{tabular}
	}
\end{table*}

\textbf{(1) Model owner (MO):} The utility of MO can be denoted:
\begin{align}
    \begin{split}
    U^{MO} &= R^{MO} - C^{MO} \\
          &= R_{\text{Now}}^{MO} + R_{\text{Future}}^{MO} - C_{\text{Deposit}}^{MO} - C_{\text{Transmit}}^{MO}
    \end{split}
\end{align}
where $R_{\text{Now}}^{MO}$ and $R_{\text{Future}}^{MO}$ refer to the 4) additional citation reward as discussed in \textbf{Step (11)} of Section \ref{sec_system}. $R_{\text{Now}}^{MO}$ is the citation reward of the current round, and $R_{\text{Future}}^{MO}$ is the revenue of the future rounds, which will be calculated as a geometric series since the future revenue has a discount rate. For the cost, MO has deposit cost $C_{\text{Deposit}}^{MO}$ for each round, but the cost is likely to be returned if the selected trainers behave normally. Moreover, MO also has transmission cost $C_{\text{Transmit}}^{MO}$ when sending the model to the selected trainers.

\textbf{(2) Trainer (T):} The utility of T can be represented:
\begin{align}
    \begin{split}
    U^{T} &= R^{T} - C^{T} = R_{\text{Now}}^T + R_{\text{Future}}^T  \\
          & - C_{\text{Train}} - C_{\text{Deposit}}^{T} - C_{\text{RecM}}^{T} - C_{\text{Encrypt}} - C_{\text{Broadcast}}
    \end{split}
\end{align}
where $R_{\text{Now}}^T$ is the revenue of the current round, which contains the revenue of the received model from MO $R_{\text{RecM}}^T$ and the revenue of the model trained by T $R_{\text{TrainedM}}^T$. And $R_{\text{Future}}^T$ is the future revenue for additional citation reward, similar to MO. The cost of T consists of five parts: 1) model training cost $C_{Train}$; 2) deposit cost $C_{\text{Deposit}}^{T}$, which will be returned if behaving normally; 3) cost of receiving the model from MO $C_{\text{RecM}}^{T}$; 4) FHE encryption cost of trained model $C_{\text{Encrypt}}$; 5) cost of broadcasting encrypted model $C_{\text{Broadcast}}$.

\textbf{(3) Deposit block miner (DBM):} The utility of DBM is:
\begin{align}
    \begin{split}
    U^{DBM} & = R^{DBM} - C^{DBM} \\
          & = R^{DBM}_{\text{Include}} - C_{\text{Mine}}
    \end{split}
\end{align}
where $R^{DBM}_{\text{Include}}$ is the incentive of miners to include deposit smart contracts as much as possible, so the revenue is proportional to the quantity of included data. $C_{\text{Mine}}$ is the cost of mining the block, i.e., the computational cost of the PoW consensus model. Note that all miners (DBM, EBM, TBM, SBM) have the aforementioned $R_{\text{Include}}$ and $C_{\text{Mine}}$. We also simply assume the block generation intervals are almost identical for all stages, thus the $C_{\text{Mine}}$ is almost fixed. 

\textbf{(4) Encryption block miner (EBM):} The utility of EBM can be formulated:
\begin{align}
    \begin{split}
    U^{EBM} & = R^{EBM} - C^{EBM}\\
          & = R^{EBM}_{\text{Include}} + R_{\text{FHEM}} - C_{\text{Mine}} - C^{EBM}_{\text{RecFHEM}} - C_{\text{GenFHEKey}}
    \end{split}
\end{align}
where $R^{EBM}_{\text{Include}}$ is the incentive of including trained models' information, containing metadata and hash values. Since the EBM is responsible for generating the FHE key pair, the EBM can use the private key to decrypt encrypted models, as discussed in \textbf{Step (7)} of Section \ref{sec_system}. Thus, $R_{\text{FHEM}}$ is the revenue for decrypting encrypted models, and $C^{EBM}_{\text{RecFHEM}}$ is the cost for receiving the encrypted model (EB can only receive and decrypt the best one). $C_{\text{Mine}}$ is the mining cost, and $C_{\text{GenFHEKey}}$ is the cost of generating the FHE key pair.

\textbf{(5) Testing block miner (TBM):} The utility of TBM is:
\begin{align}
    \begin{split}
    U^{TBM} & = R^{TBM} - C^{TBM} \\
          & = R^{TBM}_{\text{Include}} + R_{\text{GenTDCases}} - C_{\text{Mine}} - C_{\text{GenTDCases}} 
    \end{split}
\end{align}
where $R^{TBM}_{\text{Include}}$ is the incentive of including information of encrypted models using FHE, containing metadata and hash values. The TBMs are responsible for generating testing data, so they will be rewarded $R_{\text{GenTDCases}}$ according to the number of testing cases. Thus, there are corresponding costs of generating testing cases $C_{\text{GenTDCases}}$. Similar to other miners, $C_{\text{Mine}}$ is the mining cost. 

\textbf{(6) Settlement block miner (SBM):} The utility of SBM can be formulated:
\begin{align}
    \begin{split}
    U^{SBM} & = R^{SBM} - C^{SBM} \\
          & = R^{SBM}_{\text{Include}} + R_{\text{Verify}} - C_{\text{Mine}} - C^{SBM}_{\text{RecFHEMs}} - C_{\text{Verify}} 
    \end{split}
\end{align}
where $R^{SBM}_{\text{Include}}$ is the incentive of including verification confirmation details, containing metadata and performance index. The SBMs are responsible for verifying the performance of trained models, so they will be rewarded $R_{\text{Verify}}$ according to the number of verified models, corresponding to the cost for receiving all encrypted models $C^{SBM}_{\text{RecFHEMs}}$ and verifying them $C_{\text{Verify}}$. $C_{\text{Mine}}$ is the mining cost.

\begin{table*}[t!]
\renewcommand{\arraystretch}{2.0} 
\centering
\caption{Utilities of Different Participants' Strategies}
\label{tab_utilities}
\resizebox{2.0\columnwidth}{!}{
\begin{tabular}{|p{1.7cm}|p{4.7cm}|p{13cm}|}
\hline
\makecell[c]{\textbf{Participant}} & \makecell[c]{\textbf{Strategy}} & \makecell[c]{\textbf{Utility}} \\ \hline

\makecell[c]{\multirow{2}{*}{MO}} & Normal (N) & $U_{N}^{MO} = \frac{\overline{Q_{\text{Selected}}^{MO}} \cdot \mathcal{R}_{\text{Cited}}}{1 - \beta}- Q_{\text{Selected}} \cdot (1 - s) \cdot b^{MO} - k_{\text{Transmit}} \cdot |M|$ \\ \cline{2-3}
                    & Not Transmitting (NTm) & $U_{NTm}^{MO} = - Q_{\text{Selected}} \cdot b^{MO}$ \\ \hline

\makecell[c]{\multirow{3}{*}{T}}  & Normal (N) & \multicolumn{1}{l|}{$U_{N}^{T} = \left( V_{\text{RecM}} - V_{\text{Now}}^{T} + 1 \right) \cdot \circled{C} + \frac{\overline{Q_{\text{Selected}}^T} \cdot \beta \cdot \mathcal{R}_{\text{Cited}}}{1 - \beta}$} \\
                    &            & \multicolumn{1}{l|}{$- \left[ P_{\text{Comp}} \cdot D \cdot \tau \cdot |M| + (1 - s) \cdot {b}^{T} \right. + k_{\text{Transmit}} \cdot |M| + k_{\text{Encrypt}} \cdot |M| + Q_{\text{Broadcast}} \cdot k_{\text{Transmit}} \cdot k_{\text{Expand}} \cdot |M| \bigg]$} \\ \cline{2-3}

                    & Not Training (NTr) & $U_{NTr}^{T} = (V_{\text{RecM}} - V_{\text{Now}}^{T}) \cdot \circled{C} - {b}^{T} - k_{\text{Transmit}} \cdot |M|$ \\ \cline{2-3}
                    
                    & Not Broadcasting (NBr) & \multicolumn{1}{l|}{$U_{NBr}^{T} = \left( V_{\text{RecM}} - V_{\text{Now}}^{T} + 1 \right) \cdot \circled{C} - \left[ P_{\text{Comp}} \cdot D \cdot \tau \cdot |M| + {b}^{T} + k_{\text{Transmit}} \cdot |M| \right]$} \\ \hline

\makecell[c]{\multirow{3}{*}{DBM}} & Normal (N) & $U^{DBM}_{N} = Q_{\text{Deposit}} \cdot \mathcal{R}_{\text{Deposit}} - C_{\text{Mine}}$ \\ \cline{2-3}
                     & Not Packing All (NPA) & $U^{DBM}_{NPA} = Q_{\text{DepositLess}} \cdot \mathcal{R}_{\text{Deposit}} - C_{\text{Mine}}$ \\ \cline{2-3}
                     & Packing Improper Deposit Contracts (PI) & $U^{DBM}_{\text{PI}} = -C_{\text{Mine}}$ \\ \hline

\makecell[c]{\multirow{2}{*}{EBM}} & Normal (N) & \multicolumn{1}{l|}{$U^{EBM}_{N} = \left( Q_{\text{HashM}} \cdot \mathcal{R}_{\text{HashM}} + (V_{\text{FHEM}} - V_{\text{Now}}^{EBM}) \cdot \circled{C} \right) - \left( C_{\text{Mine}} + k_{\text{Transmit}} \cdot k_{\text{Expand}} \cdot |M| + C_{\text{GenFHEKey}} \right)$} \\ \cline{2-3}
                     & Not Generating FHE Key (NG) & $U^{EBM}_{NG} = -C_{\text{Mine}}$ \\ \hline

\makecell[c]{\multirow{2}{*}{TBM}} & Normal (N) & \multicolumn{1}{l|}{$U^{TBM}_{N} = Q_{\text{EncryptedM}} \cdot \mathcal{R}_{\text{EncryptedM}} + Q_{\text{Cases}} \cdot \mathcal{R}_{\text{Case}} - C_{\text{Mine}} - Q_{Cases} \cdot C_{{\text{GenTDCase}}}^{\text{Unit}}$} \\ \cline{2-3}
                     & Improper Testing Cases (IT) & $U^{TBM}_{IT} = -C_{\text{Mine}}$ \\ \hline

\makecell[c]{\multirow{2}{*}{SBM}} & Normal (N) & \multicolumn{1}{l|}{$U^{SBM}_{N} = Q_{\text{VerifiedM}} \cdot \mathcal{R}_{\text{VerifiedM}} + Q_{\text{VerifiedM}} \cdot Q_{\text{Cases}} \cdot \mathcal{R}_{\text{Verify}} - C_{\text{Mine}} - Q_{\text{VerifiedM}} \cdot k_{\text{Transmit}} \cdot k_{\text{Expand}} \cdot |M|$} \\
                     &            & \multicolumn{1}{l|}{$- Q_{\text{VerifiedM}} \cdot Q_{\text{Cases}} \cdot C_{\text{Verify}}^{\text{Unit}}$} \\ \cline{2-3}
                     & Improper Rank (IRa) & $U^{SBM}_{IRa} = -C_{\text{Mine}}$ \\ \hline

\end{tabular}
}
\end{table*}

For participants, they have different strategies to choose from, which will lead to different utilities. We utilize ``Normal (N)" to denote the participant behaves honestly following the procedure of the DeRelayL system. Specifically, MO may choose to not transmit the model to T (including transmitting fake weights), so the strategy set of MO is \{Normal (N), Not Transmitting (NTm)\}. For Ts, they may choose to not train the model (NTr) or not broadcast the trained model (NBr), so the strategy set of T is \{Normal (N), Not Training (NTr), Not Broadcasting (NBr)\}. The DBM may choose to pack partial deposit smart contracts (NPA) or pack improper ones (PI), thus the strategy set of DBM is \{Normal (N), Not Packing All (NPA), Packing Improper Deposit Contracts (PI)\}. Then, the EBM may not generate the FHE key (NG), so the strategy set of EBM is \{Normal (N), Not Generating FHE Key (NG)\}. For TBMs, they may upload improper testing cases (IT), thus the strategy set of TBM is \{Normal (N), Improper Testing Cases (IT)\}. Finally, the SBM may not rank the trained models properly (IRa), so the strategy set of SBM is \{Normal (N), Improper Rank (IRa)\}. The final utility expressions of each strategy are listed in Table \ref{tab_utilities}, and the detailed formulation, annotation, and explanation of each term can refer to Table \ref{tab_symbol} and Appendix \ref{app_utility}. Note that, in this paper, we assume the knowledge used in model training is almost equal for every round (or satisfies the same distribution), so the model performance incremental from the current round/version to the next round/version is approximate. Therefore, we introduce \circled{C} to denote the value of the knowledge gap between two adjacent model versions, which is also the unit of measurement to unify the different values formulated in the DeRelayL system, as well as for issuing incentives (cryptocurrency/coin). 

After formulating the utilities of each participant, we need to design an incentive mechanism to satisfy \textbf{IR} and \textbf{IC}. Specifically, to satisfy \textbf{IR}, the utilities of the ``Normal" strategy should be no less than 0. According to the calculation in Appendix \ref{app_icir}, the reward ($\mathcal{R}$) of each block should satisfy the following condition set (\textbf{T1 - T8}):

\textbf{T1:} To guarantee \textbf{IR} of MO, we need to let $U^{MO}_{N} \ge 0$:
\begin{equation}
    \mathcal{R}_{\text{Cited}} \ge \frac{(1 - \beta) \cdot \left( Q_{\text{Selected}} \cdot (1 - s) \cdot b^{MO} + k_{\text{Transmit}} \cdot |M| \right)}{\overline{Q_{\text{Selected}}^{MO}}}
\end{equation}

\textbf{T2:} To guarantee \textbf{IR} of T, we need to let $U^{T}_{N} \ge 0$:
\begin{align}
    \mathcal{R}_{\text{Cited}} & \ge \frac{(1 - \beta)}{\overline{Q_{\text{Selected}}^T} \cdot \beta} \cdot \Big( P_{\text{Comp}} \cdot D \cdot \tau \cdot |M| \notag \\
    &\quad + (1 - s) \cdot b^{T} + k_{\text{Transmit}} \cdot |M| + k_{\text{Encrypt}} \cdot |M| \notag \\
    &\quad + Q_{\text{Broadcast}} \cdot k_{\text{Transmit}} \cdot k_{\text{Expand}} \cdot |M| \notag \\
    &\quad - \left( V_{\text{RecM}} - V_{\text{Now}}^{T} + 1 \right) \cdot \circled{C} \Big)
\end{align}

\textbf{T3:} To guarantee \textbf{IR} of DBM, we need to let $U^{DBM}_{N} \ge 0$:
\begin{align}
    \mathcal{R}_{\text{Deposit}}  & \ge \frac{C_{\text{Mine}}}{Q_{\text{Deposit}}}
\end{align}

\textbf{T4:} To guarantee \textbf{IR} of EBM, we need to let $U^{EBM}_{N} \ge 0$:
\begin{align}
    \mathcal{R}_{\text{HashM}} & \ge \frac{1}{Q_{\text{HashM}}} \cdot \Big( C_{\text{Mine}} + k_{\text{Transmit}} \cdot k_{\text{Expand}} \cdot |M| \notag \\
    &\quad + C_{\text{GenFHEKey}} - (V_{\text{FHEM}} - V_{\text{Now}}^{EBM}) \cdot \circled{C} \Big)
\end{align}

\textbf{T5:} To guarantee \textbf{IR} of TBM, we need to let $U^{TBM}_{N} \ge 0$:
\begin{align}
\begin{split}
    Q_{\text{EncryptedM}} \cdot \mathcal{R}_{\text{EncryptedM}} + Q_{\text{Cases}} \cdot \mathcal{R}_{\text{Case}} \\
    \ge C_{\text{Mine}} + Q_{Cases} \cdot C_{{\text{GenTDCase}}}^{\text{Unit}}
\end{split}
\end{align}

\textbf{T6:} To guarantee \textbf{IR} of SBM, we need to let $U^{SBM}_{N} \ge 0$:
\begin{align}
    Q_{\text{VerifiedM}} & \cdot \mathcal{R}_{\text{VerifiedM}} + Q_{\text{VerifiedM}} \cdot Q_{\text{Cases}} \cdot \mathcal{R}_{\text{Verify}} \notag \\
    & \ge C_{\text{Mine}} + Q_{\text{VerifiedM}} \cdot k_{\text{Transmit}} \cdot k_{\text{Expand}} \cdot |M| \notag \\
    &\quad + Q_{\text{VerifiedM}} \cdot Q_{\text{Cases}} \cdot C_{\text{Verify}}^{\text{Unit}}
\end{align}

To satisfy \textbf{IC}, the utilities of the ``Normal" strategy should be greater than other strategies. According to Table \ref{tab_utilities}, some participants' other strategies have negative utilities, so they will choose ``Normal" obviously. Specifically, trainer T requires additional constraints for the strategy of ``Not Training" and ``Not Broadcasting":

\textbf{T7:} For \textbf{IC} of T with ``Not Training", there is a sufficient but not necessary condition that the deposit of T should not be lower than the value of the model (i.e., an effective deposit discussed in Section \ref{sec_system}). Otherwise, T will not have the motivation to train the model.
\begin{align}
\begin{split}
    {b}^{T} > (V_{\text{RecM}} - V_{\text{Now}}^{T}) \cdot \circled{C}
\end{split}
\end{align}

\textbf{T8:} For \textbf{IC} of T with ``Not Broadcasting", let $U_{N}^{T} - U_{NBr}^{T} > 0$, the condition can be formulated:
\begin{align}
    \mathcal{R}_{\text{Cited}} & > \frac{1}{\overline{Q_{\text{Selected}}^T} \cdot \beta} 
    \Big( (- s) \cdot {b}^{T} + k_{\text{Encrypt}} \cdot |M| \notag \\
    &\quad + Q_{\text{Broadcast}} \cdot k_{\text{Transmit}} \cdot k_{\text{Expand}} \cdot |M| \Big) \cdot (1 - \beta)
\end{align}

Detailed derivation of \textbf{IR} and \textbf{IC} can refer to Appendix \ref{app_icir}.

\section{Numerical Simulations}

To evaluate the effectiveness of the proposed DeRelayL system, we conduct a numerical simulation regarding bidding and matching of model owner and trainer, depositing, model training, performance ranking, block mining, as well as the corresponding incentive issuing. The numerical simulation mainly investigates two aspects, including sustainability and accessibility. Open-source simulation codes are available at \url{https://github.com/Tengfei-Ma13206/DeRelayL_Simulation}.

\subsection{Experimental Settings and Procedure} \label{sec_experimental_settings}

In this subsection, we will discuss the experimental settings and procedure. To better explain the numerical simulation, we provide Table \ref{tab_simulation_parameters} to list the parameter settings. The detailed procedure of the numerical simulation is as follows:

\textbf{(1) Role Allocation:} The first step is to randomly allocate the roles of all participants, e.g., randomly selecting Ts. In this simulation, we assume that the total number of participants in the DeRelayL system is fixed at $Q_{\text{TotalParticipants}} = 256$. Among them, the number of miners in each round is fixed at $Q_{\text{Miners}} = 128$. The remaining participants are potential MOs and Ts, with the number being $Q_{\text{MO\&T}} = Q_{\text{TotalParticipants}} - Q_{\text{Miners}} = 128$. Specifically, the MOs were determined based on the previous round, since the MOs in this round were the Ts who luckily ranked at the top positions during the model performance evaluation in the last round. Moreover, for the cold start, we set that only the initiator of the genesis block serves as MO in the first round. Since each MO has limited capacity, we assume that they can only collaborate with up to $Q_{\text{SelectionLimit}} = 4$ Trainers in each round.

\textbf{(2) Model Owner and Trainer Bidding:} After determining the candidate numbers for each role, MOs and Ts start the bidding as discussed in Section \ref{sec_negotiation}. The MO’s bidding strategy is based on a fixed budget, where we directly set $Budget_{\text{MO}} = 0.001$, representing the number of coins the MO possesses, denoted as $Q_{\text{CoinsOwnedByMO}}$. Therefore, the MO deposits for each Trainer can be calculated by $\frac{\min(Budget_{\text{MO}}, Q_{\text{CoinsOwnedByMO}})}{Q_{\text{SelectionLimit}}}$. Then, the bidding strategy for each T is based on the difference in model versions. We present the latest model version as $V_{\text{Latest}}$ and T’s current version as $V_{\text{Now}}^{T}$, and we assume the number of coins a T owns is $Q_{\text{CoinsOwnedByT}}$. Therefore, the bidding deposit is given by $\min(Q_{\text{CoinsOwnedByT}}, V_{\text{Latest}} - V_{\text{Now}}^{T} + 1)$, which means that the older the model version T has, the higher the motivation to place a larger bid. This setting accords the effective deposit (\textbf{T7} discussed in Section \ref{sec_incentive}) to prevent Ts from receiving the latest model without training or broadcasting the model.

\begin{table}[b!]
\centering
\caption{Simulation Parameters Setting}
\label{tab_simulation_parameters}
\resizebox{\columnwidth}{!}{%
\renewcommand{\arraystretch}{1.2}
\begin{tabular}{l|c}
\hline
\makecell[c]{\textbf{Simulation Parameter Description}} & \makecell[c]{\textbf{Value}} \\ \hline
Total number of participants & 256 \\
Number of candidate miners in each round & 128 \\
Number of MOs and candidate Ts & 128 \\
Maximum number of Trainers cooperated by an MO & 4 \\
MO's budget & 0.001 (coins) \\
Probability of successful model training by T & 0.9 \\
Number of testing data cases packaged by TBM & 100 \\
Proportion of Ts selected as successful model trainers & 0.5 \\ 
Reward base for miners, including $\mathcal{R}_{\text{Deposit}}$, $\mathcal{R}_{\text{HashM}}$, & \multirow{2}{*}{0.001 (coins)} \\
$\mathcal{R}_{\text{EncryptedM}}$, $\mathcal{R}_{\text{Case}}$, $\mathcal{R}_{\text{VerifiedM}}$, $\mathcal{R}_{\text{Verify}}$ & \\
\hline
\end{tabular}%
}
\end{table}

\textbf{(3) Model Owner and Trainer Matching:} After bidding, MO and T can be matched based on the deposit. We assume the matching process follows a simple greedy algorithm discussed in Section \ref{sec_negotiation}. In this case, MO will sort Ts in descending order based on the model performance from the previous round, and T will sort MOs based on the number of deposit coins. The first MO selects the top $Q_{\text{SelectionLimit}}$ Ts, the second MO selects the next $Q_{\text{SelectionLimit}}$ Ts, and so on, until all MOs or Ts have found a match participant.

\textbf{(4) Success of Model Training:} After MO and T matching, a miner is randomly selected from the previously grouped miners to complete the PoW mining and is designated as DBM (Note that the ``random selection" of the miner is to simulate the winner of PoW mining, rather than random selection in practical implementation). The DBM includes the deposit records into the DeRelayL blockchain. Afterward, Ts involved in the deposit start to train the model. To simulate the potential accidents that might happen in practice, we assume that Ts have a probability of $Pr_{\text{Training}} = 0.9$ to successfully train a new model, while the remaining Ts may fail to train the model due to various reasons.

\textbf{(5) Performance Ranking:} An EBM is randomly selected from the miners to include FHE keys to the DeRelayL blockchain. Following this, a randomly selected TBM includes a fixed number ($Q_{\text{Cases}} = 100$) of testing data cases into the DeRelayL blockchain. Similarly, a randomly selected SBM includes the testing results to the DeRelayL blockchain. Finally, from those who successfully trained new models, we assume that a fixed proportion of Ts ($s = 0.5$), with flooring applied, are selected to be regarded as successful trainers (i.e., rank in the top positions). Also note that the ``random selection" of the miners is to simulate the winner of PoW mining, rather than random selection in practical implementation.

\textbf{(6) Incentive Issuing:} Based on step \textbf{(5)}, the DeRelayL system will issue rewards to the MO and all predecessor MOs up to the genesis block owner (additional citation reward as discussed in \textbf{Step (11)} of Section \ref{sec_system}), each receiving one $\circled{C}$, where every successful model selection triggers this reward. Moreover, the corresponding incentives (such as mining reward, testing data reward, FHE key generation reward, etc.) will also be settled in this step. Note that, in this simulation, the base reward for miners is set at 0.001 coins, and we simply assume that no one adopts strategies other than ``Normal" due to the IR and IC ensured in Section \ref{sec_incentive}, because the mistakes would lead to forking by honest participants, which will be more effective to exclude them from the simulation in this study. In future research, the consequences of forking in DeRelayL is a promising topic that can be further investigated to discuss its unique attributes and effects.

\subsection{Sustainability}

\begin{figure}[t!]
	\centering
	\includegraphics[width=1.0\columnwidth]{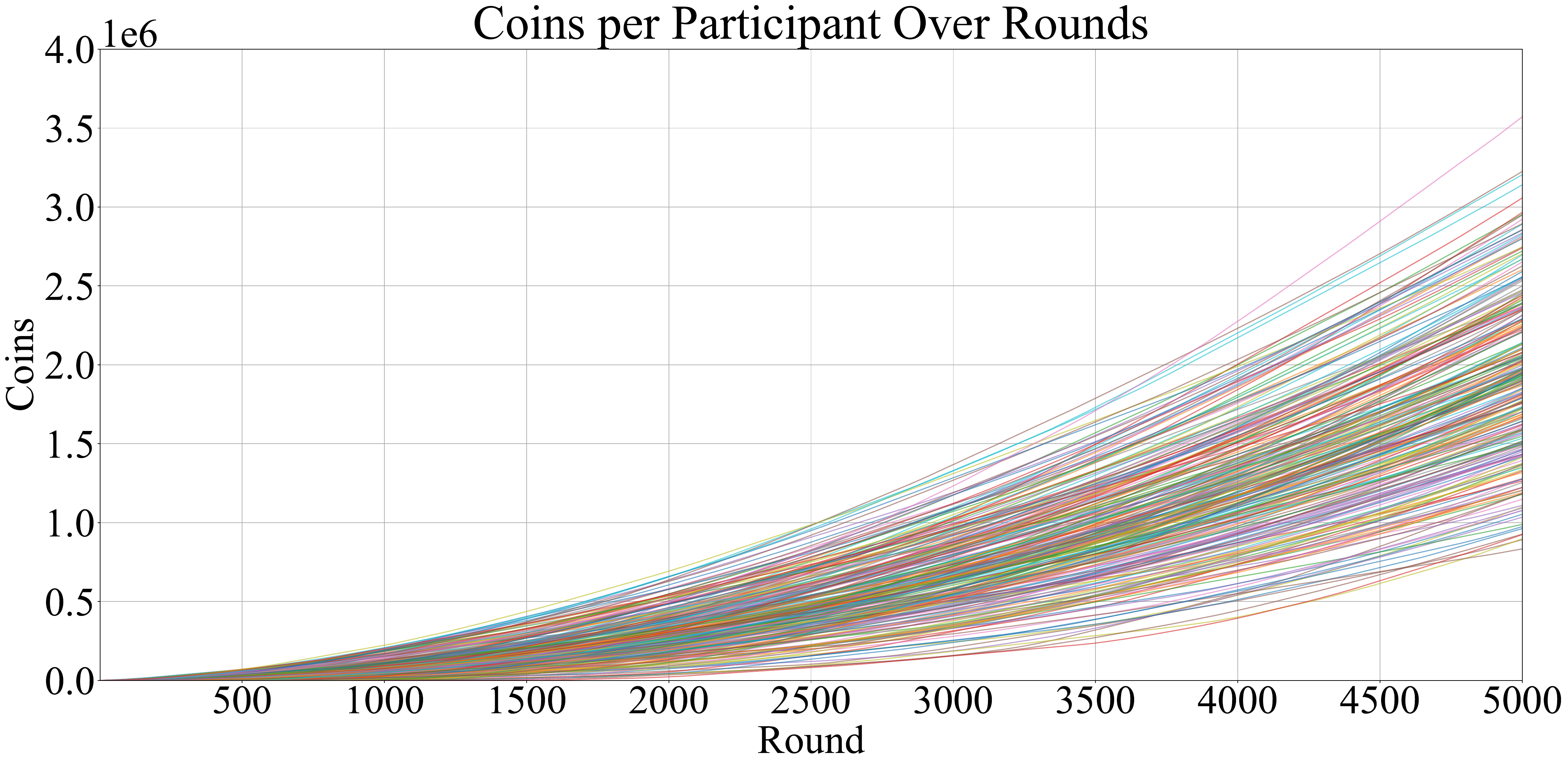}
	\caption{Coins per participants over rounds.}
	\label{fig_coin}
\end{figure}

The core motivation of the proposed DeRelayL is to build a sustainable decentralized learning system, which means that the design of a sustainable training system must first ensure that it remains operational, i.e., participants have incentives to continue their involvement. In detail, a participant's willingness to be involved largely depends on their estimated future rewards. Under stable conditions, with no abrupt changes in rules or participant behavior, their future rewards can be estimated using historical data. Therefore, we measure past rewards in terms of coins accumulated, as the cost of each training round is roughly constant. Then, we plot the changes in coin quantity for each participant as rounds progressed in Figure \ref{fig_coin}. The observed trend indicates that the growth rate of coins accelerates over time, demonstrating the sustainability of the proposed DeRelayL, where all participants have positive estimated future rewards to continue their involvement. To further explain this result, we provide a simple proof:

As discussed in Section \ref{sec_experimental_settings}, we assume that model versions are iteratively updated by selecting from all participants in a round-robin fashion, and every participant successfully uploads their trained model periodically. Simply assume that there are $Q_{\text{Participants}}$ participants, denoted as $P_{1}, P_{2}, \ldots, P_{i_{p}}, \ldots, P_{Q_{\text{Participants}}}$. Moreover, we consider a single DeRelayL blockchain where models are represented by $M_{1}, M_{2}, \ldots, M_{i_{m}}, \ldots, M_{Q_{\text{Model}}}$, where each successive model is trained based on the previous one.

\begin{table}[b!]
\centering
\caption{Key Annotations for the Numerical Simulation (In the Order of Appearance)}
\label{tab_simulation_annotation}
\resizebox{\columnwidth}{!}{%
\renewcommand{\arraystretch}{1.3}
\begin{tabular}{c|l}
\hline
\makecell[c]{\textbf{Variable}} & \makecell[c]{\textbf{Description}} \\ \hline
$Q_{\text{TotalParticipants}}$ & Total number of participants  \\
$Q_{\text{Miners}}$ & Number of candidate miners in each round \\
$Q_{\text{MO\&T}}$ & Number of MOs and candidate Ts \\
$Q_{\text{SelectionLimit}}$ & Maximum number of Trainers cooperated by an MO  \\
$Budget_{\text{MO}}$ & MO's budget  \\
$Q_{\text{CoinsOwnedByMO}}$ & Number of coins owned by MO \\
$V_{\text{Latest}}$ & Latest model version \\
$V_{\text{Now}}^{T}$ & Current model version owned by T \\
$Q_{\text{CoinsOwnedByT}}$ & Number of coins owned by T \\
$Pr_{\text{Training}}$ & Probability of successful model training by T  \\
$Q_{\text{Cases}}$ & Number of testing data cases packaged by TBM  \\
$s$ & Proportion of Ts selected as successful model trainers  \\
$Q_{\text{Participants}}$ & Total number of participants \\
$P_{i_{p}}$ & Participant indexed by $i_{p}$. \\
$M_{i_{m}}$ & Model version of the $i_{m}$-th training round \\
$Q_{\text{Model}}$ & Number of models \\
$i_{m}$ & Model index of training round \\
$i_{p}$ & Participant index of training round \\
$x$ & The $x$-th time the participant $P_{i_{p}}$ trains and uploads a model \\
$Q_{\text{Coins}}$ & Number of coins received by a participant \\
$Q_{\text{LastMO}}$ & Number of MOs in the previous (last) round \\
$Q_{\text{LastT}}$ & Number of Trainers in the previous (last) round \\
$Q_{\text{MO}}$ & Number of MOs in the current round \\
$Q_{\text{T}}$ & Number of Trainers in the current round \\
$N_r$ & Number of Trainers in round $r$ \\ \hline
\end{tabular}%
}
\end{table}

We focus only on the additional citation reward as discussed in \textbf{Step (11)} of Section \ref{sec_system}, which has the greatest impact on coin accumulation. Due to the IR and IC ensured in Section \ref{sec_incentive}, we assume all participants train diligently. Thus, we map each participant $P_{i_{p}}$ to models $M_{i_{p}}, M_{i_{p} + Q_{\text{Participants}}}, M_{i_{p} + 2 \cdot Q_{\text{Participants}}}, \ldots$, and the mapping relationship satisfies $i_{m}\ mod \ Q_{\text{Participants}} = i_{p}$.

When participant $P_{i_{p}}$ trains and uploads a model for the $x$-th time, the number of coins $Q_{\text{Coins}}$ they receive is:
\begin{align}
\begin{split}
       Q_{\text{Coins}} &= Q_{\text{Participants}} + 2 \cdot Q_{\text{Participants}} + \cdots + (x-1) \cdot Q_{\text{Participants}} \\ &= \frac{x(x-1)Q_{\text{Participants}}}{2} = O(x^2) = O((i_{m})^2) 
\end{split}
\end{align}
where the round number here equals $i_{m} = i_{p} + (x-1) \cdot Q_{\text{Participants}}$. Therefore, if every participant actively engages in training, their $Q_{\text{Coins}}$ grows quadratically with the number of rounds, implying that the coin accumulation rate accelerates over time. Detailed annotations can refer to Table \ref{tab_simulation_annotation}.

\subsection{Accessibility}

Besides the sustainability of DeRelayL, it is also crucial to ensure participants can obtain the trained models, denoted by accessibility, which fits the motivation that participants can train and share the model together. To analyze the accessibility, we plot the model version distribution over rounds, as shown in Figure \ref{fig_version}. In this figure, we illustrate the percentage of participants who possess the models of the latest 10 versions (``Latest-v" in Figure \ref{fig_version} denotes the model of the $v$-th version before the current version), older versions, or none. Note that the performance of models within the same version is approximately consistent from a global view. Figure \ref{fig_version} illustrates that all participants can possess models after rounds of training, and their owned models are kept updated following the system operation. Moreover, after convergence, the possession percentage of model versions tends to be stable, because the trainers possessing older models will tend to bid higher to compete for the training opportunity, while those possessing recent models may have less incentive to compete against them. Therefore, the models will naturally be distributed as evenly as possible among all participants. To explore this phenomenon, we also conduct a brief proof:

\begin{figure}[b!]
	\centering
	\includegraphics[width=1.0\columnwidth]{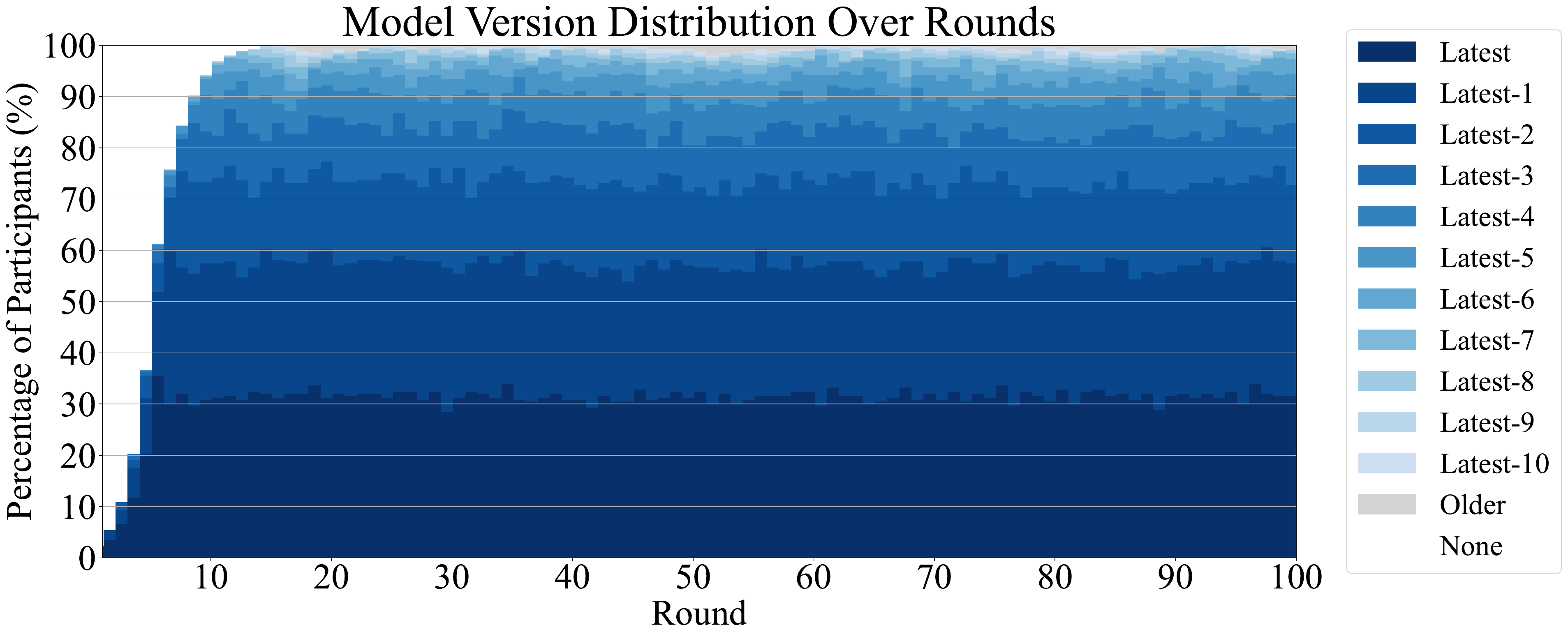}
	\caption{Model version distribution over rounds.}
	\label{fig_version}
\end{figure}

We denote the number of MOs from the previous round as $Q_{\text{LastMO}}$, and the corresponding number of trainers is:
\begin{equation}
    Q_{\text{LastT}} = \min(Q_{MO\&T} - Q_{\text{LastMO}}, Q_{\text{LastMO}} \cdot Q_{\text{SelectionLimit}})
\end{equation}
In the current round, the number of MOs is:
\begin{align}
\begin{split}
       Q_{\text{MO}} & = s \cdot Q_{\text{LastT}} \\ 
       & = s \cdot \left( \min(Q_{MO\&T} - Q_{\text{LastMO}}, Q_{\text{LastMO}} \cdot Q_{\text{SelectionLimit}}) \right) 
\end{split}
\end{align}
And the number of trainers in the current round is:
\begin{equation}
    Q_{\text{T}} = \min(Q_{MO\&T} - Q_{\text{MO}}, Q_{\text{MO}} \cdot Q_{\text{SelectionLimit}})
\end{equation}
Since $Q_{MO\&T}$ is fixed, and $Q_{\text{T}}$ initially grows but is eventually bounded by $Q_{MO\&T}$. Therefore, after convergence, we can obtain:
\begin{equation}
    Q_{\text{T}} = Q_{MO\&T} - Q_{\text{MO}} = Q_{MO\&T} - s \cdot Q_{\text{LastT}}
\end{equation}
Let the number of trainers in round $r$ be $N_r$, we can obtain:
\begin{align}
\begin{split}
       N_{r+1} & = Q_{MO\&T} - s \cdot N_r \\ 
       & = Q_{MO\&T} \cdot \left(1 + (-s) + (-s)^2 + \cdots + (-s)^{i-1}\right) \\
       & \quad + (-s)^i \cdot N_r 
\end{split}
\end{align}
Since $0 < s < 1$ is the proportion of Ts selected as successful model trainers, as $i$ tends to infinity, we can obtain:
\begin{equation}
    N_{r+i} = Q_{MO\&T} \cdot \frac{1}{1 + s} = \frac{Q_{MO\&T}}{1 + s}
\end{equation}
where $\frac{Q_{MO\&T}}{1 + s}$ is also fixed, indicating that the number of trainers remains stable at this value across the training rounds, thus the phenomenon observed in Figure \ref{fig_version} can be demonstrated. Furthermore, the fluctuations in Figure \ref{fig_version} are due to the probabilistic settings of each trainer to simulate unexpected failures of training. Detailed annotations can refer to Table \ref{tab_simulation_annotation}.

\section{Discussion}

By further clarifying the details of the proposed DeRelayL paradigm, we selected some representative and concerning points to conduct a comprehensive discussion in this section.

\subsection{System Anonymity} \label{app_anonymity}

In most public blockchain systems (e.g., Bitcoin \cite{nakamoto2008bitcoin}), anonymity is a key feature designed to mitigate the risk of a 51\% attack, which means an entity controlling over 50\% of the network nodes could potentially alter the blockchain, compromising its tamper-proof nature. However, in DeRelayL, the necessity for strict anonymity seems to be less critical due to the distinct target of the DeRelayL blockchain compared to general public blockchains. Specifically, the primary value of public blockchains lies in cryptocurrency, and any breach of consensus (such as a 51\% attack) would lead to a collapse in value. In contrast, the DeRelayL blockchain's value resides in the trained models, meaning that a 51\% attack would not yield additional profit. Even if an attacker gains control over the DeRelayL blockchain, they cannot meaningfully promote the model improvement, since the actual training is conducted by independent third-party trainers. In this case, the implications of a 51\% attack differ from those in public blockchains. For example, if such an attack occurs during $Round_{n}$, the model performance published in $Round_{n-1}$ remains unaffected, because it is an objective measure that is independent of blockchain consensus. For subsequent rounds, recognizing the 51\% attack, honest participants can fork the blockchain or create a new blockchain-based on the model from $Round_{n-1}$, ensuring the continuity of the latest models without any loss (detailed discussion on forking is provided in Section \ref{app_forking}).

From another perspective, the DeRelayL system is adaptable to different blockchain architectures. In this paper, we utilize a PoW consensus model to enable permissionless participation in model training and system maintenance. However, DeRelayL can also be deployed on a consortium blockchain, where participants are pre-approved by a governing committee. This verification process may eliminate the anonymity of participants, supervising honest behaviors. In this case, some steps in the DeRelayL workflow can be simplified. For instance, a consortium blockchain could employ consensus mechanisms other than PoW to reduce computational costs associated with block mining. Additionally, approved participants could transmit plaintext model data directly, which will also effectively reduce the resource overhead discussed in Section \ref{app_overhead}.

In summary, the DeRelayL paradigm does not strictly emphasize anonymity during system operation and can be adapted to various scenarios through appropriate modifications.

\subsection{Blockchain Forking} \label{app_forking}

As mentioned in Section \ref{app_anonymity}, the DeRelayL blockchain is designed to support flexible forking in the event of attacks or unexpected issues. Therefore, it is important to consider the potential impact that the forking of the DeRelayL blockchain will result in. In general public blockchains (e.g., Bitcoin \cite{nakamoto2008bitcoin}), forking leads to the creation of subchains, and miners must decide which subchain to follow. In theory, miners will tend to follow the longest valid subchain (or longest valid chain), which represents the most computational workers (in PoW systems) or the greatest stake (in Proof of Stake (PoS) systems). This rule helps protect the blockchain against attacks such as double spending and ensures that honest participants who follow the protocol can prevail over malicious users.

In the case of DeRelayL, normal operations will not be affected by forking in terms of model performance improvement. This is because the common model training process is inherently iterative, and model performance naturally fluctuates across training epochs. Generally, the training rounds in DeRelayL can be viewed similarly, where each forking represents an exploration of different potential paths toward a global optimum, and each block within a subchain represents a local optimum along that journey. Additionally, when considering the longest valid subchain, the risk of double spending seems to have minimal impact on the normal operation of the DeRelayL system. For instance, if the DeRelayL system allows forking in $Round_{n}$, malicious users might attempt to double spend by depositing in one subchain with one model owner and in another subchain with a different model owner, aiming to exchange the same coins for multiple models. It seems that malicious users can obtain many models through the double spending attack. However, the models they obtain will be from the same version of $Round_{n}$, which will have comparable performance from a global aspect. Furthermore, the dishonest behavior will be recorded on the blockchain, and their deposits will not be returned, preventing the attackers from joining in subsequent training rounds and obtaining newer models.

Thus, the forking of DeRelayL exhibits different attributes compared to general public blockchains. However, to seriously investigate the consequences, it is necessary to conduct a dedicated study focusing on its unique features and effects.

\subsection{Necessity of the Four-Stage Process} \label{sec_four_stage}

In this paper, we propose a four-stage process of the DeRelayL, including Deposit Block (DB), Encryption Block (EB), Testing Block (TB), and Settlement Block (SB), which is designed to ensure the integrity, consistency, fairness, and security of the training and evaluation process. We consider that each stage serves a distinct function that cannot be combined without compromising the system's objectives. The detailed schemes will be discussed as follows:

\textbf{(1) Combination of DB and EB.} In the DB stage, both the trainers and model owners are committed to the process, preventing dishonest behaviors like the trainers stealing the model weights or the model owners failing to provide the model weights. If the DB miner and EB miner were the same entity, the model would be registered after passing through the DB process, but this could result in inconsistencies between the tested and the trained models. In this case, the separation of DB and EB ensures that the model's hash value is determined independently, preventing any tampering or unintended changes. On the other hand, the EB stage is initiated after DB to enable the selection of a new miner if needed, ensuring that the process remains decentralized and fair. Thus, if DB and EB were combined, it would be harder to replace a dishonest participant, leading to potential manipulation.

\textbf{(2) Combination of EB and TB.} EB and TB cannot be combined because the testing data cannot be released during the EB stage before the model's hash ($Hash(M^k_{n})$) is fixed. Allowing this would risk overfitting, as the model would have access to the test data before the final evaluation, which means that the model trainers can use the testing data to fine-tune their models so that they can have better evaluation scores. Moreover, the TB stage also prevents unfair testing data cases by separating the training and testing phases, ensuring that the testing data is unbiased, which is crucial for fair evaluation.

\textbf{(3) Combination of TB and SB.} If TB and SB were combined, it would be possible for a malicious participant to manipulate the testing data to artificially boost a specific model's ranking. To mitigate this, the PoW mechanism \cite{nakamoto2008bitcoin} randomly selects the TB miner, ensuring that no single entity can control the process in the long term. Moreover, the SB stage serves to prevent unfair performance testing, ensuring that the model’s evaluation is independent and accurate. Additionally, waiting until the model training is completed before the TB and SB phases helps avoid cheating, such as manipulating testing cases before evaluation. Therefore, the fact the same TB miner and SB miner could allow one entity to potentially manipulate both the test and score phases.

In summary, the four-stage process is essential for maintaining fairness, preventing dishonest behavior, and ensuring the integrity of the system.

\subsection{Consensus Mechanism}

As mentioned in Section \ref{sec_system}, we utilize the Proof of Work (PoW) \cite{nakamoto2008bitcoin} consensus model for block generation as an example by default. In our system, PoW is primarily used to maintain the randomness of miner selection, which helps prevent collusion and ensures the integrity of the consensus process. However, other consensus mechanisms capable of maintaining similar randomness can also be employed, while PoW is preferred due to its well-understood and established properties. The computational difficulty of PoW in this context can be adjusted by modifying the difficulty of solving the cryptographic puzzle \cite{cai2018decentralized}. In this case, the adjustment allows the system to control the time taken for block generation, ensuring consistency in the block creation process. It is worth noting that, a key distinction from Bitcoin's PoW is that, while Bitcoin's PoW is primarily used to control the time interval between blocks, the focus in DeRelayL is more on maintaining randomness in the selection of participants for training. This setup reduces the likelihood of malicious actors benefiting from any potential system collapse, ensuring that no participant gains at the expense of others.

On the other hand, the primary purpose of PoW in this system is to randomly select a miner, but ensuring the security of the system against attacks involves more than just this random selection, such as the 51\% attack \cite{ye2018analysis}. A 51\% attack would typically occur if a malicious entity controls the majority of the network's computational power and can alter the consensus \cite{ye2018analysis}. However, in DeRelayL, as long as the majority of miners are honest, even if a dishonest group successfully mines a block, their blocks will be discarded by the network. This is because the system relies on the assumption that the majority of participants will act honestly and that any block mined by dishonest individuals will eventually be rejected by the honest majority. Furthermore, the integrity of the system depends on the assumption that the honest majority will always outweigh any dishonest minority. If a dishonest majority were to emerge, they would likely be phased out over time as the system evolves, especially given that participants are incentivized to act honestly in order to receive rewards or maintain their reputation within the network. The decentralized and distributed nature of the blockchain, along with the transparent consensus mechanism, ensures that any attempt to subvert the system by dishonest actors is self-limiting, as their blocks would not be accepted by the majority of honest miners. In fact, to further strengthen the security issues, additional mechanisms, such as staking or reputation systems \cite{zhou2021blockchain}, could be introduced to discourage dishonest behavior and incentivize honest participation, ensuring that the network remains secure and trustworthy even in the event of potential attacks.

\subsection{Potential Real-World Applications}

Currently, large model training mainly relies on crawling corpora from the Internet (e.g., models like GPT-3 \cite{gpt3} and Llama 3.1 \cite{llama3}), while the data/knowledge contributors cannot share the profits of large models. This approach leads to a significant drawback: the model trainers find it extremely difficult to obtain data that is not available online, while the data contributors are not willing to contribute knowledge to the Internet. A real-world case has occurred that some artists have refused to share their artworks with model trainers \cite{duan2024incentive}, further highlighting the limitations of relying on Internet-based data collection. Therefore, DeRelayL's potential impact lies in creating a decentralized training paradigm. By allowing the exchange of model usage rights for data, DeRelayL intends to boost the richness of training data. In this case, a wider variety of data from different offline sources can be integrated into the training process, such as specialized industry datasets, personal diaries, and private research findings. Since this mechanism does not require a large amount of capital injection to purchase data, it reduces the economic cost of further enhancing the performance of large models. Moreover, the cost-effective approach also makes it more accessible for a broader range of participants, including small-scale research teams and individual developers, to contribute to and benefit from the development of large models. Therefore, the DeRelayL system can not only increase individual influence on the development of large models but also help prevent large companies from monopolizing the values embedded in large models.

\section{Limitations and Future Research Topics}

However, it is necessary to point out that the DeRelayL is still in its early stage, and there are several limitations that remain to be completely addressed in the future.

\subsection{Resource Overhead} \label{app_overhead}

In this paper, we employ fully homomorphic encryption (FHE) \cite{martins2017survey} to transmit model weights, where FHE is an encryption scheme that enables analytical functions to be run directly on encrypted data while yielding the same encrypted results as if the functions were run on plaintext data \cite{kim2023sharp}. The purpose of utilizing FHE is to evaluate the performance of trained models without exposing the model weights. However, our theoretical framework assumes an ideal scenario where the computational, memory, and storage overhead of FHE is acceptable. In practice, high computation and memory overhead make FHE computation over $10,000 \times$ slower than unencrypted computation on conventional computing systems that process unencrypted data \cite{kim2023sharp}. This substantial overhead also increases the cost of transmitting encrypted models, as it requires significantly more memory, storage, and internet traffic. Additionally, FHE computations may introduce small errors that accumulate over FHE operations performed, leading to approximate rather than precise results \cite{martins2017survey}. Consequently, FHE is less suitable for applications requiring high numerical precision, such as scientific computations, due to its reliance on polynomial approximations.

Therefore, to implement the DeRelayL system effectively, several challenges must be addressed: (1) The computational, memory, and storage overhead of FHE needs to be reduced, potentially through the development of more effective FHE algorithms. While specialized hardware can significantly accelerate FHE operations, it may not be accessible to common users, limiting their ability to participate in the DeRelayL system. (2) Due to the storage demands of FHE, broadcasting encrypted models poses a challenge for common users. A potential solution is that trainers can upload encrypted models to decentralized storage, and then broadcast only the storage addresses to other participants, avoiding the significant transmission costs associated with P2P communication. (3) The trade-off between FHE resource overhead and the precision required for performance evaluation should be further explored, since it is intuitive that reducing numerical precision could simplify FHE operations, thereby decreasing resource consumption. (4) Alternative methodologies for evaluating the training process should also be investigated. The motivation for using FHE is to prevent model weight leakage during performance evaluation. Thus, we consider that the appropriate utilization of technologies like zero-knowledge proof (ZKP) may also provide solutions to the proposed scenario of DeRelayL. ZKP is a cryptographic method that allows one party (i.e., the trainer) to prove to another party (i.e., the SBM) that they know a piece of information (i.e., the performance of a trained model) without revealing the actual information itself (e.g., the exact model weight) \cite{liu2021zkcnn, weng2021mystique}. Additionally, other cryptographic techniques that achieve similar objectives can also be considered to improve the DeRelayL system.

\subsection{Model Size Dilemma}

In this paper, we assume that the model size (or specific model architecture) is fixed at the creation of the genesis block. This assumption is reasonable, since a fixed model can standardize the programming interface for each participant, lowering the barrier to joining the DeRelayL system. However, according to scaling laws \cite{kaplan2020scaling, zhang2024scaling}, model size influences the upper-performance limit. Therefore, over many rounds of training, a fixed model size will eventually reach its performance ceiling, where further training yields diminishing returns. In this case, the DeRelayL system meets the model size dilemma, where model improvements are minimal no matter how to conduct the continued training, making the resource costs associated with training unworthy. Consequently, our mathematical modeling of DeRelayL does not consider a stop condition when the cost of model training exceeds the incremental value gained from the model improvement.

Additionally, it is necessary to establish mechanisms for dynamically enlarging the model size without initiating a new DeRelayL blockchain. Several critical challenges need to be addressed: determining who will be responsible for managing model size enlargement, identifying the appropriate timing/block/round for implementing the larger model, and defining the optimal size for the new model. Furthermore, the performance change associated with model enlargement must be carefully evaluated, and suitable technologies must be identified for transferring knowledge from the previous model to the new one. Correspondingly, the responsibility for executing this knowledge transfer and evaluating its performance must also be clearly defined. Therefore, the dynamic adjustment of model size remains an open research topic in the DeRelayL.

In practice, the model size dilemma is an extreme case that would only append after a significant number of training rounds, which would require a considerable amount of time to achieve. Theoretically, over such a long period, other DeRelayL blockchains with higher expected performance would emerge, attracting rational participants to migrate to these new blockchains. However, this transition would lead to another dilemma: a senior participant in the old DeRelayL blockchain would become a freshman in the new one, which makes the accumulated contribution of the senior participant in the old blockchain useless. Therefore, the model size dilemma of the DeRelayL paradigm and its associated challenges are worthy of further research and investigation.

\subsection{Training Time and Training Data Dilemma} \label{sec_training_time}

In the DeRelayL system, the training time for each round is dynamically adjusted to optimize training efficiency and performance. The duration is automatically set based on the minimum time required to achieve a positive increment in model performance. If the training time is too long, it may indicate that the training process is inefficient, leading to unnecessary resource consumption. On the contrary, if the duration is too short, there is a risk that the model's performance may not improve, resulting in the efforts of the trainers ineffective. To this end, the system should prioritize the final outcome, the real growth in performance, rather than focusing too heavily on the specifics of the training process itself. We consider that the ``valuable" data can be evaluated by whether it could quickly improve the model’s performance. For example, for large datasets, the system encourages trainers to break their datasets into smaller, more manageable parts, each of which can contribute to rapid performance gains. This approach may ensure that trainers with larger datasets can participate multiple times and continue to benefit from the model's improvements. Therefore, how to adjust the training duration according to performance increments and data value is an important challenge, which influences the fairness and efficiency of collaborative training in the DeRelayL system.

Besides the training itself, we also notice that the true value of utilized data may not be immediately reflected in the model's performance in the current round, which may discourage participants who possess high-quality data from contributing. Therefore, how to effectively reflect the contribution of the participants can be further studied in the future, e.g., by applying data valuation-related approaches \cite{shi2024data}.

\section{Conclusion}

In this paper, we propose a novel collaborative learning paradigm, named \underline{De}centralized \underline{Relay} \underline{L}earning (DeRelayL), a sustainable decentralized learning system where permissionless participants can contribute to model training in a relay-like manner and share the model together. We introduce the architecture and workflow of DeRelayL and incentive mechanisms to ensure sustainability. Moreover, theoretical analysis and numerical simulations are conducted to demonstrate the effectiveness of the proposed DeRelayL. At last, we provide comprehensive discussions of DeRelayL regarding promising research topics in the future. In summary, the proposed DeRelayL training mechanism aims to solve the challenge of motivating widespread participation in large-scale model training, especially by encouraging individuals to contribute data that is not readily available on the Internet. If this mechanism operates effectively, it could lead to a more equitable distribution of the benefits derived from AI, where participants actively influence and benefit from the intelligent big data era they help create. We expect that our insights can inspire related studies into decentralized collaborative learning systems that empower common users, fostering a fairer, more sustainable digital ecosystem, where data creators have greater control and can benefit from the AI models they help develop.

\bibliographystyle{IEEEtran}
\bibliography{reference}

@String{Computing = "Computing" }

@String{Computer = "{IEEE} Computer" }

@String{Springer = "Springer-Verlag" }

@ArtifactSoftware{R,
    title = {R: A Language and Environment for Statistical Computing},
    author = {{R Core Team}},
    organization = {R Foundation for Statistical Computing},
    address = {Vienna, Austria},
    year = {2019},
    url = {https://www.R-project.org/},
}

@article{yang2019federated,
  title={Federated machine learning: Concept and applications},
  author={Yang, Qiang and Liu, Yang and Chen, Tianjian and Tong, Yongxin},
  journal={ACM Transactions on Intelligent Systems and Technology (TIST)},
  volume={10},
  number={2},
  pages={1--19},
  year={2019},
  publisher={ACM New York, NY, USA}
}

@article{martins2017survey,
  title={A survey on fully homomorphic encryption: An engineering perspective},
  author={Martins, Paulo and Sousa, Leonel and Mariano, Artur},
  journal={ACM Computing Surveys (CSUR)},
  volume={50},
  number={6},
  pages={1--33},
  year={2017},
  publisher={ACM New York, NY, USA}
}

@article{nakamoto2008bitcoin,
  title={Bitcoin: A peer-to-peer electronic cash system},
  author={Nakamoto, Satoshi},
  journal={Satoshi Nakamoto},
  year={2008}
}

@incollection{friedman2018double,
  title={The double auction market institution: A survey},
  author={Friedman, Daniel},
  booktitle={The double auction market},
  pages={3--26},
  year={2018},
  publisher={Routledge}
}

@article{kaplan2020scaling,
  title={Scaling laws for neural language models},
  author={Kaplan, Jared and McCandlish, Sam and Henighan, Tom and Brown, Tom B and Chess, Benjamin and Child, Rewon and Gray, Scott and Radford, Alec and Wu, Jeffrey and Amodei, Dario},
  journal={arXiv preprint arXiv:2001.08361},
  year={2020}
}

@inproceedings{zhang2024scaling,
  title={When Scaling Meets LLM Finetuning: The Effect of Data, Model and Finetuning Method},
  author={Zhang, Biao and Liu, Zhongtao and Cherry, Colin and Firat, Orhan},
  booktitle={The Twelfth International Conference on Learning Representations},
  year={2024}
}

@article{fan2020hybrid,
  title={Hybrid blockchain-based resource trading system for federated learning in edge computing},
  author={Fan, Sizheng and Zhang, Hongbo and Zeng, Yuchen and Cai, Wei},
  journal={IEEE Internet of Things Journal},
  volume={8},
  number={4},
  pages={2252--2264},
  year={2020},
  publisher={IEEE}
}

@article{paris2014efficient,
  title={An efficient auction-based mechanism for mobile data offloading},
  author={Paris, Stefano and Martignon, Fabio and Filippini, Ilario and Chen, Lin},
  journal={IEEE Transactions on Mobile Computing},
  volume={14},
  number={8},
  pages={1573--1586},
  year={2014},
  publisher={IEEE}
}

@inproceedings{sun2022profit,
  title={A Profit-Maximizing Model Marketplace with Differentially Private Federated Learning},
  author={Sun, Peng and Chen, Xu and Liao, Guocheng and Huang, Jianwei},
  booktitle={IEEE INFOCOM 2022-IEEE Conference on Computer Communications},
  pages={1439--1448},
  year={2022},
  organization={IEEE}
}

@article{duan2024incentive,
  title={Incentive Mechanism Design Toward a Win--Win Situation for Generative Art Trainers and Artists},
  author={Duan, Haihan and El Saddik, Abdulmotaleb and Cai, Wei},
  journal={IEEE Transactions on Computational Social Systems},
  year={2024},
  publisher={IEEE}
}

@article{cai2018decentralized,
  title={Decentralized applications: The blockchain-empowered software system},
  author={Cai, Wei and Wang, Zehua and Ernst, Jason B and Hong, Zhen and Feng, Chen and Leung, Victor CM},
  journal={IEEE access},
  volume={6},
  pages={53019--53033},
  year={2018},
  publisher={IEEE}
}

@inproceedings{mcmahan2017communication,
  title={Communication-efficient learning of deep networks from decentralized data},
  author={McMahan, Brendan and Moore, Eider and Ramage, Daniel and Hampson, Seth and y Arcas, Blaise Aguera},
  booktitle={Artificial intelligence and statistics},
  pages={1273--1282},
  year={2017},
  organization={PMLR}
}

@inproceedings{hsieh2020non,
  title={The non-iid data quagmire of decentralized machine learning},
  author={Hsieh, Kevin and Phanishayee, Amar and Mutlu, Onur and Gibbons, Phillip},
  booktitle={International Conference on Machine Learning},
  pages={4387--4398},
  year={2020},
  organization={PMLR}
}

@inproceedings{karimireddy2020scaffold,
  title={Scaffold: Stochastic controlled averaging for federated learning},
  author={Karimireddy, Sai Praneeth and Kale, Satyen and Mohri, Mehryar and Reddi, Sashank and Stich, Sebastian and Suresh, Ananda Theertha},
  booktitle={International conference on machine learning},
  pages={5132--5143},
  year={2020},
  organization={PMLR}
}

@inproceedings{fraboni2021clustered,
  title={Clustered sampling: Low-variance and improved representativity for clients selection in federated learning},
  author={Fraboni, Yann and Vidal, Richard and Kameni, Laetitia and Lorenzi, Marco},
  booktitle={International Conference on Machine Learning},
  pages={3407--3416},
  year={2021},
  organization={PMLR}
}

@article{lin2020ensemble,
  title={Ensemble distillation for robust model fusion in federated learning},
  author={Lin, Tao and Kong, Lingjing and Stich, Sebastian U and Jaggi, Martin},
  journal={Advances in neural information processing systems},
  volume={33},
  pages={2351--2363},
  year={2020}
}

@article{li2020federated,
  title={Federated optimization in heterogeneous networks},
  author={Li, Tian and Sahu, Anit Kumar and Zaheer, Manzil and Sanjabi, Maziar and Talwalkar, Ameet and Smith, Virginia},
  journal={Proceedings of Machine learning and systems},
  volume={2},
  pages={429--450},
  year={2020}
}

@article{wang20federated,
  title={Federated Learning with Matched Averaging},
  author={Wang, Hongyi and Yurochkin, Mikhail and Sun, Yuekai and Papailiopoulos, Dimitris and Khazaeni, Yasaman},
  journal={International Conference on Learning Representations},
  year={2020}
}

@article{chen21fedbe,
  title={FedBE: Making Bayesian Model Ensemble Applicable to Federated Learning},
  author={Chen, Hong-You and Chao, Wei-Lun},
  journal={International Conference on Learning Representations},
  year={2021}
}

@article{beltran2023decentralized,
  title={Decentralized federated learning: Fundamentals, state of the art, frameworks, trends, and challenges},
  author={Beltr{\'a}n, Enrique Tom{\'a}s Mart{\'\i}nez and P{\'e}rez, Mario Quiles and S{\'a}nchez, Pedro Miguel S{\'a}nchez and Bernal, Sergio L{\'o}pez and Bovet, G{\'e}r{\^o}me and P{\'e}rez, Manuel Gil and P{\'e}rez, Gregorio Mart{\'\i}nez and Celdr{\'a}n, Alberto Huertas},
  journal={IEEE Communications Surveys \& Tutorials},
  year={2023},
  publisher={IEEE}
}

@article{he2018cola,
  title={Cola: Decentralized linear learning},
  author={He, Lie and Bian, An and Jaggi, Martin},
  journal={Advances in Neural Information Processing Systems},
  volume={31},
  year={2018}
}

@inproceedings{li2022learning,
  title={Learning to collaborate in decentralized learning of personalized models},
  author={Li, Shuangtong and Zhou, Tianyi and Tian, Xinmei and Tao, Dacheng},
  booktitle={Proceedings of the IEEE/CVF Conference on Computer Vision and Pattern Recognition},
  pages={9766--9775},
  year={2022}
}

@article{qu2022blockchain,
  title={Blockchain-enabled federated learning: A survey},
  author={Qu, Youyang and Uddin, Md Palash and Gan, Chenquan and Xiang, Yong and Gao, Longxiang and Yearwood, John},
  journal={ACM Computing Surveys},
  volume={55},
  number={4},
  pages={1--35},
  year={2022},
  publisher={ACM New York, NY}
}

@inproceedings{zhang2021incentive,
  title={Incentive mechanism for horizontal federated learning based on reputation and reverse auction},
  author={Zhang, Jingwen and Wu, Yuezhou and Pan, Rong},
  booktitle={Proceedings of the Web Conference 2021},
  pages={947--956},
  year={2021}
}

@inproceedings{qin2024blockdfl,
  title={BlockDFL: A Blockchain-based Fully Decentralized Peer-to-Peer Federated Learning Framework},
  author={Qin, Zhen and Yan, Xueqiang and Zhou, Mengchu and Deng, Shuiguang},
  booktitle={Proceedings of the ACM on Web Conference 2024},
  pages={2914--2925},
  year={2024}
}

@inproceedings{kim2019blockchain,
  title={Blockchain-based node-aware dynamic weighting methods for improving federated learning performance},
  author={Kim, You Jun and Hong, Choong Seon},
  booktitle={2019 20th Asia-pacific network operations and management symposium (APNOMS)},
  pages={1--4},
  year={2019},
  organization={IEEE}
}

@article{kim2019blockchained,
  title={Blockchained on-device federated learning},
  author={Kim, Hyesung and Park, Jihong and Bennis, Mehdi and Kim, Seong-Lyun},
  journal={IEEE Communications Letters},
  volume={24},
  number={6},
  pages={1279--1283},
  year={2019},
  publisher={IEEE}
}

@article{qu2021proof,
  title={Proof of federated learning: A novel energy-recycling consensus algorithm},
  author={Qu, Xidi and Wang, Shengling and Hu, Qin and Cheng, Xiuzhen},
  journal={IEEE Transactions on Parallel and Distributed Systems},
  volume={32},
  number={8},
  pages={2074--2085},
  year={2021},
  publisher={IEEE}
}

@inproceedings{chen2024page,
  title={PAGE: Equilibrate Personalization and Generalization in Federated Learning},
  author={Chen, Qian and Wang, Zilong and Hu, Jiaqi and Yan, Haonan and Zhou, Jianying and Lin, Xiaodong},
  booktitle={Proceedings of the ACM on Web Conference 2024},
  pages={2955--2964},
  year={2024}
}

@inproceedings{zhou2024accelerating,
  title={Accelerating the Decentralized Federated Learning via Manipulating Edges},
  author={Zhou, Mingyang and Liu, Gang and Lu, KeZhong and Mao, Rui and Liao, Hao},
  booktitle={Proceedings of the ACM on Web Conference 2024},
  pages={2945--2954},
  year={2024}
}

@inproceedings{wang2023fededge,
  title={Fededge: Accelerating edge-assisted federated learning},
  author={Wang, Kaibin and He, Qiang and Chen, Feifei and Jin, Hai and Yang, Yun},
  booktitle={Proceedings of the ACM Web Conference 2023},
  pages={2895--2904},
  year={2023}
}

@inproceedings{zhang2024privacy,
  title={Privacy-preserving and fairness-aware federated learning for critical infrastructure protection and resilience},
  author={Zhang, Yanjun and Sun, Ruoxi and Shen, Liyue and Bai, Guangdong and Xue, Minhui and Meng, Mark Huasong and Li, Xue and Ko, Ryan and Nepal, Surya},
  booktitle={Proceedings of the ACM on Web Conference 2024},
  pages={2986--2997},
  year={2024}
}

@article{li2020blockchain,
  title={A blockchain-based decentralized federated learning framework with committee consensus},
  author={Li, Yuzheng and Chen, Chuan and Liu, Nan and Huang, Huawei and Zheng, Zibin and Yan, Qiang},
  journal={IEEE Network},
  volume={35},
  number={1},
  pages={234--241},
  year={2020},
  publisher={IEEE}
}

@article{li2021blockchain,
  title={Blockchain assisted decentralized federated learning (BLADE-FL): Performance analysis and resource allocation},
  author={Li, Jun and Shao, Yumeng and Wei, Kang and Ding, Ming and Ma, Chuan and Shi, Long and Han, Zhu and Poor, H Vincent},
  journal={IEEE Transactions on Parallel and Distributed Systems},
  volume={33},
  number={10},
  pages={2401--2415},
  year={2021},
  publisher={IEEE}
}

@inproceedings{ramanan2020baffle,
  title={Baffle: Blockchain based aggregator free federated learning},
  author={Ramanan, Paritosh and Nakayama, Kiyoshi},
  booktitle={2020 IEEE international conference on blockchain (Blockchain)},
  pages={72--81},
  year={2020},
  organization={IEEE}
}

@inproceedings{ekuban2023towards,
  title={Towards decentralised learning analytics (positioning paper)},
  author={Ekuban, Audrey and Domingue, John},
  booktitle={Companion Proceedings of the ACM Web Conference 2023},
  pages={1435--1438},
  year={2023}
}

@article{diskin2021distributed,
  title={Distributed deep learning in open collaborations},
  author={Diskin, Michael and Bukhtiyarov, Alexey and Ryabinin, Max and Saulnier, Lucile and Sinitsin, Anton and Popov, Dmitry and Pyrkin, Dmitry V and Kashirin, Maxim and Borzunov, Alexander and Villanova del Moral, Albert and others},
  journal={Advances in Neural Information Processing Systems},
  volume={34},
  pages={7879--7897},
  year={2021}
}

@article{ryabinin2020towards,
  title={Towards crowdsourced training of large neural networks using decentralized mixture-of-experts},
  author={Ryabinin, Max and Gusev, Anton},
  journal={Advances in Neural Information Processing Systems},
  volume={33},
  pages={3659--3672},
  year={2020}
}

@inproceedings{buyukates2023proof,
  title={Proof-of-Contribution-Based Design for Collaborative Machine Learning on Blockchain},
  author={Buyukates, Baturalp and He, Chaoyang and Han, Shanshan and Fang, Zhiyong and Zhang, Yupeng and Long, Jieyi and Farahanchi, Ali and Avestimehr, Salman},
  booktitle={2023 IEEE International Conference on Decentralized Applications and Infrastructures (DAPPS)},
  pages={13--22},
  year={2023},
  organization={IEEE}
}

@inproceedings{ebrahimi2024blockchain,
  title={Blockchain-based Federated Learning Utilizing Zero-Knowledge Proofs for Verifiable Training and Aggregation},
  author={Ebrahimi, Elmira and Sober, Michael and Hoang, Anh-Tu and Ileri, Can Umut and Sanders, William and Schulte, Stefan},
  booktitle={2024 IEEE International Conference on Blockchain (Blockchain)},
  pages={54--63},
  year={2024},
  organization={IEEE}
}

@inproceedings{yazdaninejad2024blockchain,
  title={A Blockchain-enabled and Transparent Evaluation of ML Models in the Decentralised Marketplace},
  author={Yazdaninejad, Hamid and Rajarajan, Muttukrishnan and Krol, Michal},
  booktitle={2024 IEEE International Conference on Blockchain (Blockchain)},
  pages={458--463},
  year={2024},
  organization={IEEE}
}

@inproceedings{zheng2021cerebro,
  title={Cerebro: A platform for Multi-Party cryptographic collaborative learning},
  author={Zheng, Wenting and Deng, Ryan and Chen, Weikeng and Popa, Raluca Ada and Panda, Aurojit and Stoica, Ion},
  booktitle={30th USENIX Security Symposium (USENIX Security 21)},
  pages={2723--2740},
  year={2021}
}

@article{chang2024survey,
  title={A survey on evaluation of large language models},
  author={Chang, Yupeng and Wang, Xu and Wang, Jindong and Wu, Yuan and Yang, Linyi and Zhu, Kaijie and Chen, Hao and Yi, Xiaoyuan and Wang, Cunxiang and Wang, Yidong and others},
  journal={ACM Transactions on Intelligent Systems and Technology},
  volume={15},
  number={3},
  pages={1--45},
  year={2024},
  publisher={ACM New York, NY}
}

@article{gao2023gradientcoin,
  title={Gradientcoin: A peer-to-peer decentralized large language models},
  author={Gao, Yeqi and Song, Zhao and Yin, Junze},
  journal={arXiv preprint arXiv:2308.10502},
  year={2023}
}

@article{alami2024free,
  title={Free open source communities sustainability: Does it make a difference in software quality?},
  author={Alami, Adam and Pardo, Ra{\'u}l and Lin{\aa}ker, Johan},
  journal={Empirical Software Engineering},
  volume={29},
  number={5},
  pages={114},
  year={2024},
  publisher={Springer}
}

@article{shi2022energy,
  title={An energy-efficient and privacy-aware decomposition framework for edge-assisted federated learning},
  author={Shi, Yimin and Duan, Haihan and Yang, Lei and Cai, Wei},
  journal={ACM Transactions on Sensor Networks},
  volume={18},
  number={4},
  pages={1--24},
  year={2022},
  publisher={ACM New York, NY}
}

@inproceedings{kim2023sharp,
  title={SHARP: A short-word hierarchical accelerator for robust and practical fully homomorphic encryption},
  author={Kim, Jongmin and Kim, Sangpyo and Choi, Jaewan and Park, Jaiyoung and Kim, Donghwan and Ahn, Jung Ho},
  booktitle={Proceedings of the 50th Annual International Symposium on Computer Architecture},
  pages={1--15},
  year={2023}
}

@article{bo2023relay,
  title={Relay learning: a physically secure framework for clinical multi-site deep learning},
  author={Bo, Zi-Hao and Guo, Yuchen and Lyu, Jinhao and Liang, Hengrui and He, Jianxing and Deng, Shijie and Xu, Feng and Lou, Xin and Dai, Qionghai},
  journal={NPJ Digital Medicine},
  volume={6},
  number={1},
  pages={204},
  year={2023},
  publisher={Nature Publishing Group UK London}
}

@inproceedings{liu2021zkcnn,
  title={Zkcnn: Zero knowledge proofs for convolutional neural network predictions and accuracy},
  author={Liu, Tianyi and Xie, Xiang and Zhang, Yupeng},
  booktitle={Proceedings of the 2021 ACM SIGSAC Conference on Computer and Communications Security},
  pages={2968--2985},
  year={2021}
}

@inproceedings{weng2021mystique,
  title={Mystique: Efficient conversions for Zero-Knowledge proofs with applications to machine learning},
  author={Weng, Chenkai and Yang, Kang and Xie, Xiang and Katz, Jonathan and Wang, Xiao},
  booktitle={30th USENIX Security Symposium (USENIX Security 21)},
  pages={501--518},
  year={2021}
}

@inproceedings{ruan2024optimal,
  title={Optimal Power Control for Over-the-Air Federated Learning with Gradient Compression},
  author={Ruan, Mengzhe and Li, Yunhe and Zhang, Weizhou and Song, Linqi and Xu, Weitao},
  booktitle={2024 IEEE 30th International Conference on Parallel and Distributed Systems (ICPADS)},
  pages={326--333},
  year={2024},
  organization={IEEE}
}

@article{kang2024tiny,
  title={Tiny Multi-Agent DRL for Twins Migration in UAV Metaverses: A Multi-Leader Multi-Follower Stackelberg Game Approach},
  author={Kang, Jiawen and Zhong, Yue and Xu, Minrui and Nie, Jiangtian and Wen, Jinbo and Du, Hongyang and Ye, Dongdong and Huang, Xumin and Niyato, Dusit and Xie, Shengli},
  journal={IEEE Internet of Things Journal},
  year={2024},
  publisher={IEEE}
}

@article{cao2021toward,
  title={Toward on-device federated learning: A direct acyclic graph-based blockchain approach},
  author={Cao, Mingrui and Zhang, Long and Cao, Bin},
  journal={IEEE Transactions on Neural Networks and Learning Systems},
  volume={34},
  number={4},
  pages={2028--2042},
  year={2021},
  publisher={IEEE}
}

@article{gpt3,
  title={Language models are few-shot learners},
  author={Brown, Tom and Mann, Benjamin and Ryder, Nick and Subbiah, Melanie and Kaplan, Jared D and Dhariwal, Prafulla and Neelakantan, Arvind and Shyam, Pranav and Sastry, Girish and Askell, Amanda and others},
  journal={Advances in neural information processing systems},
  volume={33},
  pages={1877--1901},
  year={2020}
}

@article{llama3,
  title={The llama 3 herd of models},
  author={Dubey, Abhimanyu and Jauhri, Abhinav and Pandey, Abhinav and Kadian, Abhishek and Al-Dahle, Ahmad and Letman, Aiesha and Mathur, Akhil and Schelten, Alan and Yang, Amy and Fan, Angela and others},
  journal={arXiv preprint arXiv:2407.21783},
  year={2024}
}

@inproceedings{ye2018analysis,
  title={Analysis of security in blockchain: Case study in 51\%-attack detecting},
  author={Ye, Congcong and Li, Guoqiang and Cai, Hongming and Gu, Yonggen and Fukuda, Akira},
  booktitle={2018 5th International conference on dependable systems and their applications (DSA)},
  pages={15--24},
  year={2018},
  organization={IEEE}
}

@article{zhou2021blockchain,
  title={Blockchain-based decentralized reputation system in E-commerce environment},
  author={Zhou, Zhili and Wang, Meimin and Yang, Ching-Nung and Fu, Zhangjie and Sun, Xingming and Wu, QM Jonathan},
  journal={Future Generation Computer Systems},
  volume={124},
  pages={155--167},
  year={2021},
  publisher={Elsevier}
}

@inproceedings{shi2024data,
  title={Data Valuation and Pricing in Internet of Things: Survey and Vision},
  author={Shi, Xinyi and Duan, Haihan},
  booktitle={2024 IEEE International Conference on Smart Internet of Things (SmartIoT)},
  pages={547--554},
  year={2024},
  organization={IEEE}
}













\begin{IEEEbiography}[{\includegraphics[width=1in,height=1.25in,clip,keepaspectratio]{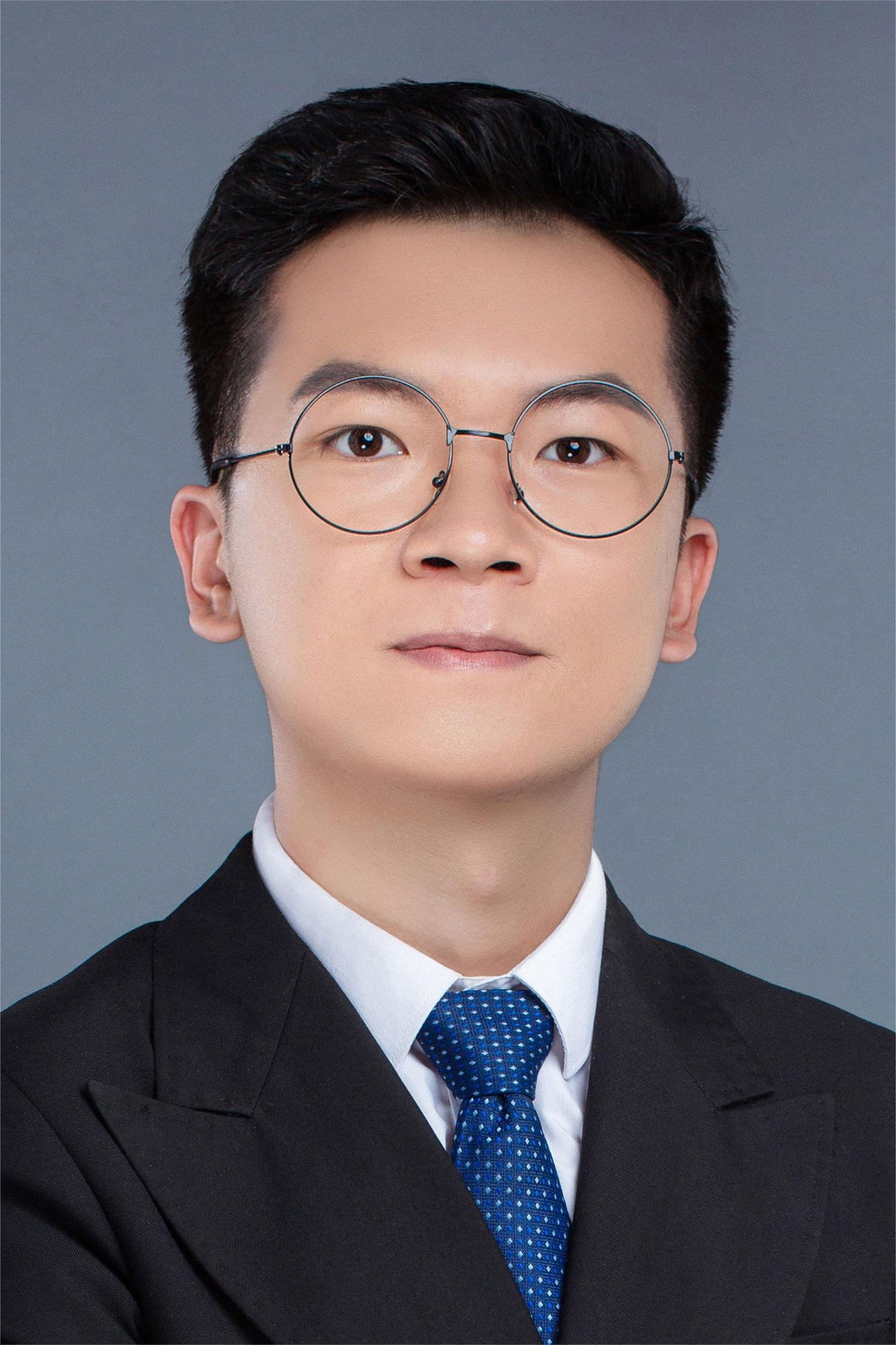}}]{Haihan Duan} (Member, IEEE) received his B.Eng. degree in Computer Science and Technology from East China Normal University, Shanghai, China, in 2017, and his M.Eng. degree in Software Engineering from Sichuan University, Chengdu, China, in 2020, and his Ph.D. degree in Computer and Information Engineering from The Chinese University of Hong Kong, Shenzhen, China, in 2023. He is currently an associate professor with Artificial Intelligence Research Institute, Shenzhen MSU-BIT University (SMBU), and also with Guangdong-Hong Kong-Macao Joint Laboratory for Emotion Intelligence and Pervasive Computing, Shenzhen, China. Before joining SMBU, he worked as a postdoctoral research fellow at University of Ottawa and Mohamed bin Zayed University of Artificial Intelligence (MBZUAI), located in Abu Dhabi, United Arab Emirates. His research interests include multimedia, blockchain and Web3, metaverse, human-centered computing, and medical image analysis.
\end{IEEEbiography}

\begin{IEEEbiography}[{\includegraphics[width=1in,height=1.25in,clip,keepaspectratio]{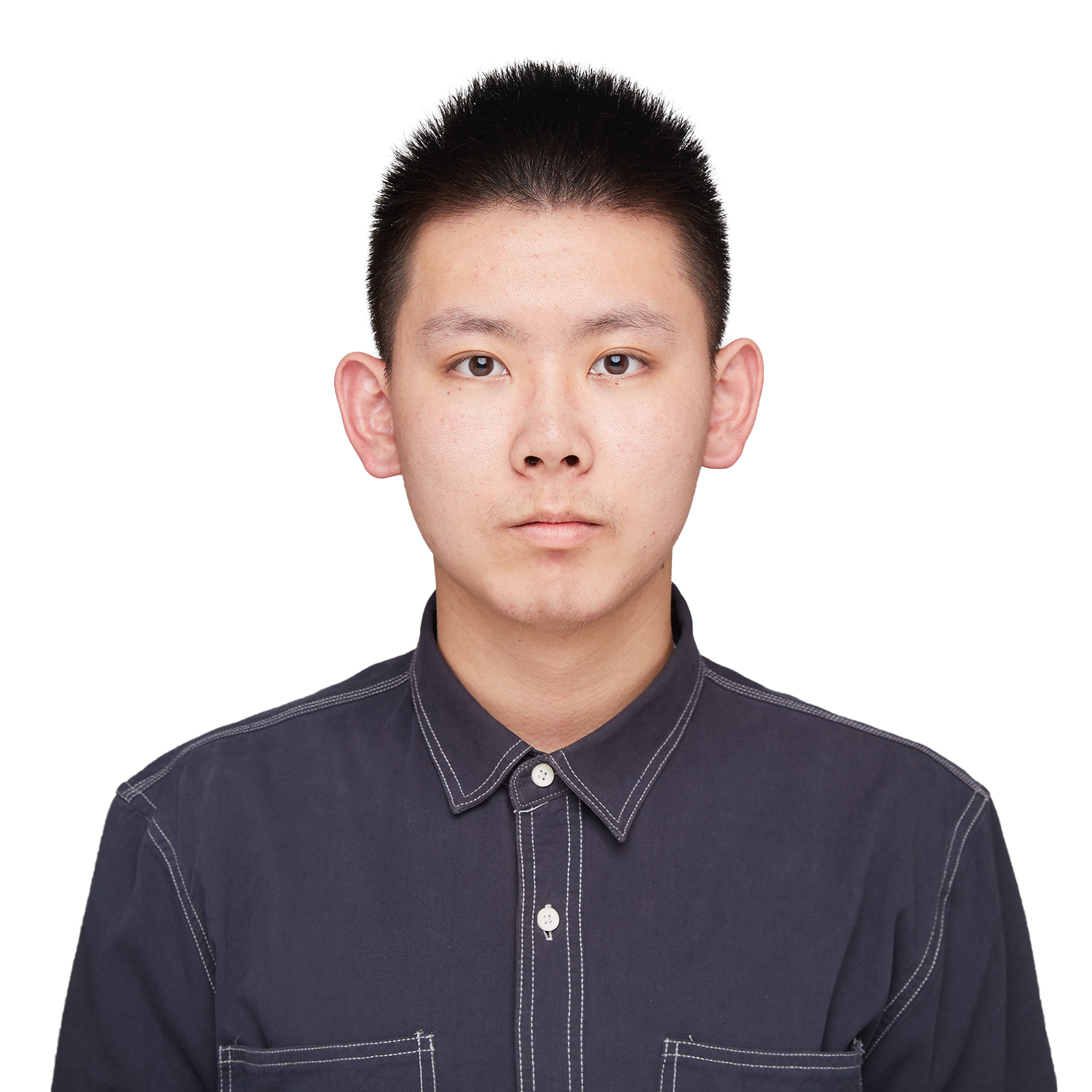}}]{Tengfei Ma} is a forth-year undergraduate student in the School of Science and Engineering at The Chinese University of Hong Kong, Shenzhen, China, pursuing his B.Eng. degree in Computer Engineering. He is currently a visiting student with the Artificial Intelligence Research Institute, Shenzhen MSU-BIT University, and also with the Guangdong-Hong Kong-Macao Joint Laboratory for Emotion Intelligence and Pervasive Computing, Shenzhen, China. His research interests include blockchain, retrieval augmented generation, and game theory.
\end{IEEEbiography}

\begin{IEEEbiography}[{\includegraphics[width=1in,height=1.25in,clip,keepaspectratio]{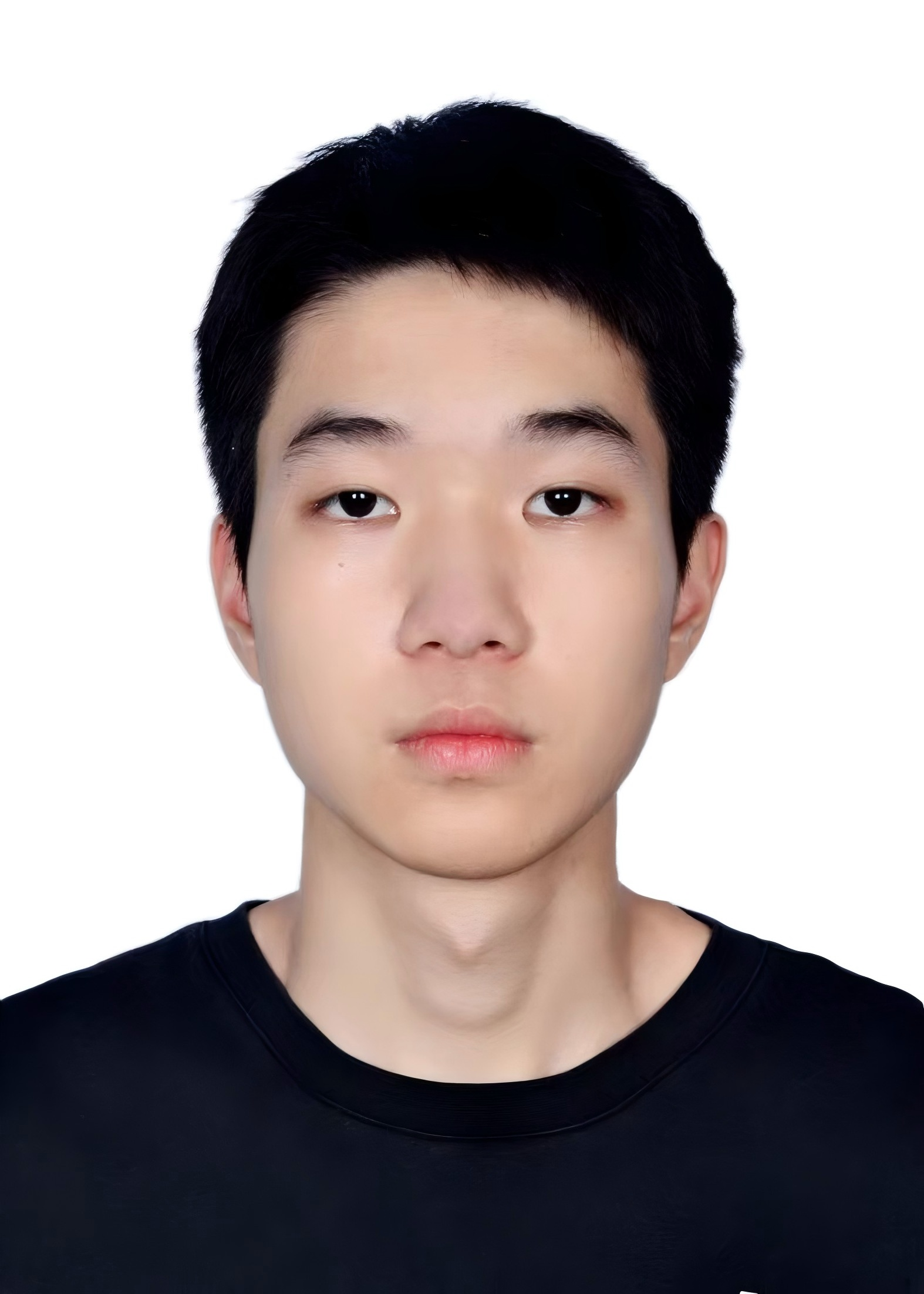}}]{Yuyang Qin} is a third-year undergraduate student in the School of Data Science at The Chinese University of Hong Kong, Shenzhen, China, pursuing his B.Sc. degree in Data Science. He is currently a visiting student with the Artificial Intelligence Research Institute, Shenzhen MSU-BIT University, and also with the Guangdong-Hong Kong-Macao Joint Laboratory for Emotion Intelligence and Pervasive Computing, Shenzhen, China. His research interests include blockchain development, data analysis, and blockchain system design.
\end{IEEEbiography}

\begin{IEEEbiography}[{\includegraphics[width=1in,height=1.25in,clip,keepaspectratio]{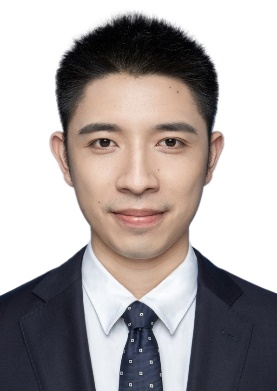}}]{Runhao Zeng} (Member, IEEE) received the PhD degree in software engineering from South China University of Technology, in 2021. He is currently an associate professor with Artificial Intelligence Research Institute, Shenzhen MSU-BIT University (SMBU), and also with Guangdong-Hong Kong-Macao Joint Laboratory for Emotion Intelligence and Pervasive Computing, Shenzhen, China. He has authored or coauthored several peer-reviewed papers on computer vision, machine learning on top-tier conferences and journals, including the Proceedings of NeurIPS, CVPR, ICCV, and TPAMI. His current research interests include machine learning, computer vision, with particular focus on video analysis.
\end{IEEEbiography}

\begin{IEEEbiography}[{\includegraphics[width=1in,height=1.25in,clip,keepaspectratio]{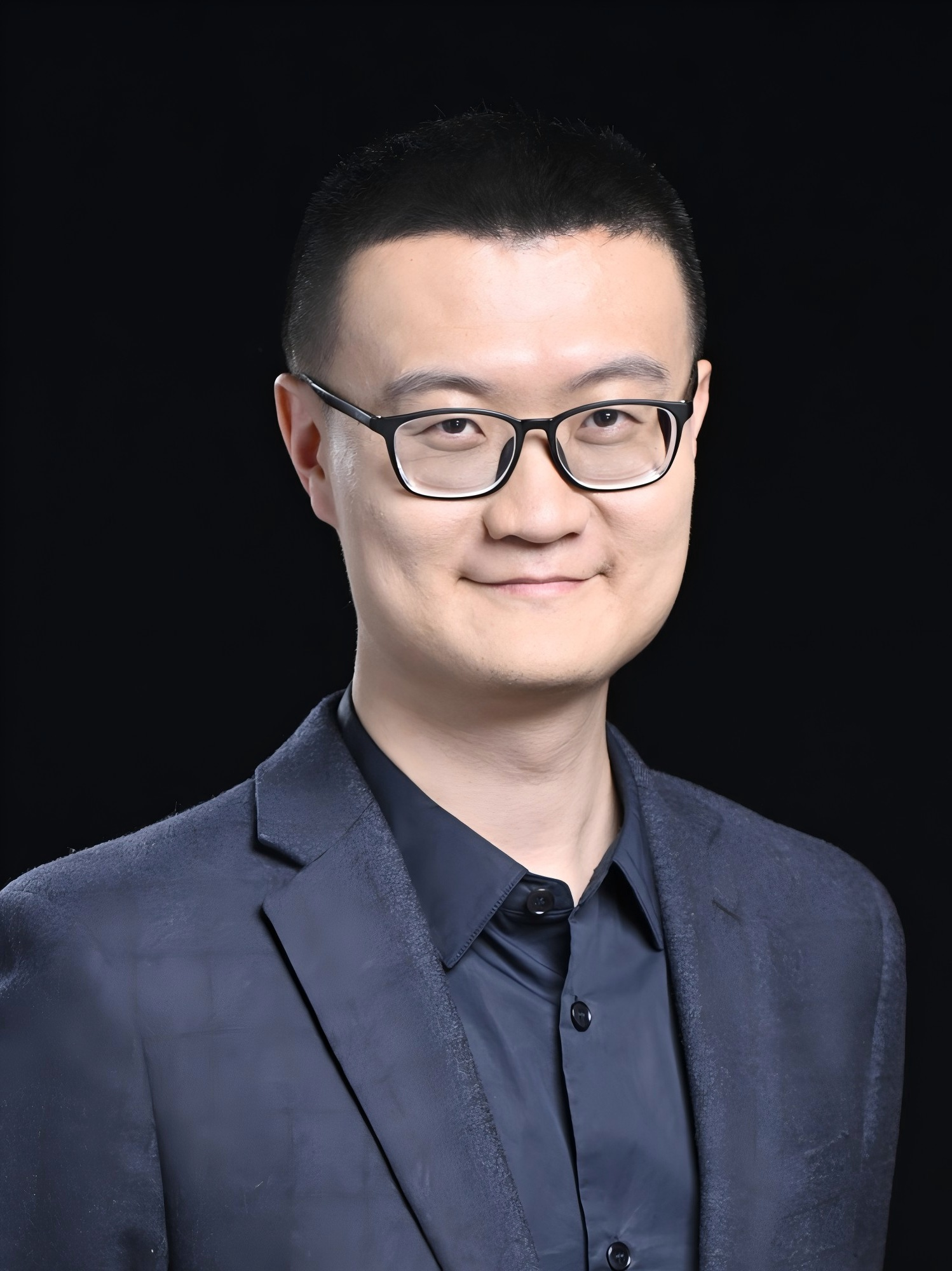}}]{Wei Cai} (Senior Member, IEEE) is a tenure-track Assistant Professor of Computer Science and Systems at the School of Engineering and Technology, University of Washington, Tacoma, WA, USA. He is now leading the Decentralized Computing Laboratory. He holds a Ph.D. in Electrical and Computer Engineering from The University of British Columbia (2016), an M.Sc. in Electrical Engineering and Computer Science from Seoul National University (2011), and a B.Eng. in Software Engineering from Xiamen University (2008). Prior to joining UW, Dr. Cai was an Assistant Professor of Electrical and Computer Engineering at The Chinese University of Hong Kong, Shenzhen, China. He has also conducted research visits at Academia Sinica (Taiwan), The Hong Kong Polytechnic University, and the National Institute of Informatics (Japan). Dr. Cai has co-authored over 100 peer-reviewed journal and conference papers and has received 6 Best Paper Awards. His research focuses on decentralized computing, with emphasis on mechanism design, social computing, multimedia, and applications. He serves as an Associate Editor for ACM Transactions on Multimedia Computing, Communications, and Applications (TOMM) and IEEE Transactions on Computational Social Systems (TCSS), and previously for IEEE Transactions on Cloud Computing (TCC). Dr. Cai is a Steering Committee member for ACM NOSSDAV, where he served as TPC co-chair in 2023, and has been an Area Chair for ACM MM since 2023. He is a Senior Member of IEEE and a member of ACM.
\end{IEEEbiography}

\begin{IEEEbiography}[{\includegraphics[width=1in,height=1.25in,clip,keepaspectratio]{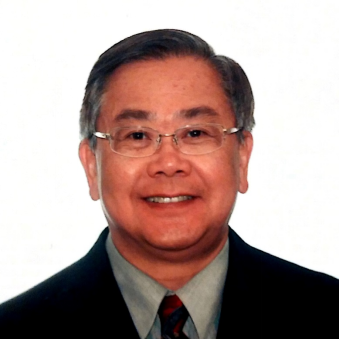}}]{Victor C.M. Leung} (Life Fellow, IEEE) is currently with Artificial Intelligence Research Institute, Shenzhen MSU-BIT University, Shenzhen, China. He is also an Emeritus Professor of electrical and computer engineering and the Director of the Laboratory for Wireless Networks and Mobile Systems, The University of British Columbia (UBC), Canada. His research interests include wireless networks and mobile systems. He has published widely in these areas. He is a Fellow of the Royal Society of Canada, the Canadian Academy of Engineering, and the Engineering Institute of Canada. He received the 1977 APEBC Gold Medal, the 1977–1981 NSERC Postgraduate Scholarships, the IEEE Vancouver Section Centennial Award, the 2011 UBC Killam Research Prize, the 2017 Canadian Award for Telecommunications Research, the 2018 IEEE TCGCC Distinguished Technical Achievement Recognition Award, and the 2018 ACM MSWiM Reginald Fessenden Award. He has coauthored papers that won the 2017 IEEE ComSoc Fred W. Ellersick Prize, the 2017 IEEE Systems Journal Best Paper Award, the 2018 IEEE CSIM Best Journal Paper Award, and the 2019 IEEE TCGCC Best Journal Paper Award. He has been serving on the editorial boards of the IEEE Transactions on Green Communications and Networking, IEEE Transactions on Cloud Computing, IEEE Access, IEEE Network, and several other journals. He is named in the current Clarivate Analytics list of ``Highly Cited Researchers." He is a fellow of the Royal Society of Canada (Academy of Science), Canadian Academy of Engineering, and Engineering Institute of Canada, and a life fellow of IEEE.
\end{IEEEbiography}

\begin{IEEEbiography}[{\includegraphics[width=1in,height=1.25in,clip,keepaspectratio]{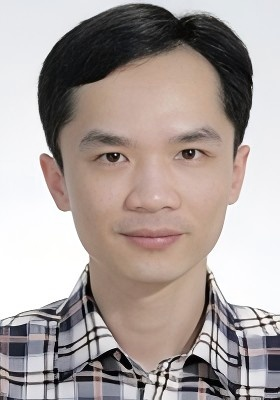}}]{Xiping Hu} (Member, IEEE) received the Ph.D.
degree from the University of British Columbia, Vancouver, BC, Canada. He is currently a
professor with Beijing Institute of Technology, and with Shenzhen MSU-BIT University, China.
He has more than 150 papers published and presented in prestigious conferences and journals, such as IEEE TPAMI/TMC/TPDS/TIP/JSAC, IEEE COMST, ACM MobiCom/MM/SIGIR/WWW, AAAI, and IJCAI. He has been serving as associate editor of IEEE TCSS, and the lead guest editors of IEEE IoT Journal and IEEE TASE etc. He has been granted several key national research projects as principal investigator. He was the Co-Founder and CTO of Bravolol Ltd., Hong Kong, a leading language learning mobile application company with over 100 million users, and listed as the top 2 language education platform globally. His research areas consist of mobile cyber-physical systems, crowd sensing and affective computing.
\end{IEEEbiography}

\clearpage

\appendix

\subsection{Utility Formulation} \label{app_utility}

This section presents the detailed derivation regarding the utilities of different participants' strategies, and the final formulations are listed in Table \ref{tab_utilities}. Since the formulation involves lots of parameters, we summarize the key annotations listed in Table \ref{tab_symbol} for better understanding. The utility formulation is following with Section \ref{sec_incentive}, and the extended derivation is shown as follows:

\textbf{(1) Model owner (MO):} The utility of MO can be denoted as:
\begin{align}
    \begin{split}
    U^{MO} &= R^{MO} - C^{MO} \\
          &= R_{\text{Now}}^{MO} + R_{\text{Future}}^{MO} - C_{\text{Deposit}}^{MO} - C_{\text{Transmit}}^{MO}
    \end{split}
\end{align}
where $R_{\text{Now}}^{MO}$ and $R_{\text{Future}}^{MO}$ refer to the 4) additional citation reward as discussed in \textbf{Step (11)} of Section \ref{sec_system}. $R_{\text{Now}}^{MO}$ is the citation reward of the current round, and $R_{\text{Future}}^{MO}$ is the revenue of the future rounds, which will be calculated as a geometric series since the future revenue has a discount rate. For the cost, MO has deposit cost $C_{\text{Deposit}}^{MO}$ for each round, but the cost is likely to be returned if the selected trainers behave normally. Moreover, MO also has transmission cost $C_{\text{Transmit}}^{MO}$ when sending the model to the selected trainers.

The strategy set of MO is \{Normal (N), Not Transmitting (NTm)\}, where ``Not Transmitting (NTm)" is an abstract presentation of dishonest behavior, including transmitting fake weights, etc. The utilities of different strategies are formulated as follows:

\textbf{(1.1) MO with N:}

The revenue of MO with N is given by:
\begin{align}
    \begin{split}
    R^{MO} &= R_{\text{Now}}^{MO} + R_{\text{Future}}^{MO} \\
          &= \overline{Q_{\text{Selected}}^{MO}} \cdot \mathcal{R}_{\text{Cited}} + \frac{\overline{Q_{\text{Selected}}^{MO}} \cdot \beta \cdot \mathcal{R}_{\text{Cited}}}{1 - \beta} \\
          &= \frac{\overline{Q_{\text{Selected}}^{MO}} \cdot \mathcal{R}_{\text{Cited}}}{1 - \beta}
    \end{split}
\end{align}
where $\overline{Q_{\text{Selected}}^{MO}}$ is the number of selected models of MO, and $\mathcal{R}_{\text{Cited}}$ is the revenue for citation reward. Kindly remind that we designed a mechanism to reward the citation of good models, discussed in \textbf{4) additional citation reward} of \textbf{Step (11)} in Section \ref{sec_system}, and it should have a discount rate $0 \le \beta \le 1$ when calculating the future revenue. In summary, the final formula of $R^{MO}$ is presented as the sum of geometric series, i.e., the total expected revenue for citation reward in now and future.

The cost of MO with N is given by:
\begin{align}
    \begin{split}
    C^{MO} &= C_{\text{Deposit}}^{MO} + C_{\text{Transmit}}^{MO} \\
           &= Q_{\text{Selected}} \cdot (1 - s) \cdot b^{MO} + k_{\text{Transmit}} \cdot |M|
    \end{split}
\end{align}
where $Q_{\text{Selected}}$ is number of selected trainers and $b^{MO}$ is MO's deposit cost for one trainer, determined by Algorithm \ref{alg_selection} in Section \ref{sec_negotiation}. And $0 \le s \le 1$ is the proportion of selected qualified models. Kindly remind that, to alleviate the lazy workers, we design a mechanism that only the trained models which have performance ranking in top-$\mathcal{K}$ can return the deposit cost for both original model owners in \textbf{3) returning qualified deposit} of \textbf{Step (11)} in Section \ref{sec_system}. Here, we use $0 \le s \le 1$ to denote the proportion that can pass the performance comparison for better generalization. Therefore, $Q_{\text{Selected}} \cdot (1 - s) \cdot b^{MO}$ is the expected deposit cost of MO. Moreover, $k_{\text{Transmit}}$ is the transmission cost coefficient, and $|M|$ is the number of parameters in the model, so $k_{\text{Transmit}} \cdot |M|$ can denote the transmission cost.

Therefore, the utility of MO with N can be formulated as:
\begin{align}
\begin{split}
    U_{N}^{MO} &= R^{MO} - C^{MO} \\
    &= \frac{\overline{Q_{\text{Selected}}^{MO}} \cdot \mathcal{R}_{\text{Cited}}}{1 - \beta}
    - Q_{\text{Selected}} \cdot (1 - s) \cdot b^{MO} \\ & \quad - k_{\text{Transmit}} \cdot |M|
\end{split}
\end{align}

\textbf{(1.2) MO with NTm:}

The revenue of MO with NTm is given by:
\begin{align}
    \begin{split}
    R^{MO}  &= 0
    \end{split}
\end{align}
which means T cannot obtain original models to train, so MO also cannot obtain any citation reward.

The cost of MO with NTm is given by:
\begin{align}
    \begin{split}
    C^{MO} = C_{\text{DepositLoss}}^{MO} = Q_{\text{Selected}} \cdot b^{MO}
    \end{split}
\end{align}
which means that the deposit of MO is $b^{MO}$ will lose, if the MO behave dishonestly. Moreover, since the MO have selected $Q_{\text{Selected}}$ Ts, the deposit loss will be $Q_{\text{Selected}} \cdot b^{MO}$.

Therefore, the utility of MO with NTm can be formulated as:
\begin{align}
\begin{split}
    U_{NTm}^{MO} &= - Q_{\text{Selected}} \cdot b^{MO}
\end{split}
\end{align}

\textbf{(2) Trainer (T):} The utility of T can be represented as:
\begin{align}
    \begin{split}
    U^{T} &= R^{T} - C^{T} = R_{\text{Now}}^T + R_{\text{Future}}^T  \\
          & - C_{\text{Train}} - C_{\text{Deposit}}^{T} - C_{\text{RecM}}^{T} - C_{\text{Encrypt}} - C_{\text{Broadcast}}
    \end{split}
\end{align}
where $R_{\text{Now}}^T$ is the revenue of the current round, which contains the revenue of the received model from MO $R_{\text{RecM}}^T$ and the revenue of the model trained by T $R_{\text{TrainedM}}^T$. And $R_{\text{Future}}^T$ is the future revenue for additional citation reward, similar to MO. The cost of T consists of five parts: 1) model training cost $C_{Train}$; 2) deposit cost $C_{\text{Deposit}}^{T}$, which will be returned if behaving normally; 3) cost of receiving the model from MO $C_{\text{RecM}}^{T}$; 4) FHE encryption cost of trained model $C_{\text{Encrypt}}$; 5) cost of broadcasting encrypted model $C_{\text{Broadcast}}$.

The T may choose to not train the model (NTr) or not broadcast the trained model (NBr), so the strategy set of T is \{Normal (N), Not Training (NTr), Not Broadcasting (NBr)\}. The utilities of different strategies are formulated as follows:

\textbf{(2.1) T with N:}

The revenue of T with N is given by:
\begin{align}
    \begin{split}
    R^{T} &= R_{\text{Now}}^T + R_{\text{Future}}^T \\
          &= (R_{\text{RecM}} + R_{\text{TrainedM}}) +  {R}_{\text{Cited}}^T \\
          &= [(V_{\text{RecM}} - V_{\text{Now}}^{T}) \cdot \circled{C} + 1 \cdot \circled{C}] + \frac{\overline{Q_{\text{Selected}}^T} \cdot \beta \cdot \mathcal{R}_{Cited}}{1 - \beta} \\
          &= \left(V_{\text{RecM}} - V_{\text{Now}}^{T} + 1  \right) \cdot \circled{C} + \frac{\overline{Q_{\text{Selected}}^T} \cdot \beta \cdot \mathcal{R}_{Cited}}{1 - \beta}
    \end{split}
\end{align}
where $V_{\text{RecM}} - V_{\text{Now}}^{T}$ denotes the version gap between the received model (the latest model) and the model that T has already owned (the out-of-date model), and we introduce \circled{C} to denote the value of knowledge gap between two adjacent model versions. The second term is the additional citation reward, similar to the MO, which is presented as the sum of geometric series based on the discount rate $0 \le \beta \le 1$.

The cost of T with N is given by:
\begin{equation}
    \begin{split}
    C^{T} &= C_{\text{Train}} + C_{\text{Deposit}}^{T} + C_{\text{RecM}}^{T} + C_{\text{Encrypt}} + C_{\text{Broadcast}} \\
          &= P_{\text{Comp}} \cdot D \cdot \tau \cdot |M| + (1 - s) \cdot {b}^{T} \\
          &\quad + k_{\text{Transmit}} \cdot |M| + k_{\text{Encrypt}} \cdot |M| \\
          &\quad + Q_{\text{Broadcast}} \cdot k_{\text{Transmit}} \cdot k_{\text{Expand}} \cdot |M|
    \end{split}
\end{equation}
where $P_{\text{Comp}}$ is the price of computational resource, $D$ represents the training data size, $\tau$ is the training duration, and $|M|$ denotes the number of parameters in the model, so the first term is the training cost. Then, $0 \le s \le 1$ denotes the proportion that can pass the performance comparison, thus $(1 - s) \cdot {b}^{T}$ is the expected deposit cost. Moreover, $k_{\text{Transmit}}$ is the transmission cost coefficient, $k_{\text{Encrypt}}$ is the FHE encryption cost coefficient, so the transmission cost and encryption cost can be presented. At last, $Q_{\text{Broadcast}}$ is the number of transmissions in broadcasting, $k_{\text{Expand}}$ is the expanding coefficient of model parameter number after FHE, and the aforementioned parameters can calculate the broadcasting cost. 

Therefore, the utility of T with N can be formulated as:
\begin{align}
\begin{split}
    U_{N}^{T} &= R^{T} - C^{T} \\
    &= \left( V_{\text{RecM}} - V_{\text{Now}}^{T} + 1 \right) \cdot \circled{C} + \frac{\overline{Q_{\text{Selected}}^T} \cdot \beta \cdot \mathcal{R}_{\text{Cited}}}{1 - \beta} \\
    &\quad - \left[ P_{\text{Comp}} \cdot D \cdot \tau \cdot |M| + (1 - s) \cdot {b}^{T} + k_{\text{Transmit}} \cdot |M| \right. \\
    &\quad \left. + k_{\text{Encrypt}} \cdot |M| + Q_{\text{Broadcast}} \cdot k_{\text{Transmit}} \cdot k_{\text{Expand}} \cdot |M| \right]
\end{split}
\end{align}

\textbf{(2.2) T with NTr:}

The revenue of T with NTr is given by:
\begin{align}
    \begin{split}
    R^{T}
          &= R_{\text{RecM}} \\
          &= (V_{\text{RecM}} - V_{\text{Now}}^{T}) \cdot \circled{C} 
    \end{split}
\end{align}
where $(V_{\text{RecM}} - V_{\text{Now}}^{T}) \cdot \circled{C}$ is the value of the version gap between the received model (the latest model) and the model that T has already owned (the out-of-date model).

The cost of T with NTr is given by:
\begin{equation}
    \begin{split}
    C^{T} &= C_{\text{DepositLost}}^{T} + C_{\text{RecM}}^{T} \\
          &=  {b}^{T} + k_{\text{Transmit}} \cdot |M|
    \end{split}
\end{equation}
where ${b}^{T}$ is the deposit of T, which will be lost since the T has failed to train. Moreover, the T has the transmission cost $k_{\text{Transmit}} \cdot |M|$ for receiving the model from MO.

Therefore, the utility of T with NTr can be formulated as:
\begin{align}
\begin{split}
    U_{NTr}^{T} &= R^{T} - C^{T} \\
    &= (V_{\text{RecM}} - V_{\text{Now}}^{T}) \cdot \circled{C} - {b}^{T} - k_{\text{Transmit}} \cdot |M|
\end{split}
\end{align}

\textbf{(2.3) T with NBr:}

The revenue of T with NBr is given by:
\begin{align}
    \begin{split}
    R^{T} 
          &= R_{\text{RecM}} + R_{\text{TrainedM}} \\
          &= (V_{\text{RecM}} - V_{\text{Now}}^{T}) \cdot \circled{C} + 1 \cdot \circled{C}\\
          &= \left(V_{\text{RecM}} - V_{\text{Now}}^{T} + 1  \right) \cdot \circled{C}
    \end{split}
\end{align}
where the Ts have finished the model training, so they will have one more $\circled{C}$ compared with ``Not Training". 

The cost of T with NBr is given by:
\begin{equation}
    \begin{split}
    C^{T} &= C_{\text{Train}} + C_{\text{DepositLost}}^{T} + C_{\text{RecM}}^{T}  \\
          &= P_{\text{Comp}} \cdot D \cdot \tau \cdot |M| + {b}^{T} + k_{\text{Transmit}} \cdot |M|
    \end{split}
\end{equation}
which has the training cost and transmission cost for receiving the model from MO, and the deposit of T ${b}^{T}$ will also be lost.

Therefore, the utility of T with NBr can be formulated as:
\begin{align}
\begin{split}
    U_{NBr}^{T} &= R^{T} - C^{T} \\
    &= \left( V_{\text{RecM}} - V_{\text{Now}}^{T} + 1 \right) \cdot \circled{C} \\
    &\quad - \left[ P_{\text{Comp}} \cdot D \cdot \tau \cdot |M| + {b}^{T} + k_{\text{Transmit}} \cdot |M| \right]
\end{split}
\end{align}

\textbf{(3) Deposit block miner (DBM):} The utility of DBM is:
\begin{align}
    \begin{split}
    U^{DBM} & = R^{DBM} - C^{DBM} \\
          & = R^{DBM}_{\text{Include}} - C_{\text{Mine}}
    \end{split}
\end{align}
where $R^{DBM}_{\text{Include}}$ is the incentive of miners to include deposit smart contracts as much as possible, so the revenue is proportional to the quantity of included data. $C_{\text{Mine}}$ is the cost of mining the block, i.e., the computational cost of the PoW consensus model. Note that, all miners (DBM, EBM, TBM, SBM) have the aforementioned $R_{\text{Include}}$ and $C_{\text{Mine}}$. We also simply assume the block generation intervals are almost identical for all stages, thus the $C_{\text{Mine}}$ is almost fixed. 

The DBM may choose to pack partial deposit smart contracts (NPA) or pack improper ones (PI), thus the strategy set of DBM is \{Normal (N), Not Packing All (NPA), Packing Improper Deposit Contracts (PI)\}. The utilities of different strategies are formulated as follows:

\textbf{(3.1) DBM with N:}

The revenue of DBM with N is given by:
\begin{align}
    \begin{split}
    R^{DBM} &= R^{DBM}_{\text{Include}}\\
    &= Q_{\text{Deposit}} \cdot \mathcal{R}_{\text{Deposit}} 
    \end{split}
\end{align}
where $Q_{\text{Deposit}}$ is the number of deposit smart contracts, and $\mathcal{R}_{\text{Deposit}}$ is the revenue to incentivize the DBM to pack deposit smart contracts as much as possible. Under this setting, a problem has emerged: what if a dishonest DBM packs invalid smart contracts to cheat for additional revenue? Due to the decentralized consensus of blockchain, other miners will check the validity of packed content, so the dishonest DBM will be recognized as ``Packing Improper Deposit Contracts". In this case, other miners will build a new fork to invalidate the block from a dishonest DBM, thus the corresponding dishonest DBM will lose all revenues.

The cost of DBM with N is given by:
\begin{align}
\begin{split}
    C^{DBM} &= C_{\text{Mine}}
\end{split}
\end{align}
where $C_{\text{Mine}}$ is the mining cost. Remind that all miners (DBM, EBM, TBM, SBM) have the  $C_{\text{Mine}}$, and we simply assume the block generation intervals are almost identical for all stages, thus the $C_{\text{Mine}}$ is almost fixed to all miners. 

Therefore, the utility of DBM with N can be formulated as:
\begin{align}
\begin{split}
    U^{DBM}_{N} &= R^{DBM} - C^{DBM}\\
                &= Q_{\text{Deposit}} \cdot \mathcal{R}_{\text{Deposit}}  - C_{\text{Mine}}
\end{split}
\end{align}

\textbf{(3.2) DBM with NPA:}

The revenue of DBM with NPA is given by:
\begin{align}
    \begin{split}
    R^{DBM} &= R^{DBM}_{\text{Include}}\\
    &= Q_{\text{DepositLess}} \cdot \mathcal{R}_{\text{Deposit}} 
    \end{split}
\end{align}
where $0 < Q_{\text{DepositLess}} < Q_{\text{Deposit}}$, which means the DBM has not packed all deposit smart contracts that broadcast in the DeRelayL network, and $Q_{\text{DepositLess}} \cdot \mathcal{R}_{\text{Deposit}}$ is the including revenue. 

The cost of DBM with NPA is given by:
\begin{align}
\begin{split}
    C^{DBM} &= C_{\text{Mine}}
\end{split}
\end{align}
where $C_{\text{Mine}}$ is the mining cost.

Therefore, the utility of DBM with NPA can be formulated as:
\begin{align}
\begin{split}
    U^{DBM}_{NPA} &= R^{DBM} - C^{DBM}\\
                &= Q_{\text{DepositLess}} \cdot \mathcal{R}_{\text{Deposit}}  - C_{\text{Mine}}
\end{split}
\end{align}

\textbf{(3.3) DBM with PI:}

The revenue of DBM with PI is given by:
\begin{align}
    \begin{split}
    R^{DBM} &= 0
    \end{split}
\end{align}
where the dishonest DBM will not obtain revenue since the cheating behavior will be recognized by other miners in the DeRelayL system. 

The cost of DBM with PI is given by:
\begin{align}
\begin{split}
    C^{DBM} &= C_{\text{Mine}}
\end{split}
\end{align}
where $C_{\text{Mine}}$ is the mining cost.

Therefore, the utility of DBM with PI is given by:
\begin{align}
\begin{split}
    U^{DBM}_{\text{PI}} &= -C_{\text{Mine}}
\end{split}
\end{align}

\textbf{(4) Encryption block miner (EBM):} The utility of EBM is:
\begin{align}
    \begin{split}
    U^{EBM} & = R^{EBM} - C^{EBM}\\
          & = R^{EBM}_{\text{Include}} + R_{\text{FHEM}} - C_{\text{Mine}} - C^{EBM}_{\text{RecFHEM}} - C_{\text{GenFHEKey}}
    \end{split}
\end{align}
where $R^{EBM}_{\text{Include}}$ is the incentive of including trained models' information, containing metadata and hash values. Since the EBM is responsible for generating the FHE key pair, the EBM can use the private key to decrypt encrypted models, as discussed in \textbf{Step (7)} of Section \ref{sec_system}. Thus, $R_{\text{FHEM}}$ is the revenue for decrypting encrypted models of FHE, and $C^{EBM}_{\text{RecFHEM}}$ is the cost for receiving the encrypted model (EB can just receive the model with best performance). $C_{\text{Mine}}$ is the mining cost, and $C_{\text{GenFHEKey}}$ is the cost of generating FHE key pair.

The EBM may not generate an FHE key (NG) or send a random number to disturb the training system. Thus, the strategy set of EBM is \{Normal (N), Not Generating FHE Key (NG)\}. The utilities of different strategies are formulated as follows:

\textbf{(4.1) EBM with N:}

The revenue of EBM with N is given by:
\begin{align}
\begin{split}
    R^{EBM} &= R^{EBM}_{\text{Include}} + R_{\text{FHEM}} \\
           &= Q_{\text{HashM}} \cdot \mathcal{R}_{\text{HashM}} + (V_{\text{FHEM}} - V_{\text{Now}}^{EBM}) \cdot \circled{C}
\end{split}
\end{align}
where $Q_{\text{HashM}}$ is the number of packed hash values of trained models, and $\mathcal{R}_{\text{HashM}}$ is the corresponding revenue. The second term is a special revenue for the EBM, as discussed in \textbf{Step (7)} of Section \ref{sec_system}, where $(V_{\text{FHEM}} - V_{\text{Now}}^{EBM}) \cdot \circled{C}$ is the value of the version gap between the encrypted model (there are lots of encrypted models are broadcast to the network, but the EBM will tend to decrypt the model that ranks at top-1 during the performance evaluation) and the model that the EBM has already owned (out-of-date model).

The cost of EBM with N is given by:
\begin{align}
\begin{split}
    C^{EBM} &= C_{\text{Mine}} + C_{\text{RecFHEM}}^{EBM} + C_{\text{GenFHEKey}} \\
    &= C_{\text{Mine}} + k_{\text{Transmit}} \cdot k_{\text{Expand}} \cdot |M| + C_{\text{GenFHEKey}}
\end{split}
\end{align}
where $C_{\text{Mine}}$ is the mining cost. If the EBMs want to obtain the latest model, they will afford the transmission cost for receiving the encrypted model, denoting as $k_{\text{Transmit}} \cdot k_{\text{Expand}} \cdot |M|$, where $k_{\text{Transmit}}$ is the transmission cost coefficient, $k_{\text{Expand}}$ is the expanding coefficient of model parameter number after FHE, and $|M|$ denotes the number of parameters in the model. 

Therefore, the utility of EBM with N can be formulated as:
\begin{align}
\begin{split}
    U^{EBM}_{N} &= R^{EBM} - C^{EBM}\\
    &= \left( Q_{\text{HashM}} \cdot \mathcal{R}_{\text{HashM}} + (V_{\text{FHEM}} - V_{\text{Now}}^{EBM}) \cdot \circled{C} \right) \\
    &\quad - \left( C_{\text{Mine}} + k_{\text{Transmit}} \cdot k_{\text{Expand}} \cdot |M| + C_{\text{GenFHEKey}} \right)
\end{split}
\end{align}

\textbf{(4.2) EBM with NG:}

The revenue of EBM with NG is given by:
\begin{align}
\begin{split}
    R^{EBM} &= 0
\end{split}
\end{align}
where the revenue of dishonest EBM will be equal to zero. This is because the invalid FHE key pair (e.g., randomly generated numbers) will be recognized by other participants, and rational participants will build a new fork to invalidate the block from a dishonest EBM.

The cost of EBM with NG is given by:
\begin{align}
\begin{split}
    C^{EBM} &= C_{\text{Mine}}
\end{split}
\end{align}
where $C_{\text{Mine}}$ is the mining cost.

Therefore, the utility of EBM with NG can be formulated as:
\begin{align}
\begin{split}
    U^{EBM}_{NG} &=  -C_{\text{Mine}}
\end{split}
\end{align}

\textbf{(5) Testing block miner (TBM):} The utility of TBM is:
\begin{align}
    \begin{split}
    U^{TBM} & = R^{TBM} - C^{TBM} \\
          & = R^{TBM}_{\text{Include}} + R_{\text{GenTDCases}} - C_{\text{Mine}} - C_{\text{GenTDCases}} 
    \end{split}
\end{align}
where $R^{TBM}_{\text{Include}}$ is the incentive of including information of encrypted models using FHE, containing metadata and hash values. The TBMs are responsible for generating testing data, so they will be rewarded $R_{\text{GenTDCases}}$ according to the number of testing cases. Thus, there are corresponding costs of generating testing cases $C_{\text{GenTDCases}}$. Similar to other miners, $C_{\text{Mine}}$ is the mining cost. 

For TBMs, they may upload improper testing cases (IT), thus the strategy set of TBM is \{Normal (N), Improper Testing Cases (IT)\}. The utilities of different strategies are formulated as follows:

\textbf{(5.1) TBM with N:}

The revenue of TBM with N is given by:
\begin{align}
\begin{split}
    R^{TBM} &= R^{TBM}_{\text{Include}} + R_{\text{GenTDCases}} \\
            &=  Q_{\text{EncryptedM}} \cdot \mathcal{R}_{\text{EncryptedM}} + Q_{\text{Cases}} \cdot \mathcal{R}_{\text{Case}}
\end{split}
\end{align}
where $Q_{\text{EncryptedM}}$ is the number of packed hash values of encrypted trained models after FHE, and $\mathcal{R}_{\text{EncryptedM}}$ is the corresponding incentive for including the hash values. Moreover, $Q_{\text{Cases}}$ is the number of included testing data cases, and $\mathcal{R}_{\text{Case}}$ is the corresponding incentive for generating the testing data cases.

The cost of TBM with N is given by:
\begin{align}
\begin{split}
    C^{TBM} &= C_{\text{Mine}} + C_{\text{GenTDCases}}\\
            &= C_{\text{Mine}} + Q_{Cases} \cdot C_{{\text{GenTDCase}}}^{\text{Unit}}
\end{split}
\end{align}
where $C_{\text{Mine}}$ is the mining cost. Moreover, the generation, preparation, or collection of testing data also has cost, and we use $C_{{\text{GenTDCase}}}^{\text{Unit}}$ to denote the generation cost of the testing data per unit/case, thus $Q_{Cases} \cdot C_{{\text{GenTDCase}}}^{\text{Unit}}$ is the total cost for generating the testing data cases.

Therefore, the utility of TBM with N can be formulated as:
\begin{align}
\begin{split}
    U^{TBM}_{N} &= R^{TBM} - C^{TBM}\\
    &= Q_{\text{EncryptedM}} \cdot \mathcal{R}_{\text{EncryptedM}} + Q_{\text{Cases}} \cdot \mathcal{R}_{\text{Case}} \\
    &\quad - C_{\text{Mine}} - Q_{Cases} \cdot C_{{\text{GenTDCase}}}^{\text{Unit}}
\end{split}
\end{align}

\textbf{(5.2) TBM with IT:}

The revenue of TBM with IT is given by:
\begin{align}
\begin{split}
    R^{TBM} &= 0
\end{split}
\end{align}
where the revenue of dishonest TBM will be equal to zero. This is because a dishonest TBM will be recognized by most participants since the testing data is accessible to the public that every participant can check, thus the rational participants will build a new fork to invalidate the block from a dishonest TBM.

The cost of TBM with IT is given by:
\begin{align}
\begin{split}
    C^{TBM} &= C_{\text{Mine}}
\end{split}
\end{align}
where the dishonest TBMs only have the mining cost $C_{\text{Mine}}$, since they will not truly prepare the testing data or just utilize old testing data uploaded by other TBMs in the previous training rounds.

Therefore, the utility of TBM with IT can be formulated as:
\begin{align}
\begin{split}
    U^{TBM}_{IT} &= -C_{\text{Mine}}
\end{split}
\end{align}

\textbf{(6) Settlement block miner (SBM):} The utility of SBM is:
\begin{align}
    \begin{split}
    U^{SBM} & = R^{SBM} - C^{SBM} \\
          & = R^{SBM}_{\text{Include}} + R_{\text{Verify}} - C_{\text{Mine}} - C^{SBM}_{\text{RecFHEMs}} - C_{\text{Verify}} 
    \end{split}
\end{align}
where $R^{SBM}_{\text{Include}}$ is the incentive of including verification confirmation details, containing metadata and performance index. The SBMs are responsible for verifying the performance of trained models, so they will be rewarded $R_{\text{Verify}}$ according to the number of verified models, corresponding to the cost for receiving all encrypted models $C^{SBM}_{\text{RecFHEMs}}$ and verifying them $C_{\text{Verify}}$. $C_{\text{Mine}}$ is the mining cost.

The SBM may not rank the trained models properly (IRa), so the strategy set of SBM is \{Normal (N), Improper Rank (IRa)\}. The utilities of different strategies are formulated as follows:

\textbf{(6.1) SBM with N:}

The revenue of SBM with N is given by:
\begin{align}
\begin{split}
    R^{SBM} &= R_{\text{Include}}^{\text{SBM}} + R_{\text{Verify}} \\
    &= Q_{\text{VerifiedM}} \cdot \mathcal{R}_{\text{VerifiedM}} + Q_{\text{VerifiedM}} \cdot Q_{\text{Cases}} \cdot \mathcal{R}_{\text{Verify}}
\end{split}
\end{align}
where $Q_{\text{VerifiedM}}$ is the number of verified models, and $\mathcal{R}_{\text{VerifiedM}}$ is the corresponding incentive for including the records. Moreover, the system will incentivize the SBM to verify trained models based on the testing data provided by the previous TBM. Thus, there is a revenue for verification, where $Q_{\text{Cases}}$ is the number of testing data cases, and $\mathcal{R}_{\text{Verify}}$ is the corresponding revenue for each case of verification per model.

The cost of SBM with N is given by:
\begin{align}
\begin{split}
    C^{SBM} &= C_{\text{Mine}} + C_{\text{RecFHEMs}}^{SBM} + C_{\text{Verify}} \\
            &= C_{\text{Mine}} +  Q_{\text{VerifiedM}} \cdot k_{\text{Transmit}} \cdot k_{\text{Expand}} \cdot |M| \\
            &\quad + Q_{\text{VerifiedM}} \cdot Q_{\text{Cases}} \cdot C_{\text{Verify}}^{\text{Unit}}
\end{split}
\end{align}
where $C_{\text{Mine}}$ is the mining cost. To verify the trained model, there is a receiving cost of encrypted models for the SBM, where $k_{\text{Transmit}}$ is the transmission cost coefficient, $k_{\text{Expand}}$ is the expanding coefficient of model parameter number after FHE, and $|M|$ denotes the number of parameters in the model. Moreover, the verification process has a computational cost, where $C_{\text{Verify}}^{\text{Unit}}$ denotes the verification cost per testing data case/unit.

Therefore, the utility of SBM with N can be formulated as:
\begin{align}
\begin{split}
    U^{SBM}_{N} &= R^{SBM} - C^{SBM}\\
           &= Q_{\text{VerifiedM}} \cdot \mathcal{R}_{\text{VerifiedM}} + Q_{\text{VerifiedM}} \cdot Q_{\text{Cases}} \cdot \mathcal{R}_{\text{Verify}}\\
           &\quad - C_{\text{Mine}} -  Q_{\text{VerifiedM}} \cdot k_{\text{Transmit}} \cdot k_{\text{Expand}} \cdot |M| \\
           &\quad - Q_{\text{VerifiedM}} \cdot Q_{\text{Cases}} \cdot C_{\text{Verify}}^{\text{Unit}}
\end{split}
\end{align}

\textbf{(6.2) SBM with IRa:}

The revenue of SBM with IRa is given by:
\begin{align}
    R^{SBM} &= 0
\end{align}
where the revenue of dishonest SBM will be equal to zero. This is because a dishonest SBM will be recognized by most participants, since everyone can check the correctness of the ranking results. Therefore, rational participants will build a new fork to invalidate the block from a dishonest SBM.

The cost of SBM with IRa is given by:
\begin{align}
    C^{SBM} &= C_{\text{Mine}}
\end{align}
where the dishonest TBMs only have the mining cost $C_{\text{Mine}}$, since they will not truly receive or verify the trained models. If the TBM has honestly finished the verification process, there is no reason that a rational TBM improperly ranks the models due to the huge cost of the verification process.

Therefore, the utility of SBM with IRa can be calculated as:
\begin{align}
\begin{split}
    U^{SBM}_{IRa} &= R^{SBM} - C^{SBM}\\
           &= 0 - C_{\text{Mine}}\\
           &= -C_{\text{Mine}}
\end{split}
\end{align}

Overall, all utilities of different participants’ strategies were formulated. Specifically, the final formulations of each strategy are summarized in Table \ref{tab_utilities} of Section \ref{sec_incentive}.

\subsection{Theoretical Analysis} \label{app_icir}

\subsubsection{Individual Rationality (IR)}

To achieve IR in the DeRelayL system, all participants that choose the ``Normal" strategy should at least have positive utilities, which means that the incentive provided by the proposed mechanism should lead to $U^{Participant}_{N} \ge 0$. Therefore, for each participant, there will be some conditions to guarantee that $U^{Participant}_{N} \ge 0$, which can be presented as follows:

\textbf{(1) IR for MO:}

Let $U_{N}^{MO} \ge 0$, the IR condition for MO can be formulated as:
\begin{align}
\begin{split}
    \frac{\overline{Q_{\text{Selected}}^{MO}} \cdot \mathcal{R}_{\text{Cited}}}{1 - \beta} 
    &- Q_{\text{Selected}} \cdot (1 - s) \cdot b^{MO} - k_{\text{Transmit}} \cdot |M| \ge 0
\end{split}
\end{align}
Rearranging the inequality, the condition can be formulated as:
\begin{align}
\begin{split}
    \frac{\overline{Q_{\text{Selected}}^{MO}} \cdot \mathcal{R}_{\text{Cited}}}{1 - \beta} 
    & \ge Q_{\text{Selected}} \cdot (1 - s) \cdot b^{MO} + k_{\text{Transmit}} \cdot |M|
\end{split}
\end{align}
Multiplying both sides by $(1 - \beta)$ to eliminate the denominator:
\begin{align}
\begin{split}
    \overline{Q_{\text{Selected}}^{MO}} \cdot \mathcal{R}_{\text{Cited}} 
    & \ge \\ (1 - \beta) \cdot & \left( Q_{\text{Selected}} \cdot (1 - s) \cdot b^{MO} + k_{\text{Transmit}} \cdot |M| \right)
\end{split}
\end{align}
Thus, to hold IR for MO, $\mathcal{R}_{\text{Cited}}$ should satisfy:
\begin{align}
    \mathcal{R}_{\text{Cited}} \ge \frac{(1 - \beta) \cdot \left( Q_{\text{Selected}} \cdot (1 - s) \cdot b^{MO} + k_{\text{Transmit}} \cdot |M| \right)}{\overline{Q_{\text{Selected}}^{MO}}}
\end{align}

\textbf{(2) IR for T:}

Let $U_{N}^{T} \ge 0$, the IR condition for T can be formulated as:
\begin{align}
    \begin{split}
    \left( V_{\text{RecM}} - V_{\text{Now}}^{T} + 1 \right) & \cdot \circled{C} + \frac{\overline{Q_{\text{Selected}}^T} \cdot \beta \cdot \mathcal{R}_{\text{Cited}}}{1 - \beta} \\
    & \ge P_{\text{Comp}} \cdot D \cdot \tau \cdot |M| + (1 - s) \cdot b^{T} \\
    &\quad + k_{\text{Transmit}} \cdot |M| + k_{\text{Encrypt}} \cdot |M| \\
    &\quad + Q_{\text{Broadcast}} \cdot k_{\text{Transmit}} \cdot k_{\text{Expand}} \cdot |M|
    \end{split}
\end{align}
Rearranging terms, the condition can be formulated as:
\begin{align}
    \begin{split}
    \frac{\overline{Q_{\text{Selected}}^T} \cdot \beta \cdot \mathcal{R}_{\text{Cited}}}{1 - \beta} & \ge P_{\text{Comp}} \cdot D \cdot \tau \cdot |M| + (1 - s) \cdot b^{T} \\
    &\quad + k_{\text{Transmit}} \cdot |M| + k_{\text{Encrypt}} \cdot |M| \\
    &\quad + Q_{\text{Broadcast}} \cdot k_{\text{Transmit}} \cdot k_{\text{Expand}} \cdot |M| \\
    &\quad - \left( V_{\text{RecM}} - V_{\text{Now}}^{T} + 1 \right) \cdot \circled{C}
    \end{split}
\end{align}
Multiplying by $(1 - \beta)$, the condition can be formulated as:
\begin{align}
    \begin{split}
    \overline{Q_{\text{Selected}}^T} & \cdot \beta \cdot \mathcal{R}_{\text{Cited}} \\ & \ge (1 - \beta) \cdot \left( P_{\text{Comp}} \cdot D \cdot \tau \cdot |M| + (1 - s) \cdot b^{T} \right. \\
    &\quad + k_{\text{Transmit}} \cdot |M| + k_{\text{Encrypt}} \cdot |M| \\
    &\quad + Q_{\text{Broadcast}} \cdot k_{\text{Transmit}} \cdot k_{\text{Expand}} \cdot |M| \\
    &\quad - \left( V_{\text{RecM}} - V_{\text{Now}}^{T} + 1 \right) \cdot \circled{C} \left) \right.
    \end{split}
\end{align}
Thus, to hold IR for T, $\mathcal{R}_{\text{Cited}}$ should satisfy:
\begin{align}
    \mathcal{R}_{\text{Cited}} & \ge \frac{(1 - \beta)}{\overline{Q_{\text{Selected}}^T} \cdot \beta} \cdot \Big( P_{\text{Comp}} \cdot D \cdot \tau \cdot |M| \notag \\
    &\quad + (1 - s) \cdot b^{T} + k_{\text{Transmit}} \cdot |M| + k_{\text{Encrypt}} \cdot |M| \notag \\
    &\quad + Q_{\text{Broadcast}} \cdot k_{\text{Transmit}} \cdot k_{\text{Expand}} \cdot |M| \notag \\
    &\quad - \left( V_{\text{RecM}} - V_{\text{Now}}^{T} + 1 \right) \cdot \circled{C} \Big)
\end{align}

\textbf{(3) IR for DBM:}

Let $U^{DBM}_{N} \ge 0$, the IR condition for DBM can be formulated as:
\begin{align}
    \begin{split}
        U^{DBM}_{N} &= Q_{\text{Deposit}} \cdot \mathcal{R}_{\text{Deposit}}  - C_{Mine} \ge 0
    \end{split}
\end{align}
Therefore, to hold IR for DBM, $\mathcal{R}_{\text{Deposit}}$ should satisfy:
\begin{align}
    \mathcal{R}_{\text{Deposit}}  & \ge \frac{C_{\text{Mine}}}{Q_{\text{Deposit}}}
\end{align}

\textbf{(4) IR for EBM:}

Let $U^{EBM}_{N} \ge 0$, the IR condition for EBM can be formulated as:
\begin{align}
    \begin{split}
        Q_{\text{HashM}} \cdot \mathcal{R}_{\text{HashM}} + (V_{\text{FHEM}} - V_{\text{Now}}^{EBM}) \cdot \circled{C} & \ge C_{\text{Mine}}\\
        + k_{\text{Transmit}} \cdot k_{\text{Expand}} \cdot |M| & + C_{\text{GenFHEKey}}
    \end{split}
\end{align}
Rearranging the terms, the condition can be formulated as:
\begin{align}
    \begin{split}
        Q_{\text{HashM}} \cdot \mathcal{R}_{\text{HashM}} & \ge C_{\text{Mine}} + k_{\text{Transmit}} \cdot k_{\text{Expand}} \cdot |M| \\
        &\quad + C_{\text{GenFHEKey}} - (V_{\text{FHEM}} - V_{\text{Now}}^{EBM}) \cdot \circled{C}
    \end{split}
\end{align}
Therefore, to hold IR for DBM, $\mathcal{R}_{\text{HashM}}$ should satisfy:
\begin{align}
    \mathcal{R}_{\text{HashM}} & \ge \frac{1}{Q_{\text{HashM}}} \cdot \Big( C_{\text{Mine}} + k_{\text{Transmit}} \cdot k_{\text{Expand}} \cdot |M| \notag \\
    &\quad + C_{\text{GenFHEKey}} - (V_{\text{FHEM}} - V_{\text{Now}}^{EBM}) \cdot \circled{C} \Big)
\end{align}

\textbf{(5) IR for TBM:}

Let $U^{TBM}_{N} \ge 0$, the IR condition for TBM can be formulated as:
\begin{align}
\begin{split}
    Q_{\text{EncryptedM}} \cdot \mathcal{R}_{\text{EncryptedM}} + Q_{\text{Cases}} \cdot \mathcal{R}_{\text{Case}} \\
    > C_{\text{Mine}} + Q_{Cases} \cdot C_{{\text{GenTDCase}}}^{\text{Unit}}
\end{split}
\end{align}
where there are two parameters that should be determined to hold IR for TBM, including $\mathcal{R}_{\text{EncryptedM}}$ and $\mathcal{R}_{\text{Case}}$.

\textbf{(6) IR for SBM:}

Let $U^{SBM}_{N} \ge 0$, the IR condition for SBM can be formulated as:
\begin{align}
    Q_{\text{VerifiedM}} & \cdot \mathcal{R}_{\text{VerifiedM}} + Q_{\text{VerifiedM}} \cdot Q_{\text{Cases}} \cdot \mathcal{R}_{\text{Verify}} \notag \\
    & \ge C_{\text{Mine}} + Q_{\text{VerifiedM}} \cdot k_{\text{Transmit}} \cdot k_{\text{Expand}} \cdot |M| \notag \\
    &\quad + Q_{\text{VerifiedM}} \cdot Q_{\text{Cases}} \cdot C_{\text{Verify}}^{\text{Unit}}
\end{align}
where there are two parameters that should be determined to hold IR for SBM, including $\mathcal{R}_{\text{VerifiedM}}$ and $\mathcal{R}_{\text{Verify}}$.

\subsubsection{Incentive Compatibility (IC)}

To achieve IC in the DeRelayL system, all rational participants will tend to choose the ``Normal" strategy, which means that the utility of the ``Normal" strategy should be greater than other strategies. Therefore, for each participant, the incentive provided by the proposed mechanism should lead to $U^{Participant}_{N} \ge U^{Participant}_{Other Strategy}$, which can be formulated as follows:

\textbf{(1) IC for MO:}

The strategy set of MO is \{Normal (N), Not Transmitting (NTm)\}. Therefore, we will compare the utility of MO with N and MO with NTm:

\begin{align}
\begin{split}
    U_{N}^{MO} - U_{NTm}^{MO} &= U_{N}^{MO} + Q_{\text{Selected}} \cdot b^{MO} > 0
\end{split}
\end{align}
where $U_{N}^{MO} \ge 0$ due to the IR for EBM, and the revenue for including records $Q_{\text{Selected}} \cdot b^{MO} > 0$. Therefore, the utility of MO with N is greater than MO with NTm, ensuring IC for MO.

\textbf{(2) IC for T:}

The strategy set of T is \{Normal (N), Not Training (NTr), Not Broadcasting (NBr)\}. Therefore, we will first compare the utilities of T with N and T with NTr:

\begin{align}
    \begin{split}
        U_{N}^{T} - U_{NTr}^{T} &= U_{N}^{T} - (V_{\text{RecM}} - V_{\text{Now}}^{T}) \cdot \circled{C}\\
        & \quad + {b}^{T} + k_{\text{Transmit}} \cdot |M|  \\
        &> -(V_{\text{RecM}} - V_{\text{Now}}^{T}) \cdot \circled{C} + {b}^{T} \\
        &> 0
    \end{split}
\end{align}
where $(V_{\text{RecM}} - V_{\text{Now}}^{T}) \cdot \circled{C}$ is the value of the version gap between the received model (the latest model) and the model that T has already owned (the out-of-date model). The formula ${b}^{T} -(V_{\text{RecM}} - V_{\text{Now}}^{T}) \cdot \circled{C} > 0$ means that the deposit of T should not be lower than the value of model (i.e., an effective deposit discussed in Section \ref{sec_system}). Otherwise, T will not have the motivation to train the model and just cheat for the latest models by depositing a small amount of coins. Therefore, the mechanism should have a condition:
\begin{align}
\begin{split}
    {b}^{T} > (V_{\text{RecM}} - V_{\text{Now}}^{T}) \cdot \circled{C}
    \label{formula_IC_1}
\end{split}
\end{align}

Then, we will compare the utilities of T with N and T with NBr:
\begin{align}
    \begin{split}
        U_{N}^{T} & - U_{NBr}^{T} = \\ 
        & \left( \left( V_{\text{RecM}} - V_{\text{Now}}^{T} + 1 \right) \cdot \circled{C} 
        + \frac{\overline{Q_{\text{Selected}}^T} \cdot \beta \cdot \mathcal{R}_{\text{Cited}}}{1 - \beta} \right) \\
        &\quad - \left( P_{\text{Comp}} \cdot D \cdot \tau \cdot |M| + (1 - s) \cdot {b}^{T} + k_{\text{Transmit}} \cdot |M| \right. \\
        &\quad \left. + k_{\text{Encrypt}} \cdot |M| + Q_{\text{Broadcast}} \cdot k_{\text{Transmit}} \cdot k_{\text{Expand}} \cdot |M| \right) \\
        &\quad - \left( \left( V_{\text{RecM}} - V_{\text{Now}}^{T} + 1 \right) \cdot \circled{C} \right) \\
        &\quad + \left( P_{\text{Comp}} \cdot D \cdot \tau \cdot |M| + {b}^{T} + k_{\text{Transmit}} \cdot |M| \right)
    \end{split}
\end{align}
Eliminating the common terms, the formula can be presented as:
\begin{align}
    \begin{split}
        U_{N}^{T} - U_{NBr}^{T} &= \frac{\overline{Q_{\text{Selected}}^T} \cdot \beta \cdot \mathcal{R}_{\text{Cited}}}{1 - \beta} \\
        &\quad - \left( (- s) \cdot {b}^{T} + k_{\text{Encrypt}} \cdot |M| \right) \\
        &\quad - \left( Q_{\text{Broadcast}} \cdot k_{\text{Transmit}} \cdot k_{\text{Expand}} \cdot |M| \right)
    \end{split}
\end{align}
Thus, to ensure \( U_{N}^{T} - U_{NBr}^{T} > 0 \), the mechanism should satisfy:
\begin{align}
    \frac{\overline{Q_{\text{Selected}}^T} \cdot \beta \cdot \mathcal{R}_{\text{Cited}}}{1 - \beta} 
    &> (- s) \cdot {b}^{T} + k_{\text{Encrypt}} \cdot |M| \notag \\
    &\quad + Q_{\text{Broadcast}} \cdot k_{\text{Transmit}} \cdot k_{\text{Expand}} \cdot |M|
\end{align}
where the condition can be formulated as:
\begin{align}
    \mathcal{R}_{\text{Cited}} & > \frac{1}{\overline{Q_{\text{Selected}}^T} \cdot \beta} 
    \Big( (- s) \cdot {b}^{T} + k_{\text{Encrypt}} \cdot |M| \notag \\
    &\quad + Q_{\text{Broadcast}} \cdot k_{\text{Transmit}} \cdot k_{\text{Expand}} \cdot |M| \Big) \cdot (1 - \beta)
    \label{formula_IC_2}
\end{align}
Therefore, there are two conditions (Formula (\ref{formula_IC_1}) and Formula (\ref{formula_IC_2})) that should be satisfied in the mechanism design to ensure IC for T.

\textbf{(3) IC for DBM:}

The strategy set of DBM is \{Normal (N), Not Packing All (NPA), Packing Improper Deposit Contracts (PI)\}. Therefore, we will first compare the utilities of DBM with N and DBM with NPA:
\begin{align}
    \begin{split}
        U^{DBM}_{N} - U^{DBM}_{NPA} &= \left( Q_{\text{Deposit}} \cdot \mathcal{R}_{\text{Deposit}} - C_{\text{Mine}} \right) \\
        &\quad - \left( Q_{\text{DepositLess}} \cdot \mathcal{R}_{\text{Deposit}} - C_{\text{Mine}} \right)
    \end{split}
\end{align}
The expression can be simplified as:
\begin{align}
    \begin{split}
        U^{DBM}_{N} - U^{DBM}_{NPA} &= \left( Q_{\text{Deposit}} - Q_{\text{DepositLess}} \right) \cdot \mathcal{R}_{\text{Deposit}}
    \end{split}
\end{align}
Since \( Q_{\text{Deposit}} > Q_{\text{DepositLess}} \), we can know:
\begin{align}
    \begin{split}
        U^{DBM}_{N} - U^{DBM}_{NPA} > 0
    \end{split}
\end{align}
Therefore, the utility of DBM with N is greater than DBM with NPA. On the other hand, we will first compare the utilities of DBM with N and DBM with PI:
\begin{align}
    \begin{split}
        U^{DBM}_{N} - U^{DBM}_{\text{PI}} &= \left( Q_{\text{Deposit}} \cdot \mathcal{R}_{\text{Deposit}} - C_{\text{Mine}} \right) \\
        &\quad - \left( -C_{\text{Mine}} \right)
    \end{split}
\end{align}
The expression can be simplified as:
\begin{align}
    \begin{split}
        U^{DBM}_{N} - U^{DBM}_{\text{PI}} = Q_{\text{Deposit}} \cdot \mathcal{R}_{\text{Deposit}} > 0
    \end{split}
\end{align}
Therefore, the utility of DBM with N is greater than DBM with PI. Overall, the IC for DBM can be ensured.

\textbf{(4) IC for EBM:}

The strategy set of EBM is \{Normal (N), Not Generating FHE Key (NG)\}. Therefore, we will compare the utilities of EBM with N and EBM with NG:
\begin{align}
\begin{split}
    U^{EBM}_{N} - U^{EBM}_{NG} = U^{EBM}_{N} + C_{\text{Mine}} > 0
\end{split}
\end{align}
where $U^{EBM}_{N} \ge 0$ due to the IR for EBM, and mining cost $C_{\text{Mine}} > 0$, so the $U^{EBM}_{N} - U^{EBM}_{NG} > 0$. Therefore, the utility of EBM with N is greater than EBM with NG, ensuring IC for EBM.

\textbf{(5) IC for TBM:}

The strategy set of TBM is \{Normal (N), Improper Testing Cases (IT)\}. Therefore, we will compare the utilities of TBM with N and TBM with IT:
\begin{align}
    \begin{split}
        U^{TBM}_{N} - U^{TBM}_{IT} = U^{TBM}_{N} + C_{\text{Mine}} > 0
    \end{split}
\end{align}
where $U^{TBM}_{N} \ge 0$ due to the IR for TBM, and mining cost $C_{\text{Mine}} > 0$, so the $U^{TBM}_{N} - U^{TBM}_{NG} > 0$. Therefore, the utility of TBM with N is greater than TBM with IT, ensuring IC for TBM.

\textbf{(6) IC for SBM:}

The strategy set of SBM is \{Normal (N), Improper Rank (IRa)\}. Therefore, we will compare the utilities of SBM with N and SBM with IRa:
\begin{align}
    \begin{split}
        U^{SBM}_{N} - U^{SBM}_{IRa} = U^{SBM}_{N} + C_{\text{Mine}} > 0
    \end{split}
\end{align}
where $U^{SBM}_{N} \ge 0$ due to the IR for TBM, and mining cost $C_{\text{Mine}} > 0$, so the $U^{SBM}_{N} - U^{SBM}_{IRa} > 0$. Therefore, the utility of SBM with N is greater than SBM with IRa, ensuring IC for SBM.

\subsection{Overall Condition Set}

Overall, according to the calculation in the previous subsections, the reward ($\mathcal{R}$) of each block should satisfy the following condition set (\textbf{T1 - T8}):

\textbf{T1:} To guarantee \textbf{IR} of MO, we need to let $U^{MO}_{N} \ge 0$, thus:
\begin{equation}
    \mathcal{R}_{\text{Cited}} \ge \frac{(1 - \beta) \cdot \left( Q_{\text{Selected}} \cdot (1 - s) \cdot b^{MO} + k_{\text{Transmit}} \cdot |M| \right)}{\overline{Q_{\text{Selected}}^{MO}}}
\end{equation}

\textbf{T2:} To guarantee \textbf{IR} of T, we need to let $U^{T}_{N} \ge 0$, that is:
\begin{align}
    \mathcal{R}_{\text{Cited}} & \ge \frac{(1 - \beta)}{\overline{Q_{\text{Selected}}^T} \cdot \beta} \cdot \Big( P_{\text{Comp}} \cdot D \cdot \tau \cdot |M| \notag \\
    &\quad + (1 - s) \cdot b^{T} + k_{\text{Transmit}} \cdot |M| + k_{\text{Encrypt}} \cdot |M| \notag \\
    &\quad + Q_{\text{Broadcast}} \cdot k_{\text{Transmit}} \cdot k_{\text{Expand}} \cdot |M| \notag \\
    &\quad - \left( V_{\text{RecM}} - V_{\text{Now}}^{T} + 1 \right) \cdot \circled{C} \Big)
\end{align}

\textbf{T3:} To guarantee \textbf{IR} of DBM, we need to let $U^{DBM}_{N} \ge 0$, thus:
\begin{align}
    \mathcal{R}_{\text{Deposit}}  & \ge \frac{C_{\text{Mine}}}{Q_{\text{Deposit}}}
\end{align}

\textbf{T4:} To guarantee \textbf{IR} of EBM, we need to let $U^{EBM}_{N} \ge 0$, that is:
\begin{align}
    \mathcal{R}_{\text{HashM}} & \ge \frac{1}{Q_{\text{HashM}}} \cdot \Big( C_{\text{Mine}} + k_{\text{Transmit}} \cdot k_{\text{Expand}} \cdot |M| \notag \\
    &\quad + C_{\text{GenFHEKey}} - (V_{\text{FHEM}} - V_{\text{Now}}^{EBM}) \cdot \circled{C} \Big)
\end{align}

\textbf{T5:} To guarantee \textbf{IR} of TBM, we need to let $U^{TBM}_{N} \ge 0$, thus:
\begin{align}
\begin{split}
    Q_{\text{EncryptedM}} \cdot \mathcal{R}_{\text{EncryptedM}} + Q_{\text{Cases}} \cdot \mathcal{R}_{\text{Case}} \\
    \ge C_{\text{Mine}} + Q_{Cases} \cdot C_{{\text{GenTDCase}}}^{\text{Unit}}
\end{split}
\end{align}

\textbf{T6:} To guarantee \textbf{IR} of SBM, we need to let $U^{SBM}_{N} \ge 0$, that is:
\begin{align}
    Q_{\text{VerifiedM}} & \cdot \mathcal{R}_{\text{VerifiedM}} + Q_{\text{VerifiedM}} \cdot Q_{\text{Cases}} \cdot \mathcal{R}_{\text{Verify}} \notag \\
    & \ge C_{\text{Mine}} + Q_{\text{VerifiedM}} \cdot k_{\text{Transmit}} \cdot k_{\text{Expand}} \cdot |M| \notag \\
    &\quad + Q_{\text{VerifiedM}} \cdot Q_{\text{Cases}} \cdot C_{\text{Verify}}^{\text{Unit}}
\end{align}

To satisfy \textbf{IC}, the utilities of the ``Normal" strategy should be greater than other strategies. According to Table \ref{tab_utilities}, some participants' other strategies have negative utilities, so they will choose ``Normal" obviously. Specifically, trainer T requires additional constraints for the strategy of ``Not Training" and ``Not Broadcasting":

\textbf{T7:} For \textbf{IC} of T with ``Not Training", there is a sufficient but not necessary condition that the deposit of T should not be lower than the value of the model (i.e., an effective deposit discussed in Section \ref{sec_system}). Otherwise, T will not have the motivation to train the model.
\begin{align}
\begin{split}
    {b}^{T} > (V_{\text{RecM}} - V_{\text{Now}}^{T}) \cdot \circled{C}
\end{split}
\end{align}

\textbf{T8:} For \textbf{IC} of T with ``Not Broadcasting", let $U_{N}^{T} - U_{NBr}^{T} > 0$:
\begin{align}
    \mathcal{R}_{\text{Cited}} & > \frac{1}{\overline{Q_{\text{Selected}}^T} \cdot \beta} 
    \Big( (- s) \cdot {b}^{T} + k_{\text{Encrypt}} \cdot |M| \notag \\
    &\quad + Q_{\text{Broadcast}} \cdot k_{\text{Transmit}} \cdot k_{\text{Expand}} \cdot |M| \Big) \cdot (1 - \beta)
\end{align}

The fore-mentioned condition set is listed in Section \ref{sec_incentive}.

\end{document}